\documentclass{article}

\usepackage[main, final]{neurips_2026}

\usepackage[utf8]{inputenc} 
\usepackage[T1]{fontenc}    
\usepackage{hyperref}       
\usepackage{url}            
\usepackage{booktabs}       
\usepackage{amsfonts}       
\usepackage{nicefrac}       
\usepackage{microtype}      
\usepackage{xcolor}         

\usepackage{subcaption}
\usepackage{graphicx} 
\usepackage{tcolorbox}
\usepackage{amssymb}
\usepackage{makecell}
\usepackage{float}
\usepackage{array}
\usepackage[ruled,vlined,noend]{algorithm2e}
\usepackage{wrapfig}
\usepackage[dvipsnames]{xcolor}
\usepackage{url}
\usepackage[table]{xcolor}
\usepackage{multirow}
\usepackage{subcaption}
\usepackage{enumitem}
\usepackage{amsmath}
\usepackage{mathtools}
\usepackage{amsthm}
\usepackage{bm}

\usepackage[capitalize,noabbrev]{cleveref}

\newcommand{\norm}[1]{\left\|#1\right\|}
\newcommand{\normsq}[1]{\left\|#1\right\|^2}

\def \1 {\mathbf{1}}

\def \L {\mathcal{L}}

\def \B {\mathcalB}

\def \E {\mathbb{E}}

\def \D {\mathcal{D}}

\def \B {\mathcal{B}}

\theoremstyle{plain}
\newtheorem{theorem}{Theorem}[section]

\newtheorem{lemma}[theorem]{Lemma}

\theoremstyle{definition}

\newtheorem{assumption}[theorem]{Assumption}
\theoremstyle{remark}

\definecolor{lightblue}{RGB}{222,235,247}
\definecolor{CorrectGreen}{HTML}{1B9E3C}

\title{Tune-Up Open-Weight CLIP: Optimization Framework for Self-Supervised Fine-tuning of CLIP}

\author{%
  Anant Mehta \qquad
  Xingyu Chen \qquad
  Xiyuan Wei \qquad
  Tianbao Yang\thanks{Corresponding author.}\\
  Department of Computer Science \& Engineering\\
  Texas A\&M University, TX, USA\\
  \texttt{\{anant\_mehta,tamucxy2023,xwei,tianbao-yang\}@tamu.edu}
}

\begin{document}

\maketitle

\begin{abstract}
CLIP has become a cornerstone of multimodal representation learning, yet improving its performance typically requires a prohibitively costly process of training from scratch on billions of samples. We ask a different question: \emph{Can we improve the performance of open-weight CLIP models across various downstream tasks using only existing self-supervised datasets?} Unlike supervised fine-tuning, which adapts a pretrained model to a single downstream task, our setting seeks to improve general performance across various tasks.  However, as both our experiments and prior studies reveal, simply applying standard training protocols starting from an open-weight CLIP model often fails, leading to performance degradation.  In this paper, we introduce \textbf{TuneCLIP}, a self-supervised fine-tuning framework that overcomes the performance degradation. TuneCLIP has two key components: (1) a warm-up stage of recovering optimization statistics to reduce cold-start bias, inspired by theoretical analysis, and (2) a fine-tuning stage of optimizing a new contrastive loss to mitigate the penalization on false negative pairs. Our extensive experiments show that TuneCLIP consistently improves performance across model architectures and scales. Notably, it elevates leading open-weight models like SigLIP (ViT-B/16), achieving gains of up to +2.5\% on ImageNet and related out-of-distribution benchmarks, and +1.2\% on the highly competitive DataComp benchmark, setting a new strong baseline for efficient post-pretraining adaptation. The code is available at \url{https://github.com/Optimization-AI/TuneCLIP}.
\end{abstract}

 \setlength\abovedisplayskip{4pt}
\setlength\belowdisplayskip{4pt}
\setlength{\textfloatsep}{4pt}

\section{Introduction}

Contrastive vision-language models such as CLIP which learn joint image-text representations at scale by training on hundreds of millions of large scale image-text pairs \citep{radford2021learning,cherti2023reproducible} show broad utility across downstream tasks spanning classification, cross-modal retrieval, multimodal reasoning \citep{shen2021much,zhao2023clip} and generation  \citep{ao2023gesturediffuclip}. Recent efforts to improve CLIP have primarily focused on pretraining by constructing ever larger datasets \citep{fang2023data}, designing novel objective functions \citep{qiu2023not,qiu2024cool}, or developing refined optimization algorithms \citep{qiu2024cool,wei2024fastclip}. While these directions have advanced the state of the art, they come at staggering cost due to billions of image-text pairs, massive GPU clusters, and days or weeks of computation.  In this work, we ask a complementary but equally important question ``{\it How can we unlock more from the CLIP we already have?}'', shifting from ``{\it How can we pretrain a better CLIP from scratch?}'', which leads to a path that is cheaper, faster, and far more compute-efficient.

A very common way to improve the model is {supervised fine-tuning}, which is performed on  specific datasets of a target domain~\citep{nguyen2024saft,srinivasan2023curriculum,goyal2023finetune}. These prior works leverage class labels or captions to steer the embedding space. An issue with these methods is that strong adaptation to the target domain can harm generalization contributing to reduced robustness to distribution shifts \citep{ding2022don,jha2024clap4clip}. Thus, supervised fine-tuning cannot be regarded as a procedure to improve a CLIP model in general, rather, it is a \textit{domain adaptation} for a specific distribution, often at the expense of transferability.

These limitations motivate an alternative paradigm that we study, namely \textit{self-supervised fine-tuning} (SSFT), which is the process of improving a pretrained CLIP model’s overall representational quality and general-purpose performance, rather than tailoring it to a specific downstream task. What makes SSFT actually ``self-supervised''? Traditionally, supervised fine-tuning is carried out on datasets such as ImageNet, CIFAR, or Flickr \citep{yang2023imagenet,krizhevsky2017imagenet,van2007flickr,wortsman2022robust,dong2022clip,fahes2024fine,goyal2023finetune,Liu_2025_CVPR}, which were constructed through human annotation or domain-specific filtering and are thereby inherently biased models toward a particular domain. By contrast, we use large-scale web corpora that were constructed for pretraining CLIP models~\footnote{While some studies treat CLIP as weakly supervised due to its web-crawled image-text training data, we follow~\cite{DBLP:journals/corr/abs-2108-07258} and classify it as self-supervised. }, e.g., \citep{fang2023data}. The result is task-agnostic corpora, positioning SSFT on them as representation refinement rather than task adaptation. 

\begin{wrapfigure}{r}{0.5\textwidth} 
  \vspace{-6pt} 
  \centering
  \includegraphics[width=\linewidth]{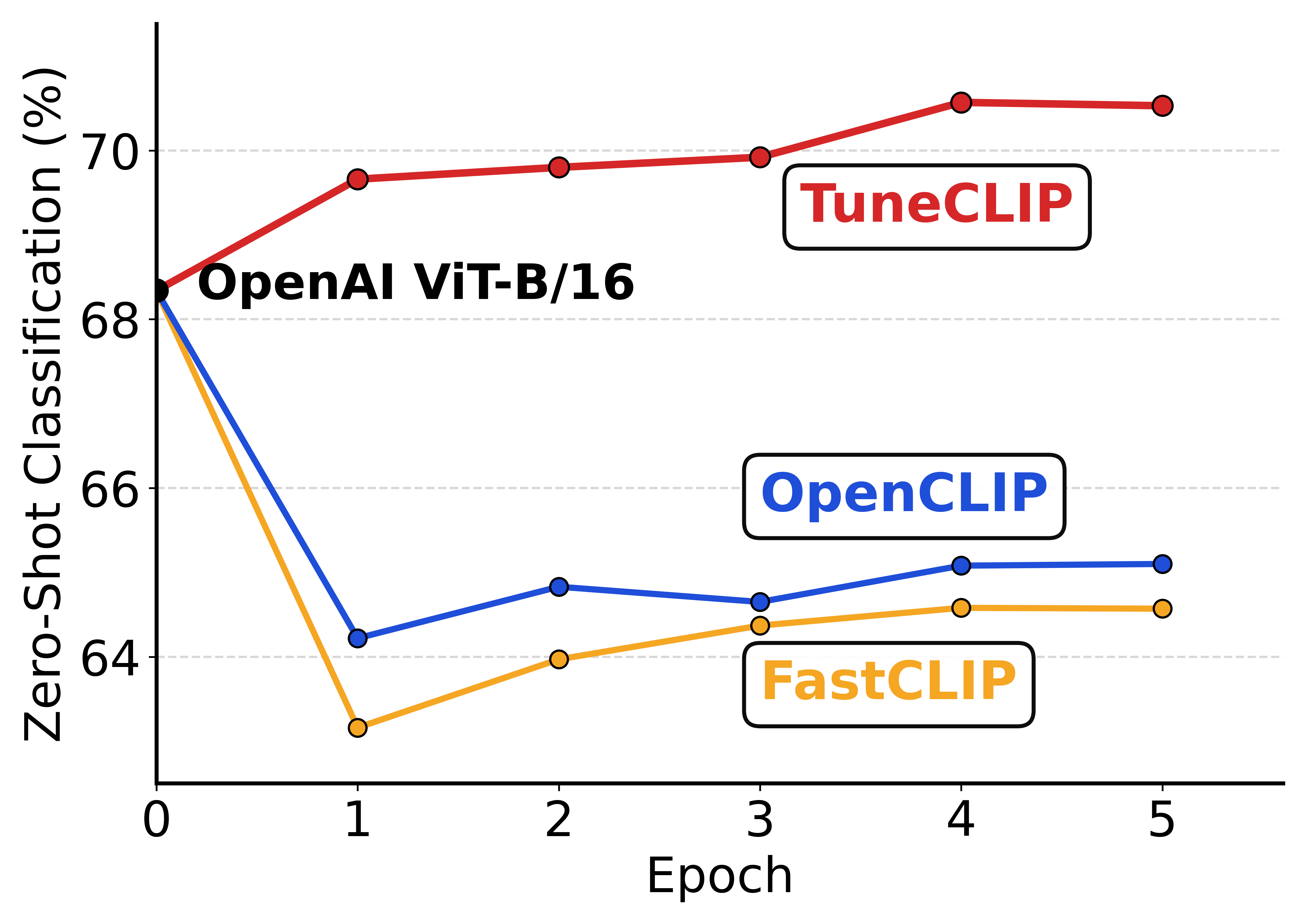}
  \vspace{-10pt} 
\caption{Early-stage performance drop  on ImageNet when fine-tuning OpenAI ViT-B/16 model on DFN-12M dataset using OpenCLIP and FastCLIP pipelines.}
  \label{fig:im1k-openai}
\end{wrapfigure}

A key challenge of applying SSFT to a well-pretrained CLIP model is a sharp 
performance degradation in the first epoch, as observed in our experiments 
(cf. Figure~\ref{fig:im1k-openai}) and corroborated by prior work~\cite{yamaguchi2025post}. 
This initial drop is difficult to recover from, even when training is extended 
for additional epochs. Although this phenomenon may appear related to catastrophic forgetting in 
continual learning~\cite{liu2025c,jha2024clap4clip,xiao2023clip,garg2023tic}, 
the underlying cause is fundamentally different. Catastrophic forgetting stems 
from distributional shift introduced by new task data; however, we observe the 
same performance drop even when training uses the same data as CLIP 
pretraining (cf. Figure \ref{img:drops-im1k} in Appendix \ref{app:in1k}). Our first contribution is a theoretical analysis of this failure  mode from an optimization perspective. We establish that the  optimization error in contrastive learning is highly sensitive to  the error of gradient estimates at initialization, and that this  sensitivity is especially pronounced when starting from a  well-pretrained model. The root cause lies in a known property of  contrastive losses — they do not admit unbiased stochastic gradient  estimators~\citep{yuan2022provable} — which renders the standard initialization of gradient estimator particularly unreliable and makes early training dynamics difficult to control. This manifests as a sharp and  hard-to-recover performance degradation at the onset of fine-tuning,  which we term \textbf{cold-start bias}.


In addition to the optimization challenge of cold-start bias, SSFT exposes a distinct learning-level issue caused by \textbf{false negatives}, whereby data semantically similar to the anchor are mistakenly treated as negatives during training. 
This problem is exacerbated as the model becomes better at capturing  semantic structure, since fine-tuning further amplifies the similarity gap  between positive and negative pairs, causing false negatives to increasingly  distort the learned embeddings. The downstream effect is a degradation in  retrieval performance, which depends critically on fine-grained ranking  among semantically similar examples. Although prior work has addressed false negatives in contrastive 
learning~\cite{DBLP:journals/corr/abs-2106-03719,DBLP:conf/icml/BalmasedaWLY25,
chuang2020debiased,huynh2022boosting}, existing methods are limited to the 
mini-batch level or require non-trivial additional overhead to detect false 
negatives globally at training time. We sidestep these limitations with a simple yet 
effective remedy: a hinge-based surrogate loss that integrates naturally into 
the global contrastive learning framework without any additional cost.



{\bf Our contributions} are summarized below:
{\setlength{\itemsep}{2pt}
 \setlength{\topsep}{2pt}
\begin{itemize}[leftmargin=*]
\item 
We provide a theoretical analysis quantifying the cold-start bias, showing how the initial gradient estimation error in contrastive loss optimization influences convergence. To mitigate this issue, we propose \emph{Optimizer Statistics Recovery (OSR)}, which restores accurate first and second-order moment estimates, along with other useful statistics of the initial model, through a warm-up stage.

\item 
To reduce the impact of false negatives, we introduce a simple yet effective remedy: the \emph{hinged global contrastive loss (HGCL)}. This loss penalizes positive and negative pairs only when their similarity gap exceeds a margin, thereby avoiding excessive penalization of false negatives. This improves retrieval performance while preserving strong zero-shot classification accuracy.

\item 
We conduct extensive experiments on SSFT across multiple pretrained models and data scales. Our results show consistent improvements over base models and demonstrate superiority to existing standard pretraining approaches that can be used for SSFT.
\end{itemize}
}

\section{Related Work}
\vspace*{-0.1in}
\textbf{Improving CLIP Pretraining.}
Extensive works improve CLIP pretraining via better objective functions and training recipes. A line of work builds upon the standard mini-batch contrastive loss~\citep{radford2021learning,cherti2023reproducible}.
For instance,  \citep{li2023scaling} incorporates image masking within the contrastive framework, \citep{DBLP:journals/corr/abs-2205-14459} enforces geometrical consistency in the learned representation of images and texts, while \cite{fan2023improving} improves CLIP training with language rewrites.  \citep{fang2023data} has proposed a data filtering network to better filter low-quality data for improving CLIP training. In addition, system-level advances in distributed training and memory efficiency have enabled scalable CLIP pretraining at web scale~\citep{sun2023eva, chen2023disco, cherti2023reproducible, wei2023improving}. 
 
However, these approaches typically rely on extremely large batch sizes, leading to significant memory and system constraints. To address this limitation, a parallel line of work replaces mini-batch contrastive loss with a global contrastive loss. Notably, \citep{yuan2022provable} introduce SogCLR  to directly optimize a global contrastive objective based on compositional optimization, achieving superior performance without requiring massive batches. Subsequent works~\citep{qiu2023not,qiu2024cool} further improve this paradigm for CLIP training through individualized temperature optimization. Later, FastCLIP~\cite{wei2024fastclip} develops an efficient distributed CLIP training framework for optimizing a global contrastive objective. Two recent works~\cite{wei2026neuclip,sun2026amorlip} address the challenge of estimating the normalization term of the softmax function by training a normalization prediction network.

Empirically, scaling laws suggest that increasing data and model capacity consistently leads to stronger CLIP models. On the data side, large-scale datasets such as Datacomp-1B~\citep{gadre2023datacomp},  DFN-2B~\citep{fang2023data}, LAION-5B~\citep{schuhmann2022laion},  and the WebLi dataset with 10B data~\cite{zhai2023sigmoid} provide high-volume web image--text pairs for training. On the model side, architectures have scaled from small Vision Transformers (ViTs)~\citep{dosovitskiy2020image} with a few million parameters to billion-parameter regimes. While data and model scaling is effective for improving performance, it also significantly increases the training budget, making it inaccessible to most researchers and developers. Recently, \cite{wei2024model} proposed a novel approach DRRHO-CLIP for improving the scaling law of CLIP training by using a pre-trained CLIP model as a reference model, which significantly reduces the  burden on large-scale data.The key distinction between our work and these prior studies is that  they train CLIP models from scratch, whereas we fine-tune an already 
well-pretrained CLIP model.

\textbf{Improving pretrained CLIP.}
The most prominent direction in this setting is supervised fine-tuning, which relies on labeled data, pseudo-labels, or task-specific supervision. FLYP~\citep{goyal2023finetune}, WiSE-FT~\citep{wortsman2022robust} constructs image--text pairs using class labels, while LP-FT~\citep{kumar2022fine} performs linear probing, again requiring labeled data. Similarly, \citep{schall2024optimizing} leverage label information in a two-stage framework based on General-Purpose Retrieval Fine-Tuning (GPR)~\citep{schall2022gpr1200}. Another line of work improves downstream performance by learning textual prompts corresponding to class labels~\citep{zhou2022learning,zhou2022conditional,khattak2023maple}. In contrast, our work advances a label-agnostic paradigm, aiming instead to enhance CLIP’s generality while preserving its robust pretrained representations.

As discussed earlier, SSFT of pretrained CLIP often suffers from early-stage performance degradation. Although not widely emphasized, this phenomenon has been observed in prior work. CLIP-Refine~\citep{yamaguchi2025post} highlights this issue, showing that directly optimizing the contrastive loss can degrade performance relative to the pretrained model. To address this, it introduces image--text feature alignment combined with a hybrid contrastive loss. While effective, its improvements rely on high-quality image--text pairs (e.g., COCO~\citep{vinyals2016show}). In our experiments, we directly compare against this framework and demonstrate the advantages of our method using web corpora.

\textbf{Continual Learning.}
An orthogonal line of work studies continual learning for CLIP, focusing on mitigating catastrophic forgetting under distributional shifts~\citep{liu2025c,jha2024clap4clip,xiao2023clip,garg2023tic}. These methods are designed to preserve performance as models adapt to new tasks or data distributions. However, we address an optimization issue that causes performance degradation even when the fine-tuning data closely matches the pretraining distribution.

\section{Preliminaries}
\vspace*{-0.1in}
\textbf{Notations:}
Let $\mathcal{D} = \{ (x_i, z_i) \}_{i=1}^n$ be a dataset of $n$ image–text pairs, where $x_i$ denotes the $i$-th image and $z_i$ denotes its corresponding text description. Given CLIP model $\mathcal{M}$ (with $|\mathcal{M}|$ parameters), we learn two separate encoders. We define $\mathbf{f}(\cdot)$ and $\mathbf{g}(\cdot)$ as the encoders for images and texts, parameterized by $\boldsymbol{\theta_1}$ and $\boldsymbol{\theta_2}$, respectively. For ease, we define the joint parameter of the image and text encoders as \(\boldsymbol{\omega} = [\boldsymbol{\theta_1}, \boldsymbol{\theta_2}]\). To ensure that cosine similarity can be consistent with the inner product, both encoders output $\ell_2$ normalized vector representations in $\mathbb{R}^d$. Thus the cosine similarity between an image $x_i$ and a text $z_j$ is 
$\mathbf{s}_{i,j} = \mathbf{f}(x_i;\boldsymbol{\omega})^\top \mathbf{g}(z_j;\boldsymbol{\omega})$. To discuss algorithms later, we need the notations for a mini-batch, so let us consider \(\mathcal{B} \subset \mathcal{D}\) having $B=|\mathcal B|$ samples to be a mini-batch sampled from the full dataset \(\mathcal{D}\). 

\textbf{Mini-batch Contrastive Loss (MBCL).}
The standard mini-batch based contrastive loss  for a batch $\mathcal{B}$ is given by~\citep{radford2021learning}:
\begin{equation}
\label{eq:CL}
\mathcal{L}_{\mathrm{MBCL}}(\boldsymbol{\omega})
= -\frac{1}{\mathcal{|B|}}\sum_{i=1}^\mathcal{|B|}
\left[
\log\frac{\exp(\mathbf{s}_{i,i}/\tau)}{\sum_{j=1}^\mathcal{|B|} \exp(\mathbf{s}_{i,j}/\tau)}
\;+\;
\log\frac{\exp(\mathbf{s}_{i,i}/\tau)}{\sum_{j=1}^\mathcal{|B|} \exp(\mathbf{s}_{j,i}/\tau)}
\right].
\end{equation}

which encourages high similarity for positive image–text pairs and low similarity for negative pairs in the shared \(\mathbb{R}^d\) space. Here \(\tau > 0\) is the temperature parameter. \cite{cherti2023reproducible} used this loss to train OpenCLIP models. 

\textbf{Global Contrastive Loss (GCL).}
One limitation of optimizing MBCL is that  it requires a large batch size in order to achieve competitive performance. To address this issue, we follow previous works~\citep{yuan2022provable} and use a Global Contrastive Loss (GCL). Without loss of generality, let us introduce a pairwise loss $\ell(\mathbf{s}_{j,i}-\mathbf{s}_{i,i})$, which measures the loss on the difference between a negative data pair and a positive data pair. Then we define two functions $\Phi_1(\cdot), \Phi_2(\cdot)$ for image-anchor data and text-anchor data, respectively, i.e., 
\begin{align*}
&\Phi_1(\boldsymbol{\omega}, i, \mathcal{D}) = \frac{1}{n}\sum_{z_j \in \mathcal{D}\setminus\{z_i\}} \exp\!\bigg(\frac{\ell(\mathbf{s}_{i,j}-\mathbf{s}_{i,i})}{\tau}\bigg),\;\Phi_2(\boldsymbol{\omega}, i, \mathcal{D}) = \frac{1}{n}\sum_{x_j \in \mathcal{D}\setminus\{x_i\}} \exp\!\bigg(\frac{\ell(\mathbf{s}_{j,i}-\mathbf{s}_{i,i})}{\tau}\bigg),
\end{align*} 
Then GCL can be defined as:
\begin{equation}
\label{eq:gcl}
\mathcal{L}_{\mathrm{GCL}}(\boldsymbol{\omega})
=
\frac{\tau}{n}\sum_{i=1}^n
\left[
\log\!\big( \varepsilon + \Phi_1(\boldsymbol{\omega}, i, \mathcal{D}) \big)
+
\log\!\big( \varepsilon + \Phi_2(\boldsymbol{\omega}, i, \mathcal{D}) \big)
\right],
\end{equation}
where \(\varepsilon>0\) is a small constant that increases numerical stability. Without explicitly mentioned, we consider $\ell(\cdot)=\cdot$ for GCL as used in~\citep{wei2024fastclip} for CLIP training from scratch.

\textbf{Optimization Algorithms.}
A fundamental challenge of optimizing GCL is that it lacks unbiased stochastic gradient estimator. To see this, the gradient of $\mathcal{L}_{\mathrm{GCL}}(\boldsymbol{\omega})$ is given by 
\[
\nabla\mathcal{L}_{\mathrm{GCL}}(\boldsymbol{\omega})=\frac{\tau}{n}\sum_{i=1}^n
\left[
\frac{1}{\varepsilon + \Phi_1(\boldsymbol{\omega}, i, \mathcal{D})}\nabla \Phi_1(\boldsymbol{\omega}, i, \mathcal{D})
+
\frac{1}{\varepsilon + \Phi_2(\boldsymbol{\omega}, i, \mathcal{D})}\nabla\Phi_2(\boldsymbol{\omega}, i, \mathcal{D})
\right]
\]
Since $\Phi_{*}(\boldsymbol{\omega}, i, \mathcal{D})$ is the denominator, simply using their mini-batch estimator will yield a biased gradient estimator. To address this issue, \cite{yuan2022provable} propose an algorithm SogCLR, which maintains and updates an estimator $u_{i,x}, u_{i,z}$ for each $\Phi_{1}(\boldsymbol{\omega}, i, \mathcal{D})$ and $\Phi_{2}(\boldsymbol{\omega}, i, \mathcal{D})$ along the optimization trajectory. At the $t$-iteration with a mini-batch $\mathcal B_{t}$, they are updated by 
\begin{equation}
\label{eq:sogclr-u}
u^{(t)}_{i,x}=(1-\gamma_t)u^{(t-1)}_{i,x}+\gamma_t\,{\Phi}_1(\boldsymbol{\omega}_{t-1},i,\mathcal{B}_t),
\qquad
u^{(t)}_{i,z}=(1-\gamma_t)u^{(t-1)}_{i,z}+\gamma_t\,{\Phi}_2(\boldsymbol{\omega}_{t-1},i,\mathcal{B}_t),
\end{equation}
Then a stochastic gradient estimator of $\mathcal{L}_{GCL}$ w.r.t.\ the shared parameters \(\boldsymbol{\omega}\) at iteration $t$ is:
\begin{equation}
\label{eq:gcl-grad-sogclr}
G(\boldsymbol{\omega}_{t-1}, \B_t)
=
\frac{\tau}{|\mathcal{B}_t|}\sum_{i\in\mathcal{B}_t}
\left[
\frac{1}{\varepsilon+u^{(t)}_{i,x}}\;\nabla_{\boldsymbol{\omega}}\Phi_1(\boldsymbol{\omega}_{t-1},i,\mathcal{B}_t)
\;+\;
\frac{1}{\varepsilon+u^{(t)}_{i,z}}\;\nabla_{\boldsymbol{\omega}}\Phi_2(\boldsymbol{\omega}_{t-1},i,\mathcal{B}_t)
\right].
\end{equation}
Then the first-order moment is updated followed by a model parameter update: 
\begin{equation}\label{eqn:mom}
\begin{aligned}
&m_{t} = \beta_1 m_{t-1} + (1-\beta_1) G(\boldsymbol{\omega}_{t-1}, \B_t)\\
&\boldsymbol{\omega}_{t}  = \boldsymbol{\omega}_{t-1} - \eta_{t} m_{t}.
\end{aligned}
\end{equation}
\cite{wei2024fastclip} has designed a distributed optimization framework FastCLIP based on the above algorithm for large-scale CLIP training. 

\section{TuneCLIP: A Two-Stage framework for Self-Supervised fine-tuning}
\label{TuneCLIP}
As outlined in the problem statement, our goal is to adapt pretrained parameters $\boldsymbol{\omega}_0$ to obtain refined weights $\boldsymbol{\omega}^\star$ that improve performance across diverse domains. In the following two subsections, we will discuss the challenges and present our solutions. 

\subsection{Stage I: Optimizer Statistics Recovery (OSR)}\label{sec:stageI}
A naive approach for SSFT with a pretrained model $\mathcal{M}$ with weights $\boldsymbol{\omega}_0$ is to just run OpenCLIP \citep{cherti2023reproducible} or FastCLIP \citep{wei2024fastclip} algorithms on an existing self-supervised learning dataset $\mathcal D$ with an initialization of $\boldsymbol{\omega}_0$. Our hypothesis is that an open-weight pretrained model $\boldsymbol{\omega}_0$ (e.g., OpenAI's ViT-B/16) is usually not an optimal model.  However, we observe a performance degradation in the first epoch of fine-tuning, see Figure~\ref{fig:im1k-openai}, with the details of training deferred to Section~\ref{sec:exp}. This phenomenon is common regardless of the model structure and datasets used for fine-tuning. 

\begin{algorithm}[t]
\caption{Optimizer Statistics Recovery (OSR)}
\label{alg:osr}
\DontPrintSemicolon  

\textbf{Init:} $ \boldsymbol{\omega_0} \;(\text{Pretrained})$, 
$m_0 \gets [0]^{|\mathcal{M}|}$, 
$v_0 \gets [0]^{|\mathcal{M}|}$, 
$u^{(0)}_{x} \gets [0]^{|\mathcal{D}|},\; 
u^{(0)}_{z} \gets [0]^{|\mathcal{D}|}$ \;

\For{iteration $t = 1$ \KwTo $T$ }{
  Sample $\mathcal{B}_t \subset \mathcal{D}$ \tcp*[r]{mini-batch sampling}
  
  $u^{(t)}_{i,x} \gets (1-\gamma_t)u^{(t-1)}_{i,x} + \gamma_t\,\Phi_1(\boldsymbol{\omega_0},i,\mathcal{B}_t), \forall i\in\mathcal{B}_t$ \tcp*[r]{refer \eqref{eq:sogclr-u}}
  $u^{(t)}_{i,z} \gets (1-\gamma_t)u^{(t-1)}_{i,z} + \gamma_t\,\Phi_2(\boldsymbol{\omega_0},i,\mathcal{B}_t), \forall i\in\mathcal{B}_t$

  Compute $g_t = G(\boldsymbol{\omega_0}, \B_t)$ \tcp*[r]{frozen $\boldsymbol{\omega_0}$}
  Update $m_t \gets \beta_1 m_{t-1} + (1-\beta_1)g_t$ \tcp*[r]{first moment}
  Update $v_t \gets \beta_2 v_{t-1} + (1-\beta_2)(g_t \odot g_t)$ \tcp*[r]{second moment}
}

\colorbox{yellow!10}{
$\textbf{Return}: \; m^\star \gets m_T,\; v^\star \gets v_T,\; 
u^\star \gets \{u^{(T)}_{i,x}, u^{(T)}_{i,z}\}_{i\in\mathcal{D}}$
}
\end{algorithm}


To understand this phenomenon, we provide a theoretical analysis of optimization error. We consider the optimization algorithm SogCLR used by FastCLIP, and note that analysis of OpenCLIP's optimization algorithm suffer from the same issue. 
To run FastCLIP algorithm, we need to initialize several statistics, including  $m_0$ and $u^{(0)}_{i,x}, u^{(0)}_{i,z},\forall i$. These statistics are usually initialized to mini-batch estimators of the first iteration in standard pretraining from scratch. Below, we show that their estimation error has a great impact on the convergence. 
To simplify the presentation, we introduce the following notations: $u_x^{(t)} = [u_{1,x}^{(t)}, \cdots, u_{n,x}^{(t)}], u_z^{(t)} = [u_{1,z}^{(t)}, \cdots, u_{n,z}^{(t)}]$, $\Phi_1(\boldsymbol{\omega_0}, \D) = [\Phi_1(\boldsymbol{\omega_0}, 1, \D), \cdots, \Phi_1(\boldsymbol{\omega_0}, n, \D)], \Phi_2(\boldsymbol{\omega_0}, \D) = [\Phi_2(\boldsymbol{\omega_0}, 1, \D), \cdots, \Phi_2(\boldsymbol{\omega_0}, n, \D)]$. Due to space limitations, all necessary assumptions and theorem proofs in this section are deferred to Appendix~\ref{app:proofs}.
\begin{theorem}\label{thm:main_thm_sox}
Let us consider the updates in~(equation \ref{eqn:mom}) with initializations $u_x^{(0)}, u_z^{(0)}$, and $m_0$. Under appropriate assumptions, with $1-\beta_1 = O(B\epsilon^2)$, $\gamma = O(B\epsilon^2)$ and $\eta = O(\frac{B^2\epsilon^2}{n})$, we  can find an $\epsilon$-stationary point $\boldsymbol{\omega}$ such that $\E[\norm{\nabla \L_{\mathrm{GCL}}(\boldsymbol{\omega})}]\leq \epsilon$ in
    \begin{align*}
        T = O\left(\frac{n}{B^2\epsilon^4}\left(\Delta_0 + \frac{B}{n}M_0 + U_{x,0} + U_{z,0}\right)\right)
    \end{align*}
    iterations, where $\Delta_0 = \L_{\mathrm{GCL}}(\boldsymbol{\omega_0}) - \min_{\boldsymbol{\omega}} \L_{\mathrm{GCL}}(\boldsymbol{\omega})$, $M_0 = \normsq{m_0 - \nabla_\omega \L_{\mathrm{GCL}}(\boldsymbol{\omega_0})}$, $U_{x,0} = \frac{1}{2n}\normsq{u_x^{(0)} - \Phi_1(\boldsymbol{\omega_0}, \D)}$, $U_{z,0} = \frac{1}{2n}\normsq{u_z^{(0)} - \Phi_2(\boldsymbol{\omega_0}, \D)}$. 
\end{theorem}
{\bf Remark:} The above theorem exhibits how the initial estimation errors of $u_x^{(0)}, u_z^{(0)}$, and $m_0$ affects the iteration complexity for finding an $\epsilon$-stationary solution. Since a pretrained model $\boldsymbol{\omega_0}$ is already well trained, we expect $\Delta_0$ to be small. However, the initial estimation errors of $u_x^{(0)}, u_z^{(0)}$, and $m_0$ could be very large if they are initialized poorly. It is these errors that cause the break in convergence and hence the performance degradation at the beginning of training. We refer to this issue caused by the initial estimation errors of statistics $\Phi_1(\boldsymbol{\omega_0}, \D), \Phi_2(\boldsymbol{\omega_0}, \D),  \nabla_\omega \L_{\mathrm{GCL}}$ as cold-start bias. 

To address the cold-start bias, we propose a simple method that aims to compute a better estimation of statistics $\Phi_1(\boldsymbol{\omega_0}, \D), \Phi_2(\boldsymbol{\omega_0}, \D),  \nabla_\omega \L_{\mathrm{GCL}}$ for updating $\boldsymbol{\omega_0}$. The idea is just run the update~(equation \ref{eqn:mom}) with the model parameter fixed at $\boldsymbol{\omega_0}$. We present the details in Algorithm~\ref{alg:osr}, which is referred to as optimizer statistics recovery (OSR). The following theorem provides a guarantee that the estimation errors of returned statistics of OSR would be much reduced. In practice, we also compute a second moment estimator for using Adam optimizer, which improves performance in our experiments. 
\begin{theorem}\label{thm:stageI}
     Let Algorithm~\ref{alg:osr} run for $E$ epochs (equivalently $T = E\cdot \frac{n}{B}$ iterations) with $1-\beta_1 = O(\sqrt{\frac{B}{E}}), \gamma = O(\sqrt{\frac{B}{E}})$, we have that:
    \begin{align}
        &\mathbb E\bigg[\frac{1}{2n}\normsq{u_x^{(\tau)} - \Phi_1(\boldsymbol{\omega_0}, \D)} + \frac{1}{2n}\normsq{u_z^{(\tau)} - \Phi_2(\boldsymbol{\omega_0}, \D)}\bigg]\leq O\left(\frac{U_{x,0} + U_{z,0}}{\sqrt{BE}} + \frac{1}{\sqrt{BE}}\right),\\
        &\mathbb E\bigg[\normsq{m_\tau - \nabla_\omega \L_{\mathrm{GCL}}(\boldsymbol{\omega_0})}\bigg]\leq O\left(\frac{\frac{B}{N}M_0 + U_{x,0} + U_{z,0}}{\sqrt{BE}} + \frac{1}{\sqrt{BE}}\right)
    \end{align}
    where $\tau\in\{0,\ldots, T-1\}$ is randomly sampled. 
\end{theorem}
We observe that $E=5$ epochs for OSR is good enough to ensure stable training in the second stage of updating the model parameters.

We also note that MBCL is a special case of GCL with $\gamma=1$, i.e., replacing Eq.~\eqref{eq:sogclr-u} with $u_{i,x}^{(t)}=\Phi_1(\omega_{t-1},i,\mathcal{B}_t)$ and $u_{i,z}^{(t)}=\Phi_2(\omega_{t-1},i,\mathcal{B}_t)$. Thus, a similar analysis applies to MBCL, yielding a similar dependence on the error of gradient momentum initialization. For more detail analyses, please refer to Appendix.~\ref{sec:analysis_MBCL}

\begin{algorithm}[t]
\caption{TuneCLIP Algorithm}
\label{alg:sogclr-hgcl}
\DontPrintSemicolon  

\textbf{Given:} $\boldsymbol{\omega}_0 \;(\text{Pretrained})$, dataset $\mathcal{D}$, batch size $|\mathcal{B}|$, 
epochs $E'$, $\tau$, margin $m$, $\gamma_t$, Adam $(\beta_1,\beta_2)$ \;

$(m^\star, v^\star, \{u^\star_{i,x}, u^\star_{i,z}\}_{i\in\mathcal{D}})
\gets \mathrm{OSR}(\boldsymbol{\omega}_0,\mathcal{D})$ \tcp*{Alg.~\ref{alg:osr}}

\textbf{Init:} $\boldsymbol{\omega}\gets\boldsymbol{\omega}_0$; $m_0\gets m^\star$; $v_0\gets v^\star$; 
$u^{(0)}_{i,x}\gets u^\star_{i,x},\; u^{(0)}_{i,z}\gets u^\star_{i,z}$ for all $i\in\mathcal{D}$ \;

\For{iteration $t = 1 \;\KwTo\; T'$ }{
  Sample $\mathcal{B}_t \subset \mathcal{D}$ \tcp*[r]{mini-batch sampling}
  
  \For{each $i \in \mathcal{B}_t$}{
    ${\Phi}^{m}_{1}(\boldsymbol{\omega},i,\mathcal{B}_t)\gets\frac{1}{|\mathcal{B}_t|}\sum_{z_j\in\mathcal{B}_t\setminus\{z_i\}}\exp\!\left(\frac{\ell(\mathbf{s}_{i,j}-\mathbf{s}_{i,i})}{\tau}\right)$ \tcp*[r]{\eqref{eq:phi12-hinge}}
    
    ${\Phi}^{m}_{2}(\boldsymbol{\omega},i,\mathcal{B}_t)\gets\frac{1}{|\mathcal{B}_t|}\sum_{x_j\in\mathcal{B}_t\setminus\{x_i\}}\exp\!\left(\frac{\ell(\mathbf{s}_{j,i}-\mathbf{s}_{i,i})}{\tau}\right)$ \tcp*[r]{\eqref{eq:phi12-hinge}}

    $u^{(t)}_{i,x}\gets(1-\gamma_t)u^{(t-1)}_{i,x}+\gamma_t\,{\Phi}^{m}_{1}(\boldsymbol{\omega},i,\mathcal{B}_t),\;
    u^{(t)}_{i,z}\gets(1-\gamma_t)u^{(t-1)}_{i,z}+\gamma_t\,{\Phi}^{m}_{2}(\boldsymbol{\omega},i,\mathcal{B}_t)$ \;
    
  }
  
  $\tilde{g}_t \gets \dfrac{\tau}{|\mathcal{B}_t|}\sum_{i\in\mathcal{B}_t}\!\left[
    \dfrac{1}{\varepsilon+u^{(t)}_{i,x}}\nabla_{\boldsymbol{\omega}}{\Phi}^{m}_{1}(\boldsymbol{\omega},i,\mathcal{B}_t)+
    \dfrac{1}{\varepsilon+u^{(t)}_{i,z}}\nabla_{\boldsymbol{\omega}}{\Phi}^{m}_{2}(\boldsymbol{\omega},i,\mathcal{B}_t)
  \right]$ \;
  
  Update $m_t$, $v_t$, and $\boldsymbol{\omega}$ using Adam-style optimizer with gradient $\tilde{g}_t$ \;
}

\textbf{Return:} $\boldsymbol{\omega}^\star \gets \boldsymbol{\omega}$ \tcp*[r]{Best parameters after last iteration}
\end{algorithm}

\subsection{Stage II: Hinged Global Contrastive Loss}
With accurate initializations of $m_0$ and $u^{(0)}_{i,x}, u^{(0)}_{i,z},\forall i$ found by OSR, we continue fine-tuning $\boldsymbol{\omega_0}$ with the SogCLR algorithm. This brings evident improvement across a variety of tasks. However, one issue is that the retrieval performance could still decline as fine-tuning progresses. We depict result of the fine-tuning of SigLIP ViT-B/16 on the DFN dataset (see Figure~\ref{fig:ssft_combined}), where the retrieval performance on the fine-tuning dataset keeps increasing but the retrieval performance on testing data such as Flickr decreases. This trend holds across pretrained models (Figure~\ref{fig:main_results_row1}, Appendix~\ref{app:benefits_hgcl}).


\setlength{\fboxrule}{1pt} 
\setlength{\fboxsep}{1pt}  

\begin{figure}[t]
  \centering
  \fbox{%
    \colorbox{red!20}{%
      \setlength{\fboxsep}{1pt}%
      \begin{subfigure}{0.30\textwidth}
        \centering
        \includegraphics[width=\linewidth]{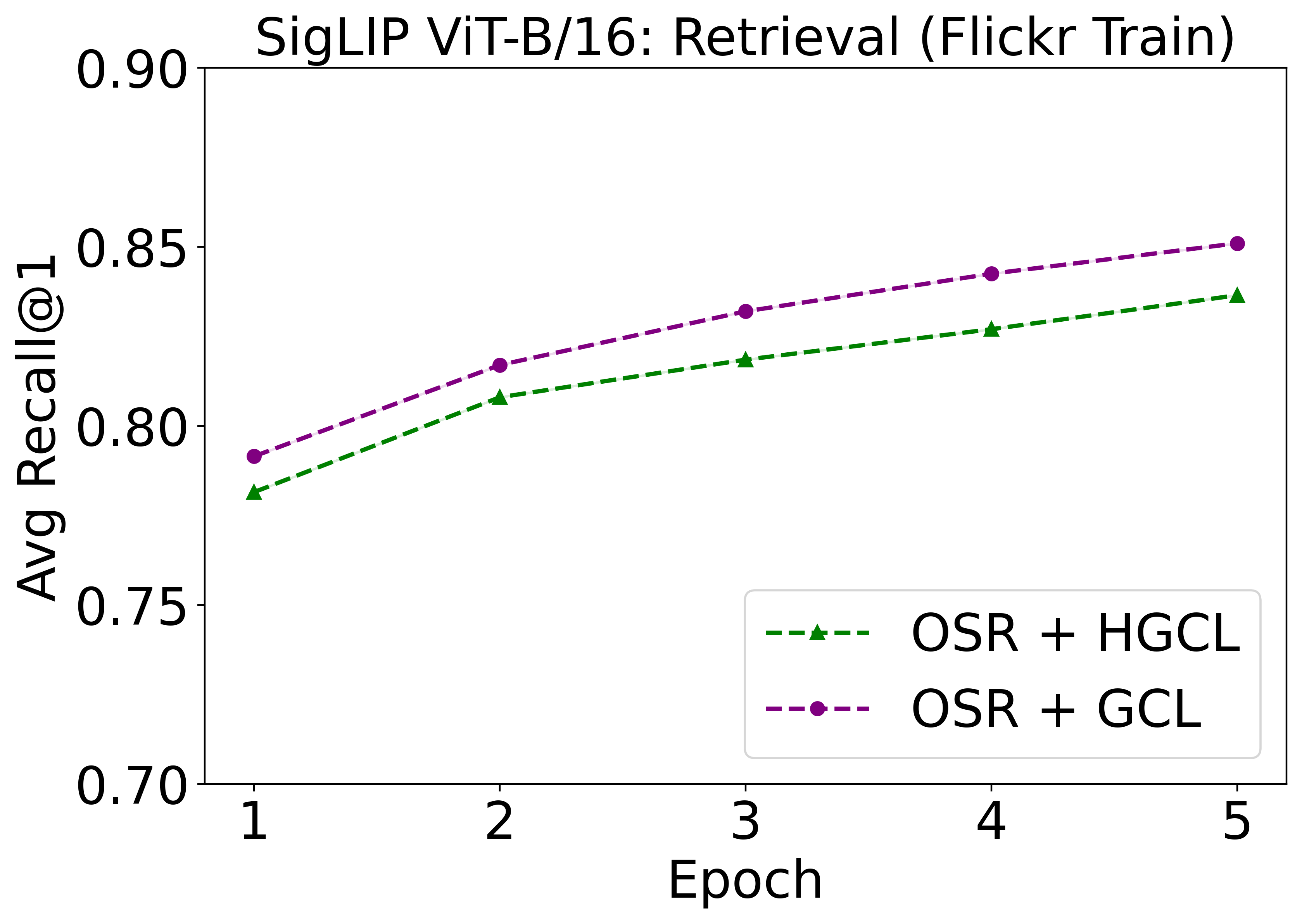}
        \caption{Supervised finetuning}
        \label{fig:flickr2}
      \end{subfigure}%
    }%
  }\hfill
  \fbox{%
    \colorbox{blue!10}{%
      \setlength{\fboxsep}{2pt}%
      \begin{subfigure}{0.65\textwidth}
        \centering
        \includegraphics[width=0.46\linewidth]{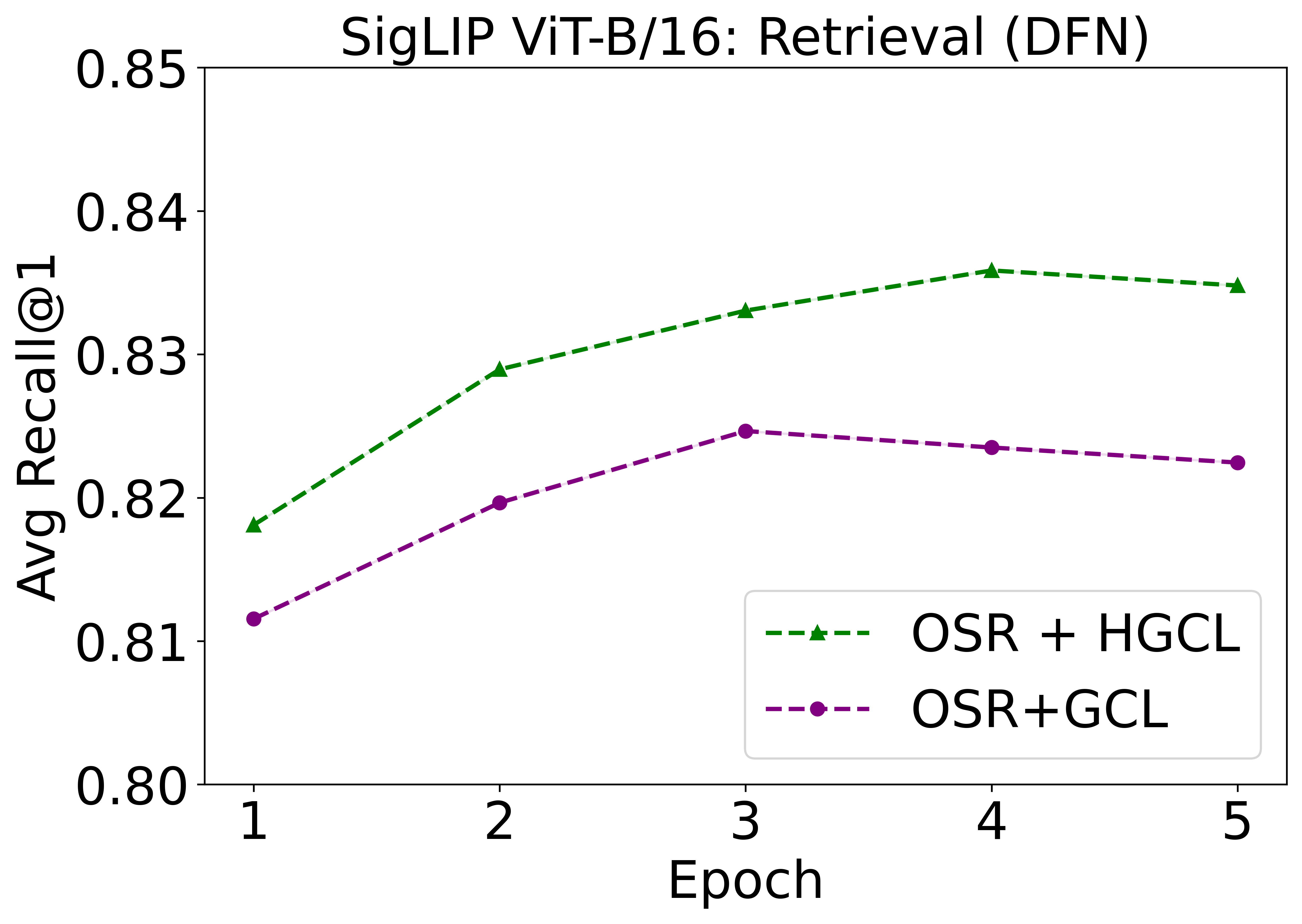}\hspace{8pt}%
        \includegraphics[width=0.46\linewidth]{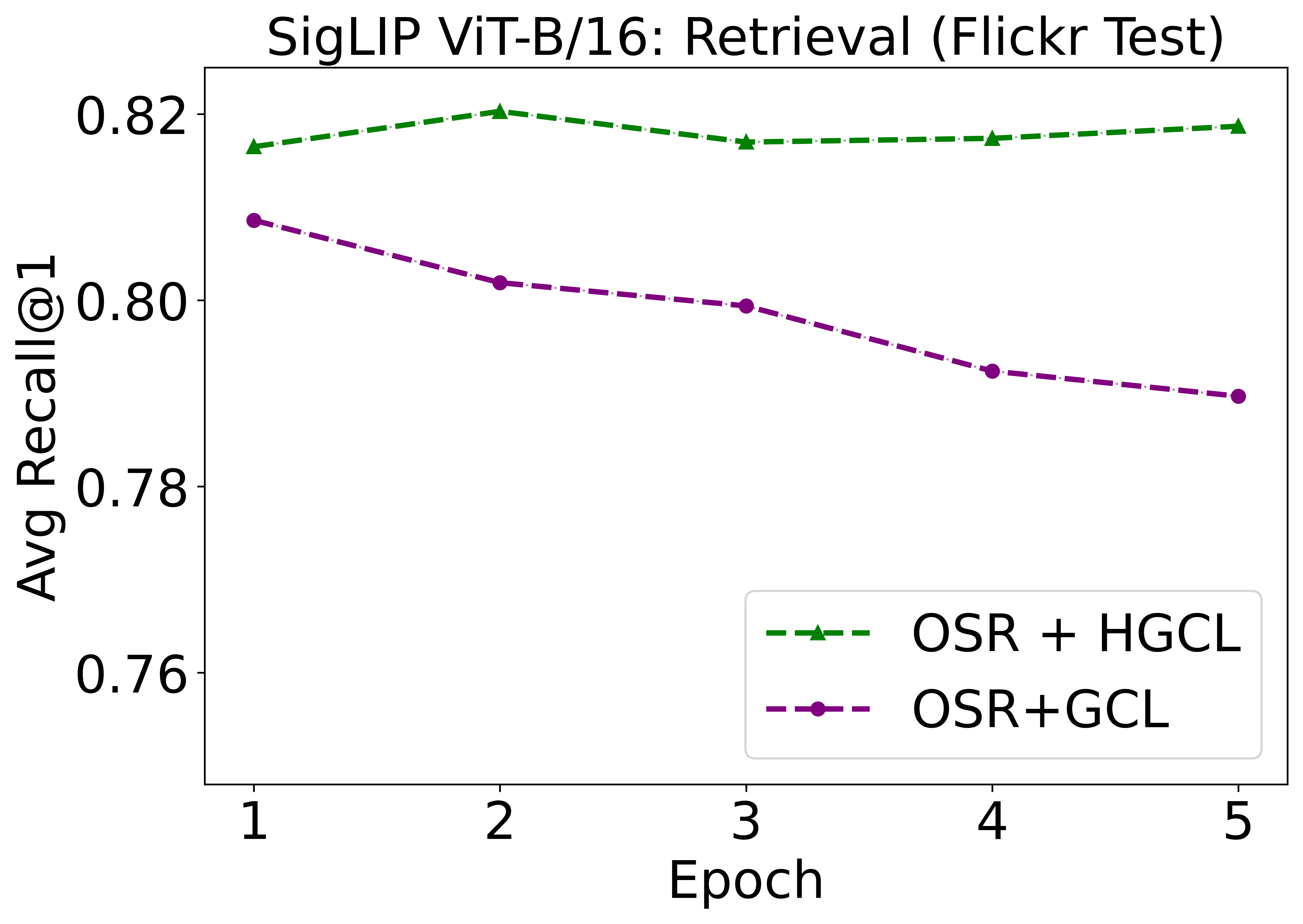}
        \caption{Self-Supervised finetuning}
        \label{fig:ssft_combined}
      \end{subfigure}%
    }%
  }
\caption{
In supervised fine-tuning (red), OSR+GCL outperforms OSR+HGCL (TuneCLIP) because true negative labels justify separating negatives. In contrast, under self-supervised fine-tuning (blue), the absence of such labels makes OSR+HGCL more suitable, leading to improved retrieval performance on Flickr when fine-tuned with SSFT (see Appendix~\ref{app:benefits_hgcl} for OpenAI CLIP \& details).
}
  \label{fig:main_results_row2}
\end{figure}


We attribute this generalization gap to the prevalence of false negatives in web-scale datasets. By optimizing GCL with $\ell(\cdot)=\cdot$, we keep decreasing the similarity gap $s_{ij} - s_{ii}$ and $s_{ji} - s_{ii}$ across iterations. If $(x_i, z_j)$ are semantically similar, e.g., $z_j$ is the caption of an image $x_j$ that is semantically similar to $x_i$, then minimizing $s_{ij} - s_{ii}$ would distort well learnt embeddings of $x_i, z_j$.  This strict separation on training data will undermine the testing performance due to distributional shift. This is the reason that leads to the retrieval performance drop. 

To mitigate this over-penalization of false negatives, we introduce a simple yet effective remedy by using a hinge-based pairwise surrogate loss $\ell(s_{ij}-s_{ii}) = \max(s_{ij} - s_{ii}+m,0)^2$ with $m>0$ being a margin hyperparameter constant. It means that as long as $s_{ii}>s_{ij}+m$, its gradient will become zero and hence will not affect the model updates anymore. Illustrative examples of this phenomenon are provided in Table ~\ref{tab:false_negatives} (Appendix \ref{app:fn_hgcl}). 
Accordingly, we define new $\Phi_1^{m}(.)$ (image-anchored) and $\Phi_2^{m}(.)$ (text-anchored) as:
\begin{equation}
\label{eq:phi12-hinge}
\begin{aligned}
\Phi^{m}_{1}(\boldsymbol{\omega}, i, \mathcal{D}) 
&= \frac{1}{|\mathcal{D}|}\sum_{j \in \mathcal{D}\setminus\{i\}} 
   \exp\!\left(\frac{\ell(\mathbf{s}_{i,j}-\mathbf{s}_{i,i})}{\tau}\right), 
&\quad \ell(\mathbf{s}_{i,j}-\mathbf{s}_{i,i}) = [\,\mathbf{s}_{i,j}-\mathbf{s}_{i,i}+m\,]^2_+, \\[6pt]
\Phi^{m}_{2}(\boldsymbol{\omega}, i, \mathcal{D}) 
&= \frac{1}{|\mathcal{D}|}\sum_{j \in \mathcal{D}\setminus\{i\}} 
   \exp\!\left(\frac{\ell(\mathbf{s}_{j,i}-\mathbf{s}_{i,i})}{\tau}\right), 
&\quad \ell(\mathbf{s}_{j,i}-\mathbf{s}_{i,i}) = [\,\mathbf{s}_{j,i}-\mathbf{s}_{i,i}+m\,]^2_+.
\end{aligned}
\end{equation}

Equation~\ref{eq:phi12-hinge} leads to the Hinged Global Contrastive Loss (HGCL), defined below.  
\begin{equation}
\label{eq:lhgcl}
\mathcal{L}_{\mathrm{HGCL}}(\boldsymbol{\omega})
= \frac{\tau}{|\mathcal{D}|}\sum_{i \in \mathcal{D}}
\left[
\log\!\big(\varepsilon + \Phi^{m}_1(\boldsymbol{\omega}, i, \mathcal{D})\big)
+ 
\log\!\big(\varepsilon + \Phi^{m}_2(\boldsymbol{\omega}, i, \mathcal{D})\big)
\right].
\end{equation}

We optimize \(\mathcal{L}_{\mathrm{HGCL}}\) using the SogCLR algorithm with OSR. Algorithm~\ref{alg:sogclr-hgcl} presents the details of our final algorithm named TuneCLIP, combining OSR with HGCL. 

Crucially, we note that the margin \(m\) is a hyperparameter that controls how aggressively negatives are separated from the positive. A larger \(m\) enforces stricter separation, pushing more false negatives downward until their similarity ($s_{i,j}$ or $s_{j,i}$) lies at least \(m\) below the positive score ($s_{i,i}$), even when they start with relatively high similarity. Conversely, a smaller \(m\) relaxes this constraint, allowing higher-scoring false negatives to be retained but at the risk of insufficient separation of true negatives. Choosing \(m\) therefore presents a \textit{tradeoff} between alleviating the over-suppression of semantically related false negatives and preventing true negatives from remaining too close to the anchor.

\section{Experiments}
\label{sec:exp}
\textbf{CLIP models \& datasets.} We explore a range of pretrained CLIP models at different scales (ranging from 151 to 987 million parameters), including OpenAI's CLIP ViT-B/32 and ViT-B/16~\cite{radford2021learning}, LAION's CLIP ViT-B/32~\cite{schuhmann2022laion} and SigLIP ViT-B/16~\cite{zhai2023sigmoid}, where ViT-B/X refers to ViT based image encoder.  We report results for fine-tuning OpenAI's ViT-B/16 and SigLIP ViT-B/16 in the main paper, and provide results of fine-tuning other models in Appendix~\ref{app:fullresults}. We additionally evaluated our method for fine-tuning the state-of-the-art CLIP ViT-H/14 pretrained on DFN-5B ~\citep{fang2023data}. To study how performance scales with data under fixed training conditions, we fine-tune on two subsets of the DFN datasets~\citep{fang2023data} containing 12 million (DFN-12M) and 60 million (DFN-60M) samples.

\begin{table*}[t]
\centering
\caption{Summary of mean zero-shot performance across ImageNet \& distributional variants, retrieval benchmarks, and the DataComp benchmark, together with wall-clock training time per GPU.}
\begin{tabular}{llcccc}
\toprule
\textbf{Base Model} & \textbf{Method} 
& \textbf{Wall C.T. (hrs)} 
& \textbf{IN \& Variants} 
& \textbf{Retrieval} 
& \textbf{DataComp} \\
\midrule

\multirow{4}{*}{\makecell[c]{OpenAI\\ViT-B/16}}
& Baseline               
& N/A
& 57.67 & 57.46 & 56.26 \\
& FastCLIP           
& 4.21
& 54.57 ($\downarrow$) & 51.88 ($\downarrow$) & 53.53 ($\downarrow$) \\
& OpenCLIP    
& 5.46
& 54.99 ($\downarrow$) & 57.81 & 55.11 ($\downarrow$) \\
& CLIP-Ref.  
& 1.01
& 55.96($\downarrow$) & 58.82 & 56.18($\downarrow$) \\
\rowcolor{lightblue!100}
& \textbf{TuneCLIP}  
& \textbf{8.62}
& \textbf{59.36} \textcolor{CorrectGreen}{\textbf{(+1.69)}}
& \textbf{64.12} \textcolor{CorrectGreen}{\textbf{(+6.66)}}
& \textbf{58.62} \textcolor{CorrectGreen}{\textbf{(+2.36)}} \\
\midrule

\multirow{4}{*}{\makecell[c]{SigLIP\\ViT-B/16}}
& Baseline               
& N/A
& 63.12 & 69.32 & 62.32 \\
& FastCLIP           
& 4.28
& 39.22 ($\downarrow$) & 43.37 ($\downarrow$) & 45.80 ($\downarrow$) \\
& OpenCLIP   
& 7.55
& 40.21 ($\downarrow$) & 51.54 ($\downarrow$) & 48.10 ($\downarrow$) \\
& CLIP-Ref.  
& 1.07
& 39.12($\downarrow$) & 49.83 ($\downarrow$) & 44.72($\downarrow$)\\
\rowcolor{lightblue!100}
& \textbf{TuneCLIP}  
& \textbf{9.27}
& \textbf{65.58} \textcolor{CorrectGreen}{\textbf{(+2.46)}}
& \textbf{69.44} \textcolor{CorrectGreen}{\textbf{(+0.11)}}
& \textbf{63.47} \textcolor{CorrectGreen}{\textbf{(+1.15)}} \\
\bottomrule
\end{tabular}
\label{tab:summary_means}
\end{table*}

\textbf{Training hyperparameters \& algorithms.} We run OSR for $E=5$ epochs, and another $E^{'}=5$ epochs for fine-tuning. Optimizer used is AdamW~\citep{loshchilov2018decoupled} (\(\beta_1{=}0.9,\,\beta_2{=}0.98\)). The CLIP temperature \(\tau\) remains fixed as provided with checkpoint (no scheduling). We sweep learning rates \(\{10^{-4},10^{-5},10^{-6}\}\). Batch sizes are \(256\times 8\) GPUs for ViT-B/16 and \(512\times 8\) GPUs for ViT-B/32 CLIPs. The margin $m$ is swept from 0.5 down to 0.01, with values around $0.1$ proving to be the most effective across the majority of architectures. More details are provided in Appendices~\ref{app:clip-details} \& \ref{app:hyp}.

\textbf{Evaluation protocol and metrics.}  
We follow the DataComp protocol \citep{gadre2023datacomp} and use 38 benchmark datasets. Our main results are reported in three evaluation groups: (1) ImageNet-1k and six robustness variants \citep{krizhevsky2017imagenet} for assessing zero-shot classification accuracy,  (2) \textcolor{black}{MSCOCO~\citep{vinyals2016show}} and Flickr30k \citep{van2007flickr} \textcolor{black}{for measuring multi \& single-object retrieval performance}, and (3) the full DataComp~\citep{gadre2023datacomp} benchmark.

\begin{table*}[t]
\centering
\caption{Impact of OSR statistics transfers on HGCL fine-tuning (OpenAI ViT-B/16 CLIP): $(m_t,v_t)$ yield a large gain, with $u_t$ providing additional improvement for the best overall performance.}
\begin{tabular}{lcccc}
\toprule
($m_t$, $v_t$) $u_t$ & \textbf{IN \& Variants} & \textbf{Retrieval} & \textbf{DataComp} & \textbf{Mean} \\
\midrule
($\times$, $\times$) $\times$ & 54.91 & 58.64 & 54.49 & 56.01 \\
(\checkmark, \checkmark) $\times$ & \textbf{59.48} \textcolor{CorrectGreen}{\textbf{(+4.57)}} & 63.70 & 58.56 & 60.58 \\
\rowcolor{lightblue!100}
(\checkmark, \checkmark) \checkmark & 59.36 & \textbf{64.12} \textcolor{CorrectGreen}{\textbf{(+5.48)}} & \textbf{58.62} \textcolor{CorrectGreen}{\textbf{(+4.13)}} & \textbf{60.70} \textcolor{CorrectGreen}{\textbf{(+4.69)}} \\
\bottomrule
\end{tabular}
\label{tab:ut_transfer}
\end{table*}

{\bf Baselines.} We compare with three baselines, FastCLIP~\cite{wei2024fastclip}, OpenCLIP~\cite{cherti2023reproducible} and CLIP-Refine~\cite{yamaguchi2025post}. We run 5 epochs for all except for CLIP-Refine, which is run for 1 epoch as suggested in their work. 

{\bf Main Results.} We present results on three evaluation suites for fine-tuning various models on DFN-12M in Table~\ref{tab:summary_means}.  We observe that TuneCLIP delivers substantial gains over the base model, most notably for OpenAI ViT-B/16 with $6.7\%$ improvement on retrieval and $1.7\%$ improvement on zero-shot classification, while improvements for SigLIP are smaller given its stronger baseline. In contrast, the baseline methods OpenCLIP~\citep{cherti2023reproducible} and FastCLIP~\citep{wei2024fastclip} not only fail to improve the performance over the base model but also suffer significant performance drop in retrieval and zero-shot classification. Similar to other baselines, CLIP-Refine also fails to consistently improve performance. This behavior aligns with observations in~\citep{yamaguchi2025post}, where performance is shown to be sensitive to data quality. In contrast, TuneCLIP avoids such degradation and delivers stable gains. For reference, we also provide more detailed results for all 38 benchmarks in Appendix ~\ref{app:fullresults}. 

Table~\ref{tab:summary_means} also reports wall-clock time and GPU-hours of different methods. Despite a modest overhead from its two-stage design, TuneCLIP consistently improves performance, whereas baselines show limited gains even with similar compute budget. This is illustrated in Figure~\ref{img:longer-ft} (Appendix~\ref{app:in1k}).

We also plot the \textcolor{black}{curves} of zero-shot classification performance on ImageNet-1k across training epochs for  different checkpoints of TuneCLIP for fine-tuning OpenAI ViT-B/16 in Figures~\ref{fig:im1k-openai}, and for other models  in Appendix~\ref{app:in1k}. 

Finally, TuneCLIP for fine-tuning the state-of-the-art model ViT-H/14-quickgelu~\citep{fang2023data} achieves new SOTA  accuracy on ImageNet and its variants as shown in Appendix \ref{app:h14}, surpassing ViT-H/14 at 224$\times$224 image resolution by about 1.5\% \textcolor{black}{(from 71.80\% to 73.23\%)}, while maintaining comparable performance on Retrieval and DataComp. 

\subsection{Ablation, Scaling \& Comparison of TuneCLIP}
We begin with an ablation study on the effect of OSR, comparing TuneCLIP with full statistics recovery, partial recovery of $m_t$ and $v_t$, and no recovery at all.
The results are reported in Table~\ref{tab:ut_transfer}, which shows that using the full recovered statistics from OSR achieves the best, and the recovery of first and second-order moments is more important than the recovery of $u_t$ (i.e., $u_x, u_z$). Beyond the ablations conducted within OSR itself, we also compare against simple cold-start mitigation heuristics that practitioners might reasonably try to stabilize fine-tuning, as presented in Appendix~\ref{app:cold-start_bias} (Table~\ref{tab:cold_start_baselines}). These alternatives offer only limited stability and smaller gains, reinforcing that OSR provides a more effective and reliable solution to cold-start bias.

\begin{figure*}[t]
\centering
\begin{minipage}[t]{0.45\textwidth}
    \centering
     \includegraphics[width=\textwidth]{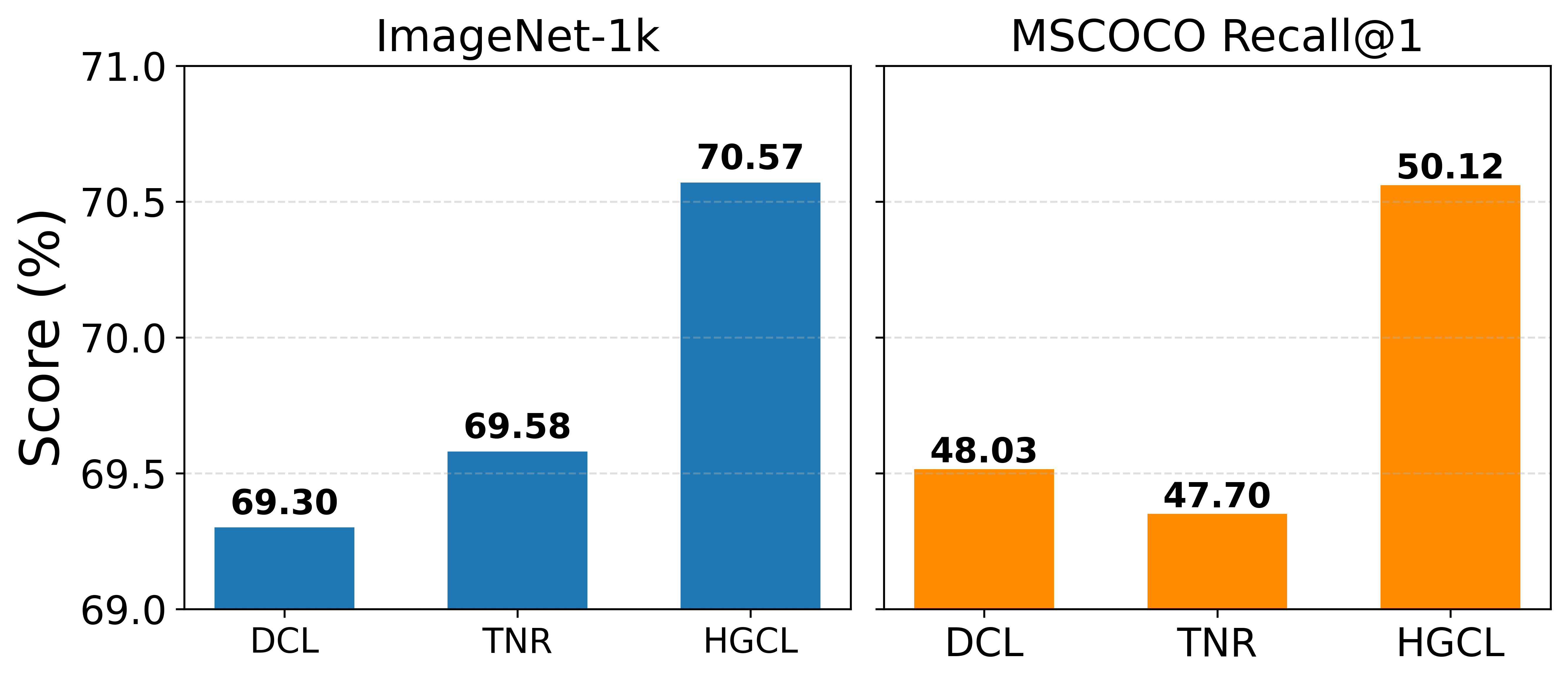}   
    \caption*{(a)}
\end{minipage}
\hfill
\begin{minipage}[t]{0.45\textwidth}
    \centering
     \includegraphics[width=0.9\linewidth]{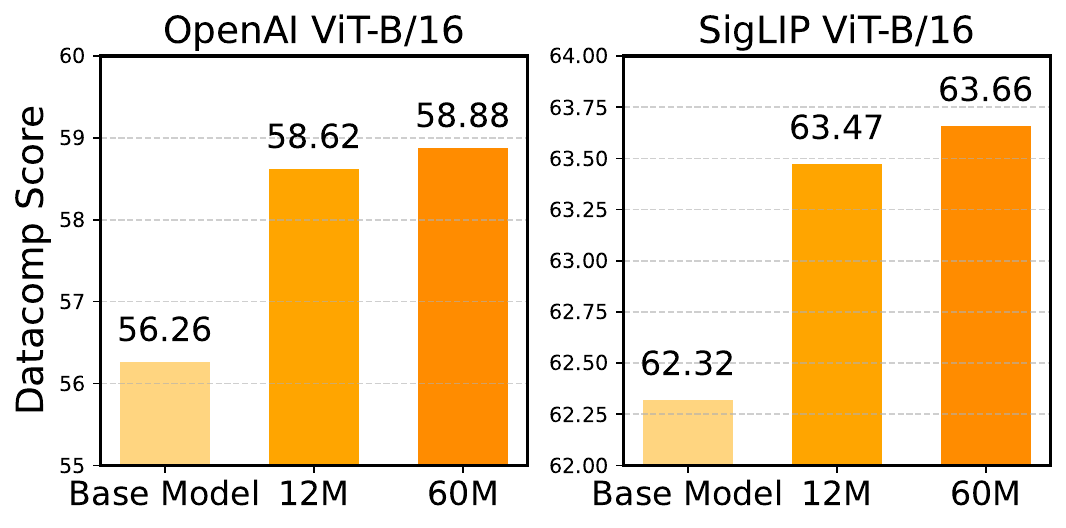}
    \caption*{(b)}
\end{minipage}
\caption{(a) Comparison of different false negative mitigation strategies in the second stage. DCL extends debiased contrastive learning~\cite{chuang2020debiased} to CLIP, TNR removes top negatives during gradient estimation (based on ~\cite{huynh2022boosting}). All use OSR in first stage, initialized from OpenAI B/16. (b) Effect of data scaling on TuneCLIP, demonstrating stable monotonic improvements as data scale increases.}
\label{fig:combined}
\end{figure*}

\begin{table*}[h]
\centering
\caption{TuneCLIP performance across datasets of varying noise levels.}
\label{tab:cc12m_dfn12m}
\begin{tabular}{lcccc}
\toprule
{\color{black}\textbf{Method}} 
& {\color{black}\textbf{Data}} 
& {\color{black}\textbf{IN \& Variants}} 
& {\color{black}\textbf{Retrieval}} 
& {\color{black}\textbf{DataComp}} \\
\midrule
{\color{black}OpenAI CLIP} 
& {\color{black}--}
& {\color{black}57.67} 
& {\color{black}57.46} 
& {\color{black}56.26} \\
\rowcolor{lightblue!100}
{\color{black}TuneCLIP} 
& {\color{black}CC12M}
& {\color{black}57.68 {\textbf{\color{CorrectGreen}(+0.01)}}} 
& {\color{black}\textbf{65.83} {\textbf{\color{CorrectGreen}(+8.37)}}} 
& {\color{black}56.47 {\textbf{\color{CorrectGreen}(+0.21)}}} 
\\
\rowcolor{lightblue!100}
{\color{black}TuneCLIP} 
& {\color{black}DFN12M}
& {\color{black}\textbf{59.36} {\textbf{\color{CorrectGreen}(+1.69)}}}
& {\color{black}64.12 {\textbf{\color{CorrectGreen}(+6.66)}}}
& {\color{black}\textbf{58.62} {\textbf{\color{CorrectGreen}(+2.36)}}} \\
\bottomrule
\end{tabular}
\end{table*}

We also conduct an ablation study comparing GCL with HGCL, both with OSR for supervised fine-tuning and SSFT. For supervised fine-tuning, we fine-tune a pretrained model on the training set of Flickr30k data and evaluate on a testing set of Flickr1k. For SSFT, we finetune the same pretrained model on DFN-12M and evaluate on the same testing set of Flickr1k. The results are shown in Figure~\ref{fig:main_results_row2} for fine-tuning SigLIP ViT-B/16 and in Figure~\ref{fig:main_results_row1} (Appendix~\ref{app:benefits_hgcl}) for fine-tuning OpenAI's CLIP ViT-B/16. These results show that GCL with OSR performs better for supervised fine-tuning, while HGCL with OSR is superior for SSFT, highlighting the role of false negatives in SSFT and the effectiveness of HGCL in improving retrieval performance. In Appendix~\ref{app:fn_hgcl}, we further show that HGCL reduces the variance of similarity scores for false negatives. 
Additionally, the true positive (top-1) distribution becomes closer to that of false negatives after fine-tuning. We analyze the sensitivity of the HGCL margin parameter $m$ in Figure~\ref{fig:margin_comp} (Appendix~\ref{app:hyp}).

We further compare HGCL with prior false negative mitigation methods~\cite{chuang2020debiased,huynh2022boosting} in Figure~\ref{fig:combined}(a), which shows that HGCL consistently outperforms both baselines. 

We also evaluate TuneCLIP on CNN-based CLIP, showing improvements beyond Vision Transformer backbones (Table~\ref{tab:resnet50_results}, Appendix~\ref{app:resnet}). Finally, we examine how TuneCLIP scales with the amount of 
fine-tuning data and how it performs under varying data quality. As shown in Figure~\ref{fig:combined}(b), performance remains stable even with a fivefold increase (DFN12M to DFN60M). While additional data yields modest gains, improvements are limited as fine-tuning operates on already well-structured pretrained representations.
The DFN datasets are constructed by filtering web image-text pairs using a learned filter, and thus form a relatively less noisy source of self-supervised training data. To examine robustness to raw web-scale (noisy corpora), we fine-tune TuneCLIP on CC12M~\citep{changpinyo2021conceptual}, a web dataset with weaker image-text alignment. As shown in Table~\ref{tab:cc12m_dfn12m}, TuneCLIP consistently achieves positive gains across both datasets, with larger improvements on the less noisy DFN-12M.

\section{Conclusion}
TuneCLIP solves a core problem in model adaptation by showing how to fine-tune a pre-trained model into a superior version with broad, multi-domain improvements. Our two-stage approach combines optimizer recovery with a hinge-based loss to deliver consistent, interpretable gains across tasks. This work goes beyond proposing a new method, opening a concrete and promising direction for self-supervised fine-tuning toward more general-purpose foundation model enhancement.

\section{Limitations and Future Works}
\label{limitations}
One limitation of this work is that we use all data for self-supervised fine-tuning without data selection or filtering.  As a future direction, we consider how to select the most informative data given the knowledge of the pretrained model to accelerate the fine-tuning. Extending our framework beyond CLIP models to other self-supervised architectures (e.g. DINO) is also an interesting direction.

Another limitation is that our framework uses a fixed temperature and a fixed margin, with the margin tuned for the chosen temperature. Making the margin adaptive or learnable is an interesting extension. How to optimize these parameters jointly while preserving the hinge-based loss's ability to mitigate false negatives remains an interesting direction for future work.

\section*{Acknowledgments and Disclosure of Funding}
We thank Texas A\&M University's High Performance Research Computing (HPRC) and the National Center for Supercomputing Applications (NCSA) for providing access to supercomputing resources. A.~Mehta, X.~Chen, X.~Wei, and T.~Yang were partially supported by National Science Foundation Award \#2306572.

\bibliographystyle{unsrt}
\bibliography{ref}

\newpage
\section*{NeurIPS Paper Checklist}

\begin{enumerate}

\item {\bf Claims}
    \item[] Question: Do the main claims made in the abstract and introduction accurately reflect the paper's contributions and scope?
    \item[] Answer: \answerYes{} 
    \item[] Justification: Yes. Our claims regarding TuneCLIP’s improvements over baselines and its superior performance are supported by extensive experiments presented in Section~\ref{sec:exp} and Appendix~\ref{app:fullresults}. Theoretical analysis for the framework is presented in Section~\ref{sec:stageI} and Appendix~\ref{app:proofs}.

\item {\bf Limitations}
    \item[] Question: Does the paper discuss the limitations of the work performed by the authors?
    \item[] Answer: \answerYes{} 
    \item[] Justification: We discuss the limitations of this work in Section~\ref{limitations}. In particular, we note that improved data selection and filtering strategies could further enhance performance, which is currently limited by the availability of noisy data.

\item {\bf Theory assumptions and proofs}
    \item[] Question: For each theoretical result, does the paper provide the full set of assumptions and a complete (and correct) proof?
    \item[] Answer: \answerYes{} 
    \item[] Justification: We provide complete proofs and the associated assumptions for all theorems in Appendix~\ref{app:proofs}.

    \item {\bf Experimental result reproducibility}
    \item[] Question: Does the paper fully disclose all the information needed to reproduce the main experimental results of the paper to the extent that it affects the main claims and/or conclusions of the paper (regardless of whether the code and data are provided or not)?
    \item[] Answer: \answerYes{} 
    \item[] Justification: We present the full TuneCLIP algorithm in Section~\ref{TuneCLIP} (Algorithm~\ref{alg:sogclr-hgcl}). We further detail the architectures, hyperparameters, and experimental setup in Appendix~\ref{app:clip-details} and~\ref{app:hyp} to ensure reproducibility. The code is available at \url{https://github.com/Optimization-AI/TuneCLIP}.

\item {\bf Open access to data and code}
    \item[] Question: Does the paper provide open access to the data and code, with sufficient instructions to faithfully reproduce the main experimental results, as described in supplemental material?
    \item[] Answer: \answerYes{} 
    \item[] Justification: All training datasets and evaluation datasets described in Section~\ref{sec:exp} are publicly available. The code is available at \url{https://github.com/Optimization-AI/TuneCLIP}.

\item {\bf Experimental setting/details}
    \item[] Question: Does the paper specify all the training and test details (e.g., data splits, hyperparameters, how they were chosen, type of optimizer) necessary to understand the results?
    \item[] Answer: \answerYes{} 
    \item[] Justification: All training and evaluation details are given in Section~\ref{sec:exp}, Appendix Sections ~\ref{app:clip-details} and ~\ref{app:hyp}.

\item {\bf Experiment statistical significance}
    \item[] Question: Does the paper report error bars suitably and correctly defined or other appropriate information about the statistical significance of the experiments?
    \item[] Answer: \answerNo{} 
    \item[] Justification: Due to the substantial computational cost of large-scale CLIP training, and following prior works such as~\cite{cherti2023reproducible,wei2024model}, we do not report error bars or confidence intervals for the main experimental results in Section~\ref{sec:exp}. 

\item {\bf Experiments compute resources}
    \item[] Question: For each experiment, does the paper provide sufficient information on the computer resources (type of compute workers, memory, time of execution) needed to reproduce the experiments?
    \item[] Answer: \answerYes{} 
    \item[] Justification: Compute resource details are provided in Appendix~\ref{app:cost-analysis}.
    
\item {\bf Code of ethics}
    \item[] Question: Does the research conducted in the paper conform, in every respect, with the NeurIPS Code of Ethics \url{https://neurips.cc/public/EthicsGuidelines}?
    \item[] Answer: \answerYes{} 
    \item[] Justification: Research conducted follows NeurIPS Code of Ethics.

\item {\bf Broader impacts}
    \item[] Question: Does the paper discuss both potential positive societal impacts and negative societal impacts of the work performed?
    \item[] Answer: \answerNA{} 
    \item[] Justification: There is no societal impact of the work performed.
    
\item {\bf Safeguards}
    \item[] Question: Does the paper describe safeguards that have been put in place for responsible release of data or models that have a high risk for misuse (e.g., pre-trained language models, image generators, or scraped datasets)?
    \item[] Answer: \answerNA{} 
    \item[] Justification: The paper poses no such risks.

\item {\bf Licenses for existing assets}
    \item[] Question: Are the creators or original owners of assets (e.g., code, data, models), used in the paper, properly credited and are the license and terms of use explicitly mentioned and properly respected?
    \item[] Answer: \answerYes{} 
    \item[] Justification: All asset owners are properly credited, including datasets, models, and codebases.

\item {\bf New assets}
    \item[] Question: Are new assets introduced in the paper well documented and is the documentation provided alongside the assets?
    \item[] Answer: \answerNA{} 
    \item[] Justification: This paper does not release new assets at submission time. All datasets used are publicly available, and code will be released after decisions.

\item {\bf Crowdsourcing and research with human subjects}
    \item[] Question: For crowdsourcing experiments and research with human subjects, does the paper include the full text of instructions given to participants and screenshots, if applicable, as well as details about compensation (if any)? 
    \item[] Answer: \answerNA{} 
    \item[] Justification: This paper does not involve crowdsourcing nor research with human subjects.

\item {\bf Institutional review board (IRB) approvals or equivalent for research with human subjects}
    \item[] Question: Does the paper describe potential risks incurred by study participants, whether such risks were disclosed to the subjects, and whether Institutional Review Board (IRB) approvals (or an equivalent approval/review based on the requirements of your country or institution) were obtained?
    \item[] Answer: \answerNA{} 
    \item[] Justification: This paper does not involve crowdsourcing nor research with human subjects.

\item {\bf Declaration of LLM usage}
    \item[] Question: Does the paper describe the usage of LLMs if it is an important, original, or non-standard component of the core methods in this research? Note that if the LLM is used only for writing, editing, or formatting purposes and does \emph{not} impact the core methodology, scientific rigor, or originality of the research, declaration is not required.
    \item[] Answer: \answerNA{} 
    \item[] Justification: LLMs were not used in research.

\end{enumerate}

\appendix
\newpage

\section{Proof for theorems in subsection~\ref{sec:stageI}}\label{app:proofs}
In this part we consider a general FCCO problem:
\[\min_\omega \frac{1}{N}\sum_{i=1}^{N-1}f(g_i(\omega))\]
and the corresponding SOX algorithm~\cite{wang2022finite}. As discussed in preliminary, by specifying $N = 2n$, $f(\cdot) = \log(\epsilon + \cdot)$, $g_i(\cdot) = g_i(\cdot, \D) = \Phi_1(\cdot, i, \D)$ if $i\leq n$ otherwise $\Phi_2(\cdot, i-n, \D)$, we recover the GCL loss and the SogCLR algorithm. We also set $|\B_1| = |\B_2| = B$ in SOX when presenting its convergence analysis to simplify notation.
Before starting our proofs, we make the following standard, commonly used assumptions as in~\cite{wang2022finite} under which theorems in subsection~\ref{sec:stageI} hold:
\begin{assumption}\label{asm:func}
    We assume that:\\
    $\bullet$  $f(\cdot)$ and $\nabla f(\cdot)$ are $C_f$ and $L_f$-Lipschitz continuous, respectively.\\
    $\bullet$  $g_i(\cdot)$ and $\nabla g_i(\cdot)$ are $C_g$ and $L_g$-Lipschitz continuous, respectively.
\end{assumption}
\begin{assumption}\label{asm:oracle}
There exist constants $\sigma_0\geq0$ and $\sigma_1\geq0$ such that the following statements hold for $g_i(\omega)$ and $g_i(\omega,\xi_i)$ for $i=1,\dots,N$ for any $\omega\in \mathbb{R}^d$ 
   $\mathbb{E}\|g_i(\omega, \xi_i) - g_i(\omega)\|^2\leq \sigma_0^2$, $\mathbb{E}\normsq{\nabla g_i(\omega, \xi_i) - \nabla g_i(\omega)}\leq \sigma_1^2$.
\end{assumption}

\subsection{Technical Lemma}
The following descent lemma is commonly used in the analysis of momentum methods for non-convex optimization(see e.g.~\cite{wang2022finite, chen2026stochastic}).
\begin{lemma}
	Consider a sequence $\omega_{t+1} = \omega_t - \eta m_{t+1}$ and the $L_F$-smooth function $F$ and the step size $\eta L_F\leq 1/2$. 
	\begin{align}\label{eq:nonconvex_starter}
		F(\omega_{t+1}) & \leq F(\omega_t) + \frac{\eta}{2}M_t - \frac{\eta}{2}\norm{\nabla F(\omega_t)}^2 - \frac{\eta}{4}\norm{m_{t+1}}^2,
	\end{align}	
	where $M_t \coloneqq \norm{m_{t+1} - \nabla F(\omega_t)}^2$.
	\label{lem:nonconvex_starter}
\end{lemma}	

We build a recursion for the gradient variance $M_t$ by proving the following lemma. \begin{lemma}\label{lem:grad_recursion}
	If $\beta\leq \frac{2}{7}$, the gradient variance $M_t$ can be bounded as
	\begin{align}\label{eq:grad_recursion}
        \E\left[M_{t+1}\right] & \leq (1-\beta)\E\left[M_t\right] + \frac{2L_F^2\eta^2}{\beta}\E\left[\norm{m_{t+1}}^2\right] +  \frac{2\beta^2 C_f^2(\sigma_1^2+C_g^2)}{B}+ 5\beta L_f^2C_1^2 \E\left[U_{t+1}\right]
	\end{align}
	where $U_{t} = \frac{1}{N}\norm{u_{t+1} -g(\omega_t;\D)}^2$, $u_t = [u_1^{(t)},\dotsc,u_N^{(t)}]^\top$, $g(\omega_t;\D) = [g_1(\omega_t;\D_1),\dotsc, g_N(\omega_t;\D_N)]^\top$ and $C_1^2 = C_g^2 + \frac{\sigma_1^2}{B}$. Also note that we follow the tradition usage of $\beta$ in~\cite{wang2022finite} so $\beta = 1 - \beta_1$ where $\beta_1$ is used in algorithm~\ref{alg:osr}.
\end{lemma}
\textbf{Remark}: We point out that the lemma is similar to lemma 9 in~\cite{wang2022finite} without a term corresponding to $\normsq{u_{t+1} - u_{t}}$, the gap is caused by the different usage of $u$ when constructing the overall gradient estimator $G(\omega_{t-1}, \B_t)$: instead of using old $u_{t-1}$ as in SOX we use a newer version $u_t$ in this paper. This would require sampling one more iid minibatch per iteration to derive a bound as shown in the lemma. However in practice we typically sample only a single minibatch.
\begin{lemma}\label{lem:fval_recursion}
 	If $\gamma\leq 1/5$, function value variance $U_t$ can be bounded as
 	\begin{align}\label{eq:fval_recursion}
 		\E\left[U_{t+1}\right] &\leq \left(1-\frac{\gamma B}{4N}\right)\E\left[U_t\right] + \frac{5N\eta^2C_g^2}{\gamma B}\E\left[\norm{m_{t+1}}^2\right] + \frac{2\gamma^2\sigma_0^2}{N}
 	\end{align}	
\end{lemma}	
\textbf{Remark:} We directly drop the negative term in lemma 2 in~\cite{wang2022finite}.
\subsection{Proof of theorem}
\begin{proof}[Proof of Theorem~\ref{thm:main_thm_sox}] The proof is almost the same as theorem 3 in~\cite{wang2022finite} so we only make necessary clarifications.
Summing \eqref{eq:nonconvex_starter}, $\frac{\eta}{\beta}\times$\eqref{eq:grad_recursion}, and $\frac{20L_f^2C_1^2N\eta}{\gamma B}\times$\eqref{eq:fval_recursion} leads to
\begin{align*}
	& \E\left[F(\omega_{t+1})- F^* + \frac{\eta}{\beta}M_{t+1} + \frac{20L_f^2C_1^2N\eta}{\gamma B}\left(1-\frac{\gamma B}{4N}\right)U_{t+1}\right]\\
	& \leq  \E\left[F(\omega_t)- F^* + \frac{\eta}{\beta}\left(1-\frac{\beta }{2}\right)M_t + \frac{20L_f^2C_1^2N\eta}{\gamma B}\left(1-\frac{\gamma B}{4N}\right)U_t\right] -\frac{\eta}{2}\E\left[\norm{\nabla F(\omega_t)}^2\right]\\
	& \quad\quad\quad  - \eta\left(\frac{1}{4} - \frac{2L_F^2\eta^2}{\beta^2} - \frac{100L_f^2 N^2C_1^2\eta^2 C_g^2}{\gamma^2}\right) \E\left[\norm{m_{t+1}}^2\right] + \frac{2\beta\eta C_f^2(\sigma_1^2 + C_g^2)}{B} + \frac{40\eta\gamma L_f^2 C_1^2\sigma_0^2}{B}.
\end{align*}
Set $\beta=\min\{\frac{B\epsilon^2}{12C_f^2(\sigma_1^2+C_g^2)}, \frac{2}{7}\}$, $\gamma = \min\left\{\frac{B\epsilon^2}{240L_f^2C_1^2\sigma_0^2}, \frac{1}{5}\right\}$, and $\eta = \min\left\{\frac{\beta}{4 L_F}, \frac{\gamma B}{30L_f N C_1 C_g}\right\}$. Define the Lyapunov function as $\Gamma_t\coloneqq F(\omega_t) - F^* + \frac{\eta}{\beta}M_t + \frac{20L_f^2 C_1^2N}{B}\frac{\eta}{\gamma}\left(1-\frac{\gamma B}{4N}\right)U_t$. Then,
\begin{align}\label{eq:sox_grad_norm_bound}
	\frac{1}{T}\sum_{t=0}^{T-1} \E\left[\norm{\nabla F(\omega_t)}^2\right] \leq \frac{2\Gamma_0}{\eta T} + \frac{4\beta C_f^2(\sigma_1^2 + C_g^2)}{B} + \frac{80 \gamma L_f^2 C_1^2\sigma_0^2}{B},
\end{align}
discarding the non-dominant terms and unimportant constants, to guarantee $\frac{1}{T}\sum_{t=0}^{T-1} \E\left[\norm{\nabla F(\omega_t)}^2\right] \leq \epsilon^2$ we need at most
\begin{align*}
    T = O\left(\frac{n\Gamma_0}{B^2\epsilon^4}\right) = O\left(\frac{n}{B^2\epsilon^4}(\Delta_0 + \frac{B}{n}M_0 + U_0)\right)
\end{align*}
iterations, which leads to conclusion by noting that $U_{0} = \frac{1}{N}\norm{u_0 -g(\boldsymbol{\omega_0};\D)}^2 = \frac{1}{2n}\normsq{u_x^{(0)} - \Phi_1(\boldsymbol{\omega_0}, \D)} + \frac{1}{2n}\normsq{u_z^{(0)} - \Phi_2(\boldsymbol{\omega_0}, \D)}$.\\
\end{proof}
\begin{proof}[Proof of Theorem~\ref{thm:stageI}]
    Note that algorithm~\ref{alg:osr} is essentially SOX~\cite{wang2022finite} without updating the model parameters $\omega$(i.e. learning rate $\eta=0$), we can still leverage lemma~\ref{lem:grad_recursion} and~\ref{lem:fval_recursion} by plugging $\eta=0$ into them and have the following bound:
    \begin{align}
        \E\left[M_{t+1}\right] & \leq (1-\beta)\E\left[M_t\right]  +  \frac{2\beta^2 C_f^2(\sigma_1^2+C_g^2)}{B}+ 5\beta L_f^2C_1^2 \E\left[U_{t+1}\right]\label{eq:grad_recur_OSR}\\
        \E\left[U_{t+1}\right] &\leq \left(1-\frac{\gamma B}{4N}\right)\E\left[U_t\right] + \frac{2\gamma^2\sigma_0^2}{N}\label{eq:fval_recur_OSR}
    \end{align}
    Note that now we are not updating $\omega$ so $M_t = \normsq{m_{t+1} - \nabla_\omega \L_{\mathrm{GCL}}(\boldsymbol{\omega_0})}$, $U_t = \frac{1}{N}\norm{u_{t+1} - g(\boldsymbol{\omega_0};\D)}^2$. Rearranging terms and divide both side for~\eqref{eq:grad_recur_OSR}, ~\eqref{eq:fval_recur_OSR} by $\beta$ and $\frac{\gamma B}{4N}$, respectively, then we have:
    \begin{align}
        \E\left[M_t\right] & \leq \frac{1}{\beta}\E\left[M_t - M_{t+1}\right] + \frac{2\beta C_f^2(\sigma_1^2+C_g^2)}{B}+ 5 L_f^2C_1^2 \E\left[U_{t+1}\right]\\
        \E\left[U_t\right] &\leq \frac{4N}{\gamma B}\E\left[U_t - U_{t+1}\right] + \frac{8\gamma\sigma_0^2}{B}\label{eq:u_bound}
    \end{align}
    combining the above two inequalities we have
    \begin{align}
        \E\left[M_t\right] & \leq \frac{1}{\beta}\E\left[M_t - M_{t+1}\right]  +  \frac{2\beta C_f^2(\sigma_1^2+C_g^2)}{B} + 5 L_f^2C_1^2 \left((\frac{4N}{\gamma B}-1)\E\left[U_t - U_{t+1}\right] + \frac{8\gamma\sigma_0^2}{B}\right)\nonumber\\
        &\leq \E\left[\Psi_t - \Psi_{t+1}\right] + \frac{2\beta C_f^2(\sigma_1^2+C_g^2)}{B} + \frac{40\gamma\sigma_0^2L_f^2C_1^2}{B}
    \end{align}
    where $\Psi_t = \frac{1}{\beta}M_t + \frac{20NL_f^2C_1^2}{\gamma B}\left(1 - \frac{\gamma B}{4N}\right)U_t$. Sum over $t = 0,1,\cdots, T-1$ and divide both side by $T$ then we have
    \begin{align}
        \frac{1}{T}\sum_{t=0}^{T-1}\E[M_t] \leq \frac{\Psi_0}{T} + \frac{2\beta C_f^2(\sigma_1^2+C_g^2)}{B} + \frac{40\gamma\sigma_0^2L_f^2C_1^2}{B}
    \end{align}
    Convergence of $U_t$ can be easily derived from~\eqref{eq:u_bound} by summing over $t = 0,1,\cdots, T-1$ and divide both side by $T$:
    \begin{align}
        \frac{1}{T}\sum_{t=0}^{T-1}\E[U_t] \leq \frac{4NU_0}{\gamma BT} + \frac{8\gamma\sigma_0^2}{B}
    \end{align}
   By setting $\beta = O(\sqrt{\frac{N}{T}}), \gamma = O(\sqrt{\frac{N}{T}})$ and omitting unimportant constants, we have
    \begin{align}
        \E_{\hat{t}}\left[\E[M_{\hat{t}}]\right] = \frac{1}{T}\sum_{t=0}^{T-1}\E[M_t] \leq O\left(\frac{M_0}{\sqrt{NT}} + \frac{U_0}{\sqrt{BE}} + \frac{1}{\sqrt{BE}}\right)\\
        \E_{\hat{t}}\left[\E[U_{\hat{t}}]\right] = \frac{1}{T}\sum_{t=0}^{T-1}\E[U_t] \leq O\left(\frac{U_0}{\sqrt{BE}} + \frac{1}{\sqrt{BE}}\right)
    \end{align}
    which directly leads to the conclusion by noting that $N=2n, T = \frac{nE}{B}$ and $U_0 = U_{x,0} + U_{z,0}$.
\end{proof}
\subsection{Convergence Analysis of MBCL}\label{sec:analysis_MBCL}
We only highlight the difference to the analysis in the previous section by noting that MBCL is a special case of GCL with $\gamma=1$. Note that in this case we have $u_{i,x}^{(t)}=\Phi_1(\omega_{t-1},i,\mathcal{B}_t)$ and $u_{i,z}^{(t)}=\Phi_2(\omega_{t-1},i,\mathcal{B}_t)$, which leads to a naive bound:
\begin{align}
    U_t\leq \frac{\sigma_0^2}{B}
\end{align}
Plug it in eqn.~\eqref{eq:grad_recursion}. Then summing \eqref{eq:nonconvex_starter}, $\frac{\eta}{\beta}\times$\eqref{eq:grad_recursion} leads to
\begin{align*}
	& \E\left[F(\omega_{t+1})- F^* + \frac{\eta}{\beta}M_{t+1}\right]\\
	& \leq  \E\left[F(\omega_t)- F^* + \frac{\eta}{\beta}\left(1-\frac{\beta }{2}\right)M_t\right] -\frac{\eta}{2}\E\left[\norm{\nabla F(\omega_t)}^2\right]\\
	& \quad\quad\quad  - \eta\left(\frac{1}{4} - \frac{2L_F^2\eta^2}{\beta^2}\right) \E\left[\norm{m_{t+1}}^2\right] + \frac{2\beta\eta C_f^2(\sigma_1^2 + C_g^2)}{B} + \frac{5\eta L_f^2 C_1^2\sigma_0^2}{B}.
\end{align*}
Set $B=\Theta(\frac{15L_f^2C_1^2\sigma_0^2}{\epsilon^2})$, $\beta=\min\{\frac{B\epsilon^2}{12C_f^2(\sigma_1^2+C_g^2)}, \frac{2}{7}\}$ and $\eta = \Theta(\beta)$. Define the Lyapunov function as $\Gamma_t\coloneqq F(\omega_t) - F^* + \frac{\eta}{\beta}M_t$. Then,
\begin{align}\label{eq:sox_grad_norm_bound}
	\frac{1}{T}\sum_{t=0}^{T-1} \E\left[\norm{\nabla F(\omega_t)}^2\right] \leq \frac{2\Gamma_0}{\eta T} + \frac{4\beta C_f^2(\sigma_1^2 + C_g^2)}{B} + \frac{10L_f^2 C_1^2\sigma_0^2}{B},
\end{align}
discarding the non-dominant terms and unimportant constants, to guarantee $\frac{1}{T}\sum_{t=0}^{T-1} \E\left[\norm{\nabla F(\omega_t)}^2\right] \leq \epsilon^2$, the sample complexity is 
\begin{align*}
    T*B = O\left(\frac{\Gamma_0}{\epsilon^4}\right) = O\left(\frac{\Delta_0 + M_0}{\epsilon^4}\right)
\end{align*}
iterations, which concludes the proof.\\
\textbf{Remark}: The above analysis shows that the iteration complexity of MBCL also depends on the initial estimation error of $m_0$.

\newpage

\section{Additional details on CLIP models}
\label{app:clip-details}

The CLIP models (ViT-B/32 and ViT-B/16) use an embedding dimension of 512 for contrastive learning. In contrast, SigLIP employs a larger embedding dimension of 768.
Tables~\ref{tab:vision_tower} and \ref{tab:text_tower} summarize the configurations of the vision and text encoders, respectively. Table~\ref{tab:model_specs} further reports the overall model specifications, including parameter counts and developers.

\begin{table}[H]
\centering
\caption{Vision tower configurations of CLIP models.}
\begin{tabular}{lcccc}
\toprule
\textbf{Model} & \textbf{Image Size} & \textbf{Layers} & \textbf{Width} & \textbf{Patch Size} \\
\midrule
CLIP ViT-B/32      & 224 & 12 & 768 & 32 \\
CLIP ViT-B/16      & 224 & 12 & 768 & 16 \\
SigLIP ViT-B/16    & 224 & 12 & 768 & 16 \\
\bottomrule
\end{tabular}
\label{tab:vision_tower}
\end{table}

\begin{table}[H]
\centering
\caption{Text tower configurations of CLIP models.}
\begin{tabular}{lccccc}
\toprule
\textbf{Model} & \textbf{Context Length} & \textbf{Vocab Size} & \textbf{Width} & \textbf{Heads} & \textbf{Layers} \\
\midrule
CLIP ViT-B/32      & 77 & 49408 & 512 & 8  & 12 \\
CLIP ViT-B/16      & 77 & 49408 & 512 & 8  & 12 \\
SigLIP ViT-B/16    & 64 & 32000 & 768 & 12 & 12 \\
\bottomrule
\end{tabular}
\label{tab:text_tower}
\end{table}

\begin{table}[ht]
\centering
\caption{Model specifications of different CLIP variants.}
\begin{tabular}{lcccc}
\toprule
\textbf{Model} & \textbf{Vision Encoder} & \textbf{Text Encoder} & \textbf{Parameters (M)} & \textbf{Developer} \\
\midrule
CLIP ViT-B/32            & ViT & Transformer & 151.28 & OpenAI~\cite{radford2021learning} \\
CLIP ViT-B/16            & ViT & Transformer & 149.62 & OpenAI~\cite{radford2021learning} \\
CLIP ViT-B/32            & ViT & Transformer & 151.28 & LAION~\cite{schuhmann2022laion}  \\
SigLIP ViT-B/16     & ViT & Transformer & 203.16 & Google~\cite{zhai2023sigmoid} \\
\bottomrule
\end{tabular}
\label{tab:model_specs}
\end{table}

\begin{table}[h]
\centering
\caption{Key hyperparameters used for training the models.}
\begin{tabular}{ll}
\toprule
\textbf{Hyperparameter} & \textbf{Value} \\
\midrule
Image size         & 224x224 \\
Learning rate (lr) & 1e-5 \\
Optimizer          & AdamW \\
Beta1, Beta2       & 0.9, 0.98 \\
Weight decay (wd)  & 0.02 \\
Scheduler          & Cosine \\
Precision          & AMP (mixed precision) \\
Margin             & 0.1 \\
Margin smoothing   & 2.0 \\
Gamma              & 0.9 \\
Gamma schedule     & Cosine (decay every 4 epochs) \\
World size         & 8 (DDP) \\
Workers            & 6 \\
\bottomrule
\end{tabular}
\label{tab:fastclip_hparams}
\end{table}

\section{Hyperparameter details}
\label{app:hyp}

A brief summary of the key hyperparameters is provided in Table~\ref{tab:fastclip_hparams}. Owing to the availability of 40GB A100s and 80GB H100s, we conducted all experiments to an image resolution of $224 \times 224$. Across models, the best-performing base learning rate was consistently around $1\times10^{-5}$. We used cosine scheduling on the learning rate in the second stage of fine-tuning. We adopted AdamW as the optimizer with momentum parameters $(\beta_1, \beta_2) = (0.9, 0.98)$, and a weight decay of $0.02$ to improve generalization. A cosine learning rate scheduler was used to provide smooth decay to $\gamma$, following a cosine schedule until the 4\textsuperscript{th} epoch and fixed to $0.9$ thereafter.
 We also applied mixed-precision training (\textit{AMP}) to balance performance and efficiency. Each experiment used a world size of 8 for DDP and 6 data-loading workers per GPU to optimize throughput.

\subsection{\textcolor{black}{Ablation on different learning rates}}
\begin{table}[h]
\centering
\caption{Learning rate sweep on TuneCLIP}
\begin{tabular}{lcc}
\toprule
\textcolor{black}{\textbf{Learning Rate}} &
\textcolor{black}{\textbf{ImageNet-1k (\%)}} &
\textcolor{black}{\textbf{MS COCO (Avg R@1) (\%)}} \\
\midrule
\textcolor{black}{1e-4}   & \textcolor{black}{69.66} & \textcolor{black}{48.98} \\
\textcolor{black}{1e-5}   & \textcolor{black}{70.57} & \textcolor{black}{50.11} \\
\textcolor{black}{5e-6}   & \textcolor{black}{70.23} & \textcolor{black}{49.30} \\
\bottomrule
\end{tabular}
\label{tab:lr_sweep}
\end{table}

\textcolor{black}{After sweeping learning rates across $\{10^{-4}, 10^{-5}, 10^{-6}\}$, we observe that performance drops slightly above $10^{-4}$, while learning rates in the range of $10^{-6}$ to $10^{-5}$ remain comparably strong. For example, Table~\ref{tab:lr_sweep} shows that TuneCLIP achieves stable performance around $1\text{e--}5$ and $5\text{e--}6$.}

\subsection{\textcolor{black}{Ablation on margin $m$ used with HGCL}}
\label{app:m_hyp}

\begin{figure}[h]
    \centering
    \includegraphics[width=0.4\linewidth]{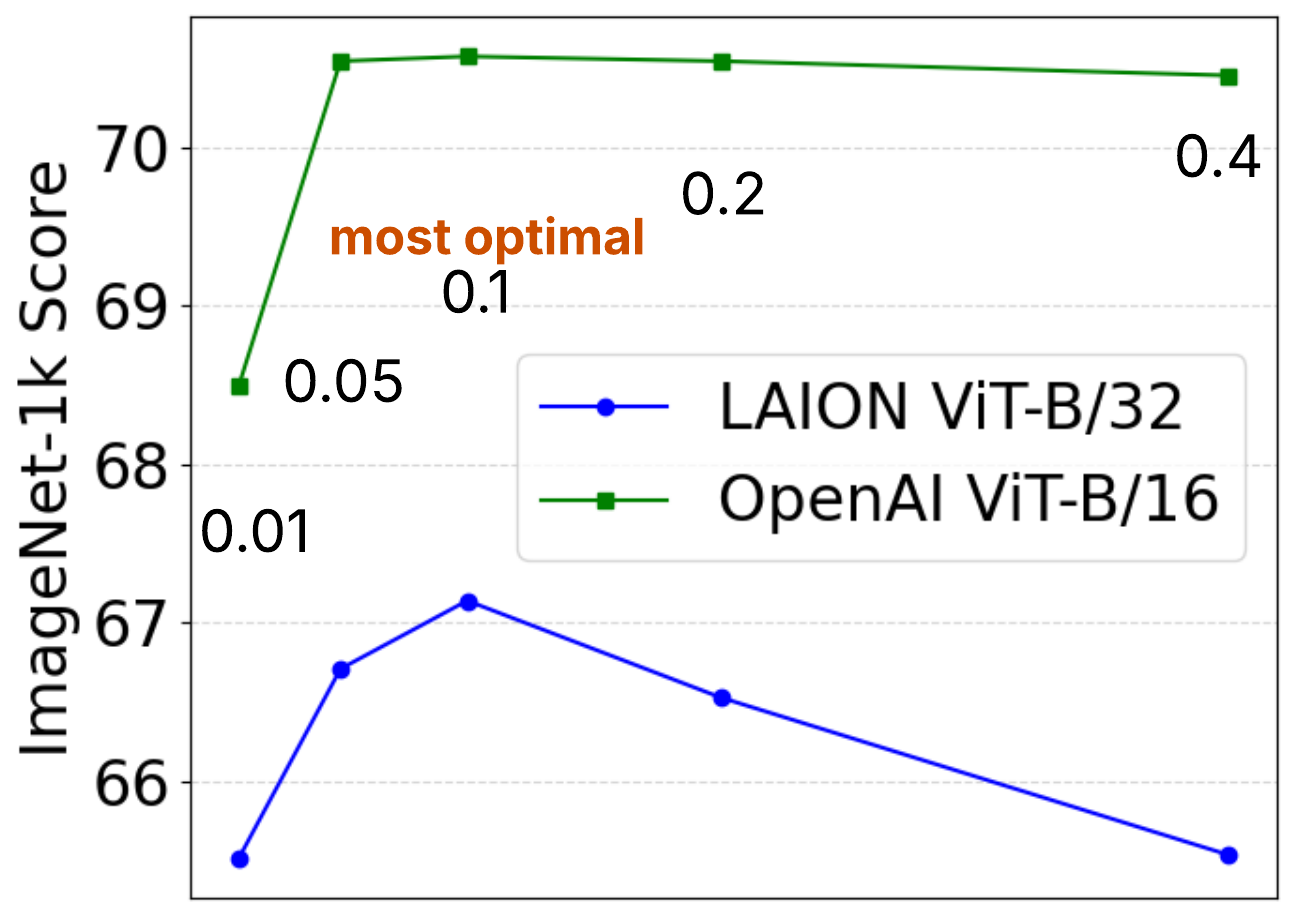}
    \caption{Sensitivity of margin ($m$).}
    \label{fig:margin_comp}
\end{figure}

To study the impact of the margin $m$, we consider two different CLIP
architectures, LAION ViT-B/32 and OpenAI ViT-B/16, and sweep over a
representative set of values $m \in \{0.01, 0.05, 0.10, 0.20, 0.40\}$. 
The resulting ImageNet-1k accuracies are plotted in
Figure~\ref{fig:margin_comp}. Using this as a representative metric, we observe that margins around $m = 0.1$ consistently perform well across different model scales. 

\section{Baseline details}

\begin{table}[ht]
\centering
\caption{Training configurations for compared methods.}
\begin{tabular}{lll}
\toprule
\textbf{Method} & \textbf{Loss} & \textbf{Optimization Strategy} \\
\midrule
FastCLIP \cite{wei2024fastclip} & GCL & SogCLR + AdamW \\
OpenCLIP \cite{cherti2023reproducible} & MBCL & AdamW \\
CLIP-Refine \cite{yamaguchi2025post} & RaFA+HyCD\cite{yamaguchi2025post} & AdamW \\
\rowcolor{yellow!15}
TuneCLIP (ours) & HGCL (ours) & OSR (ours) + SogCLR + AdamW \\
\bottomrule
\end{tabular}
\label{tab:algo_loss_opt}
\end{table}

Table \ref{tab:algo_loss_opt} summarizes the training setups of the algorithms used in our comparison. 
FastCLIP \citep{wei2024fastclip} employs the standard Global Contrastive Loss (GCL) with the SogCLR optimization algorithm and AdamW. OpenCLIP \citep{cherti2023reproducible} relies on a minibatch contrastive loss (MBCL) combined with AdamW. CLIP-Refine~\cite{yamaguchi2025post} uses random feature alignment (RaFA) and
hybrid contrastive-distillation (HyCD). Our TuneCLIP introduces the proposed Hinged Global Contrastive Loss (HGCL) loss 
and leverages Optimizer Statistics Recovery (OSR) alongside SogCLR and AdamW.

\newpage
\section{Impact of Hinged Global Contrastive Loss}  
\label{app:benefits_hgcl}

The controlled study in Figures~\ref{fig:main_results_row2}, \ref{fig:main_results_row1} and \ref{fig:supervised_testing} is set up as follows: supervised fine-tuning is performed on Flickr30k, while SSFT uses DFN-12M. Train Retrieval trends are computed from 15,000 random DFN samples (SSFT) and 1,000 Flickr30k samples (SFT). Both models are evaluated on the Flickr1k test set.

\setlength{\fboxrule}{1pt} 
\setlength{\fboxsep}{1pt}  

\begin{figure}[h]
  \centering
  \fbox{%
    \colorbox{red!20}{%
      \setlength{\fboxsep}{1pt}%
      \begin{subfigure}{0.30\textwidth}
        \centering
        \includegraphics[width=\linewidth]{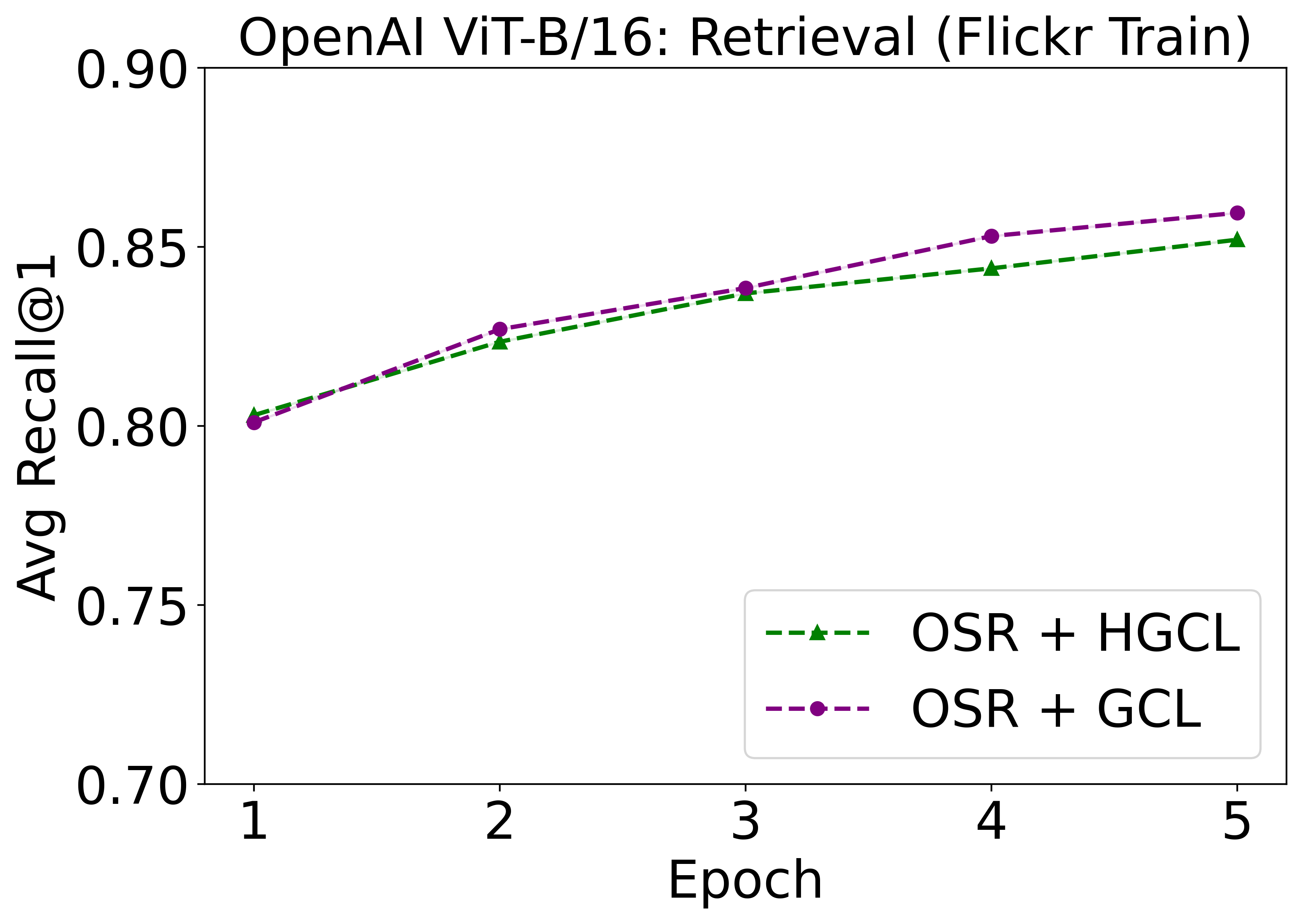}
        \caption{Supervised finetuning}
        \label{fig:flickr2}
      \end{subfigure}%
    }%
  }\hfill
  \fbox{%
    \colorbox{blue!10}{%
      \setlength{\fboxsep}{2pt}%
      \begin{subfigure}{0.65\textwidth}
        \centering
        \includegraphics[width=0.46\linewidth]{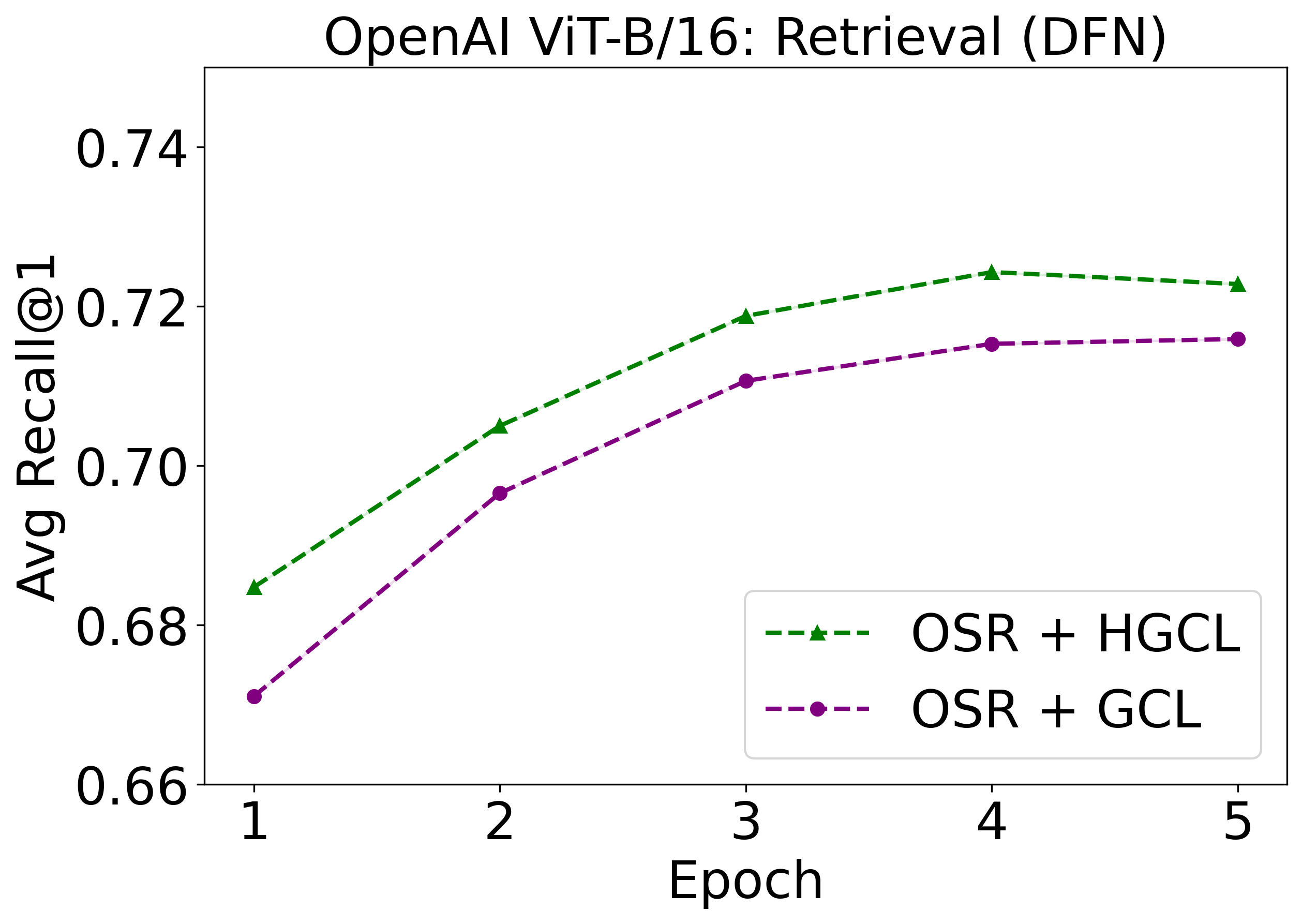}\hspace{8pt}%
        \includegraphics[width=0.46\linewidth]{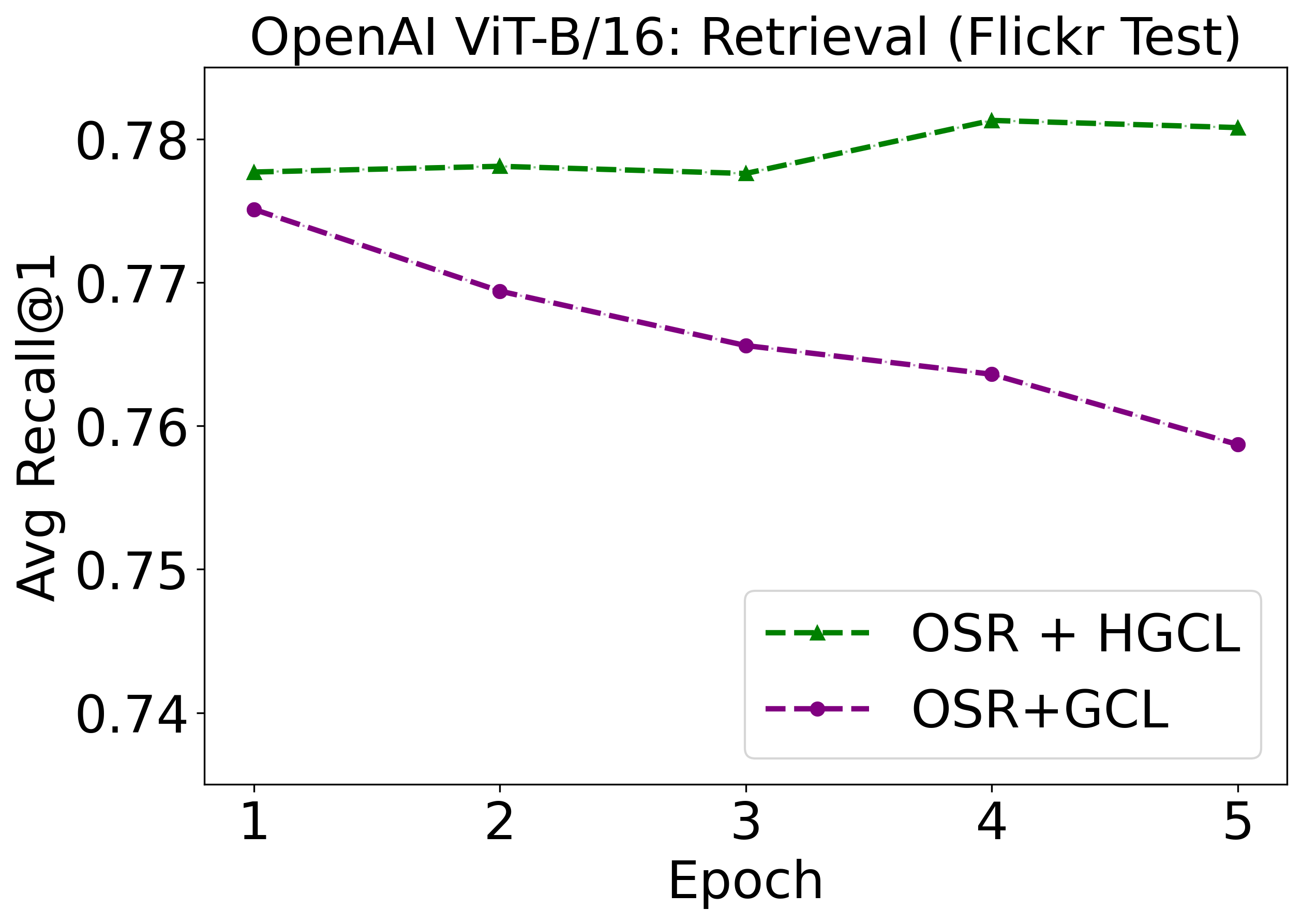}
        \caption{Self-Supervised finetuning}
        \label{fig:ssft_combined-2}
      \end{subfigure}%
    }%
  }
\caption{
Similar to Figure~\ref{fig:main_results_row2}, supervised fine-tuning (red) shows stronger performance with OSR+GCL than with OSR+HGCL (TuneCLIP), since true negative labels justify separating negatives. 
By contrast, in self-supervised fine-tuning (blue), the absence of such labels makes OSR+HGCL more effective.}

  \label{fig:main_results_row1}
\end{figure}

\begin{figure*}[h]
    \centering
    \begin{subfigure}[t]{0.40\textwidth}
        \centering
        \includegraphics[width=\linewidth]{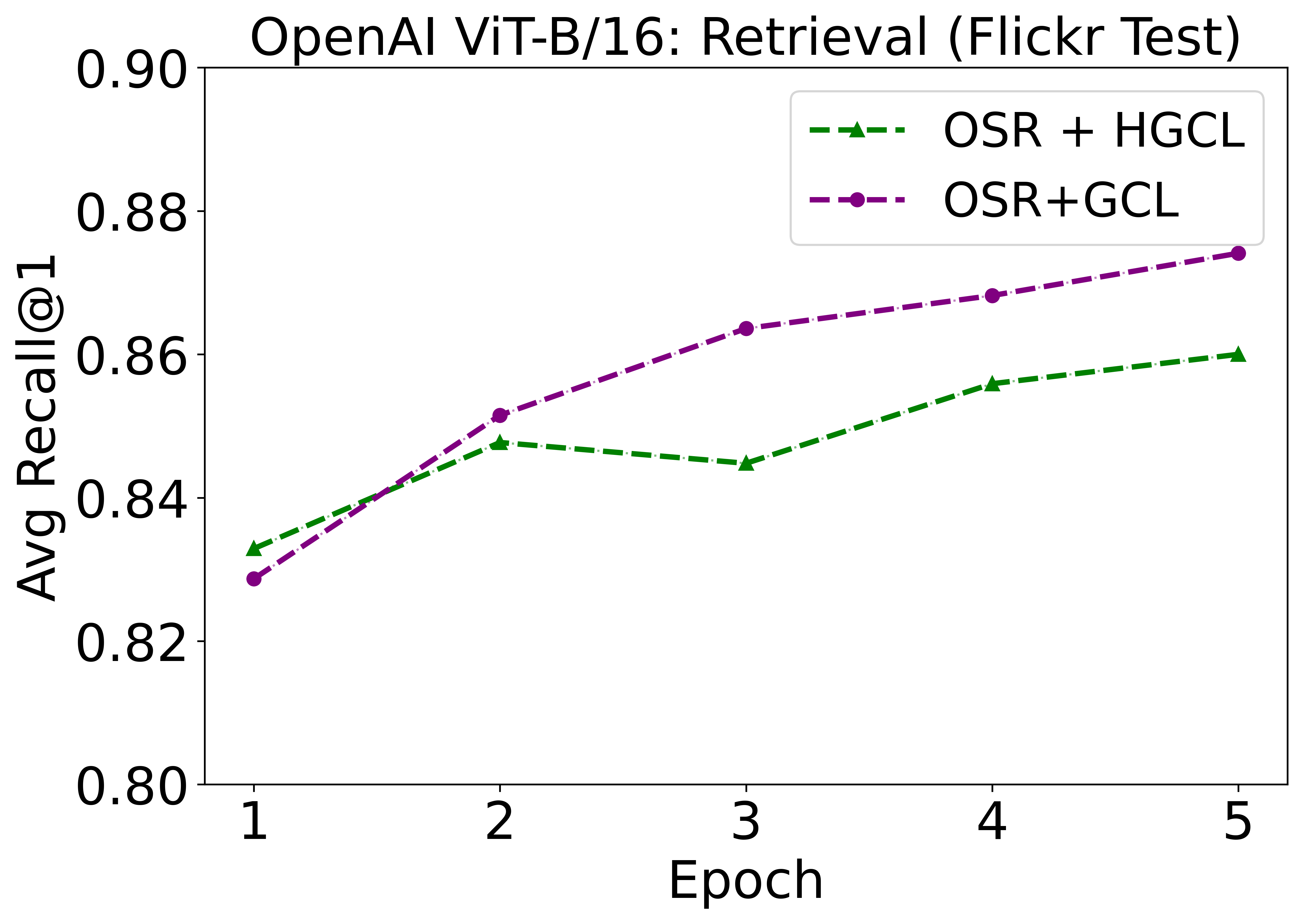}
        \caption{}
        \label{fig:sft-val1}
    \end{subfigure}
    \begin{subfigure}[t]{0.40\textwidth}
        \centering
        \includegraphics[width=\linewidth]{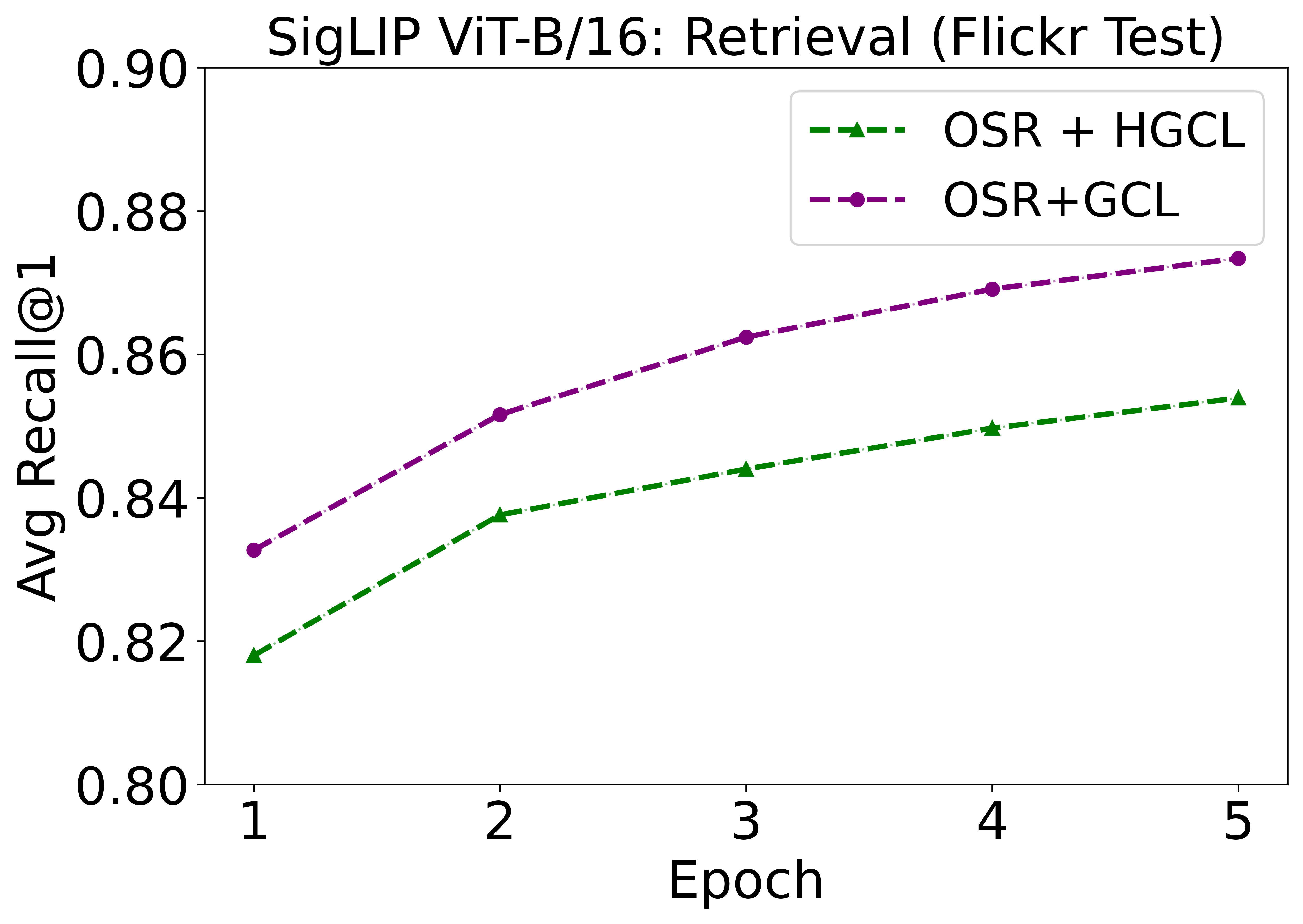}
        \caption{}
        \label{fig:sft-val2}
    \end{subfigure}
\caption{
Retrieval trends on the Flickr1k test set under supervised fine-tuning. 
In the supervised setting, explicit labels guide the separation of positives from negatives, making OSR+GCL outperform OSR+HGCL. 
By contrast, as shown in Figures~\ref{fig:ssft_combined} and \ref{fig:ssft_combined-2}, the absence of supervision and margin regularization in SSFT reverses this trend, with OSR+HGCL (TuneCLIP) achieving superior retrieval performance. 
}

\label{fig:supervised_testing}
\end{figure*}

\newpage
\section{Fine-tuning performance on ImageNet-1k}
\label{app:in1k}

\begin{figure}[h]
    \centering
    
    \begin{subfigure}[t]{0.95\textwidth}
        \centering
        \begin{subfigure}[t]{0.45\textwidth}
            \centering
            \includegraphics[width=\linewidth]{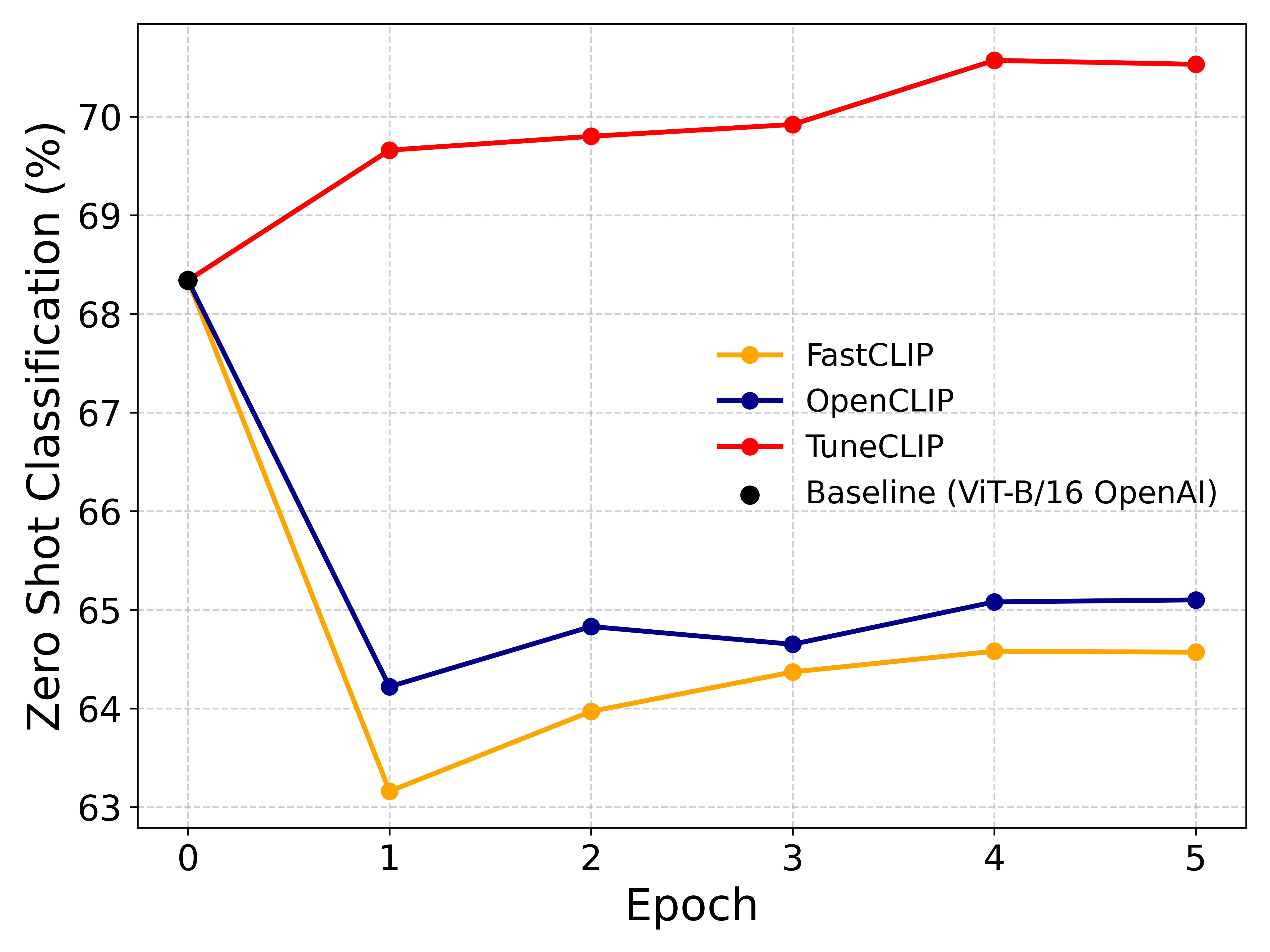}
            \label{fig:in1k-vitb16openai}
        \end{subfigure}
        \hfill
        \begin{subfigure}[t]{0.45\textwidth}
            \centering
            \includegraphics[width=\linewidth]{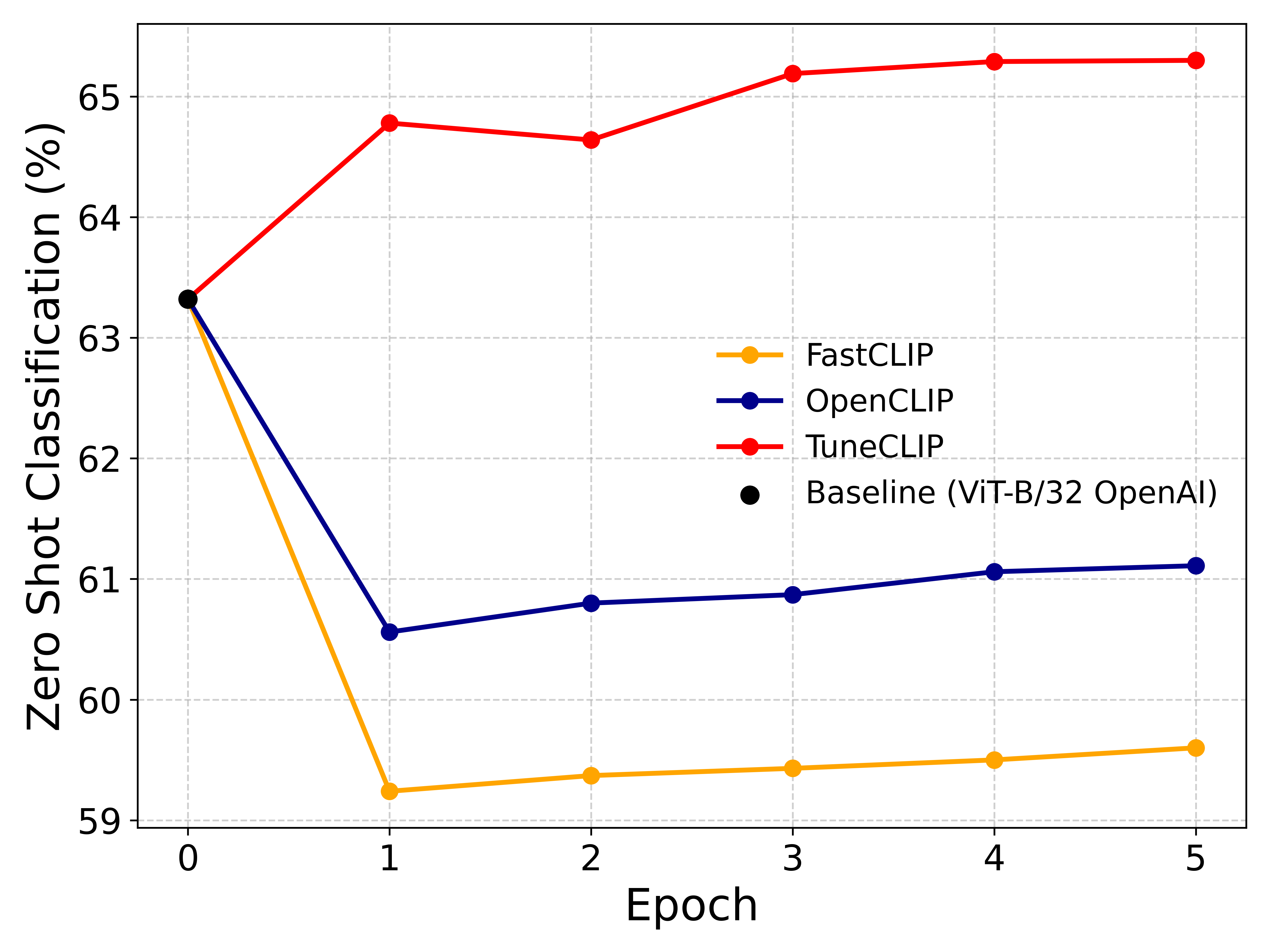}
            \label{fig:in1k-vitb32openai}
        \end{subfigure}
        \caption{OpenAI models.}
        \label{fig:in1k-openai}
    \end{subfigure}
    
    \vspace{0.8em}
    
    \begin{subfigure}[t]{0.95\textwidth}
        \centering
        \begin{subfigure}[t]{0.45\textwidth}
            \centering
            \includegraphics[width=\linewidth]{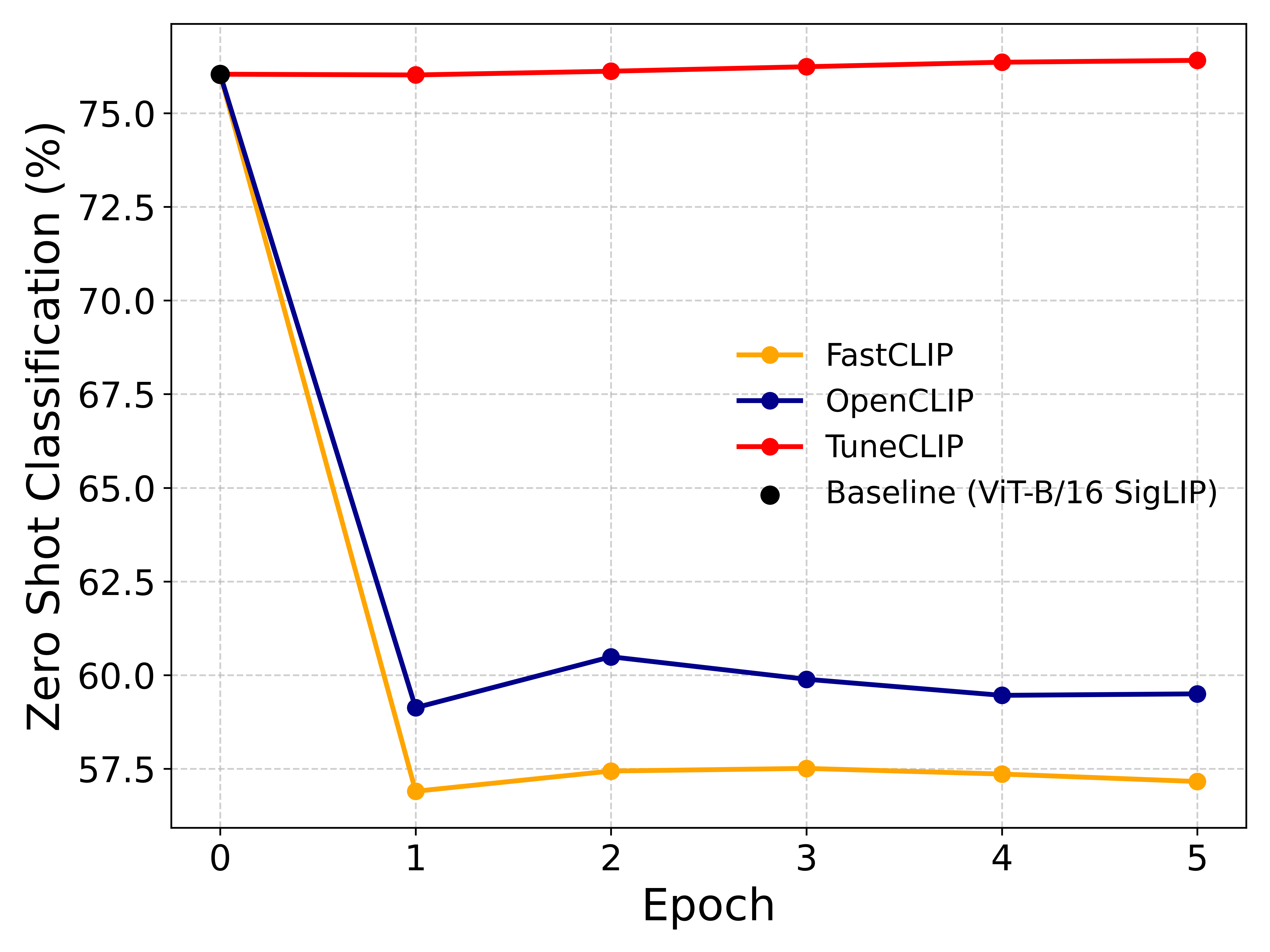}
            \label{fig:in1k-vitb16siglip}
        \end{subfigure}
        \hfill
        \begin{subfigure}[t]{0.45\textwidth}
            \centering
            \includegraphics[width=\linewidth]{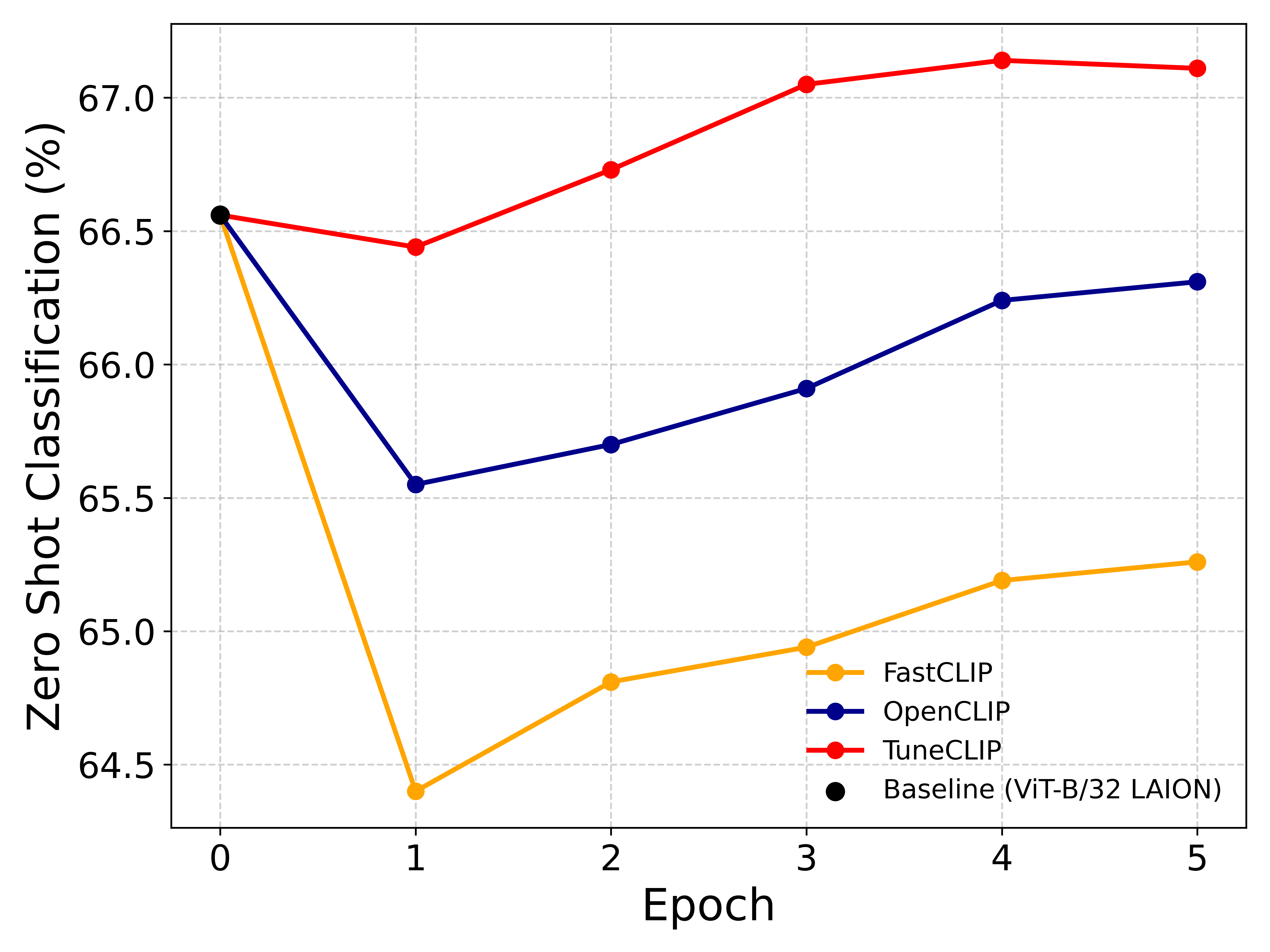}
            \label{fig:in1k-vitb32laion}
        \end{subfigure}
        \caption{SigLIP and LAION models.}
        \label{fig:in1k-others}
    \end{subfigure}

    \caption{Zero-shot classification scores  (ImageNet-1k) over 5 epochs of fine-tuning for all CLIPs.}
    \label{fig:imagenet_finetuning}
\end{figure}

Across all four models, the baselines exhibit degraded initialization and fail to recover in early epochs. In contrast, TuneCLIP consistently starts stronger and outperforms them throughout fine-tuning (Figure~\ref{fig:imagenet_finetuning}).

\begin{figure}[ht]
    \centering
    \includegraphics[width=0.50\linewidth]{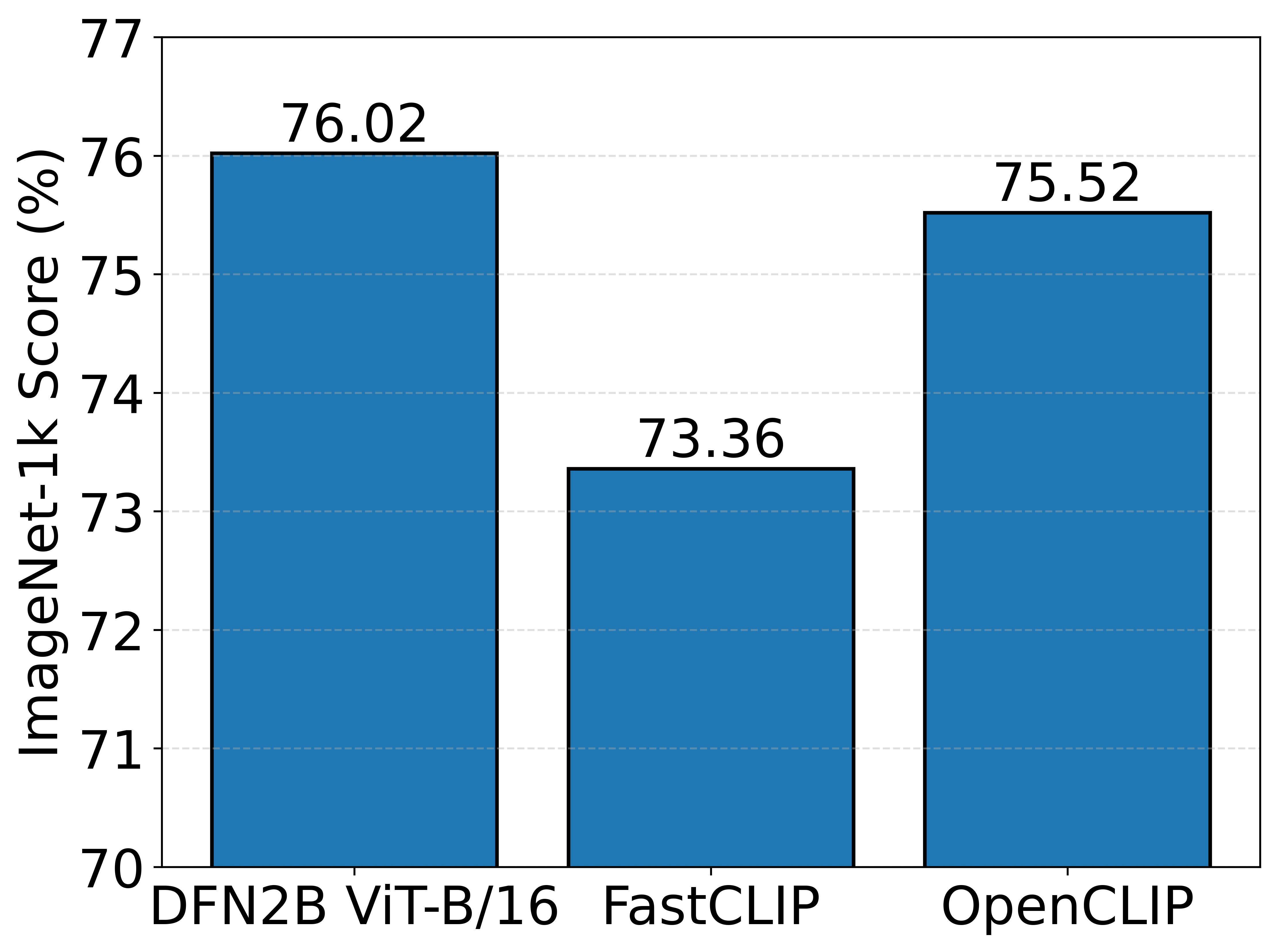}%
    \caption{ImageNet-1k accuracy after \emph{one epoch} of fine-tuning on DFN-12M (a subset of DFN2B). Despite matching the pretraining data distribution, na\"ive fine-tuning methods (FastCLIP, OpenCLIP) exhibit clear performance degradation compared to the DFN2B-pretrained ViT-B/16 baseline.}
    \label{img:drops-im1k}
\end{figure}

{\color{black}
\begin{figure}[ht]
    \centering
    \includegraphics[width=0.45\linewidth]{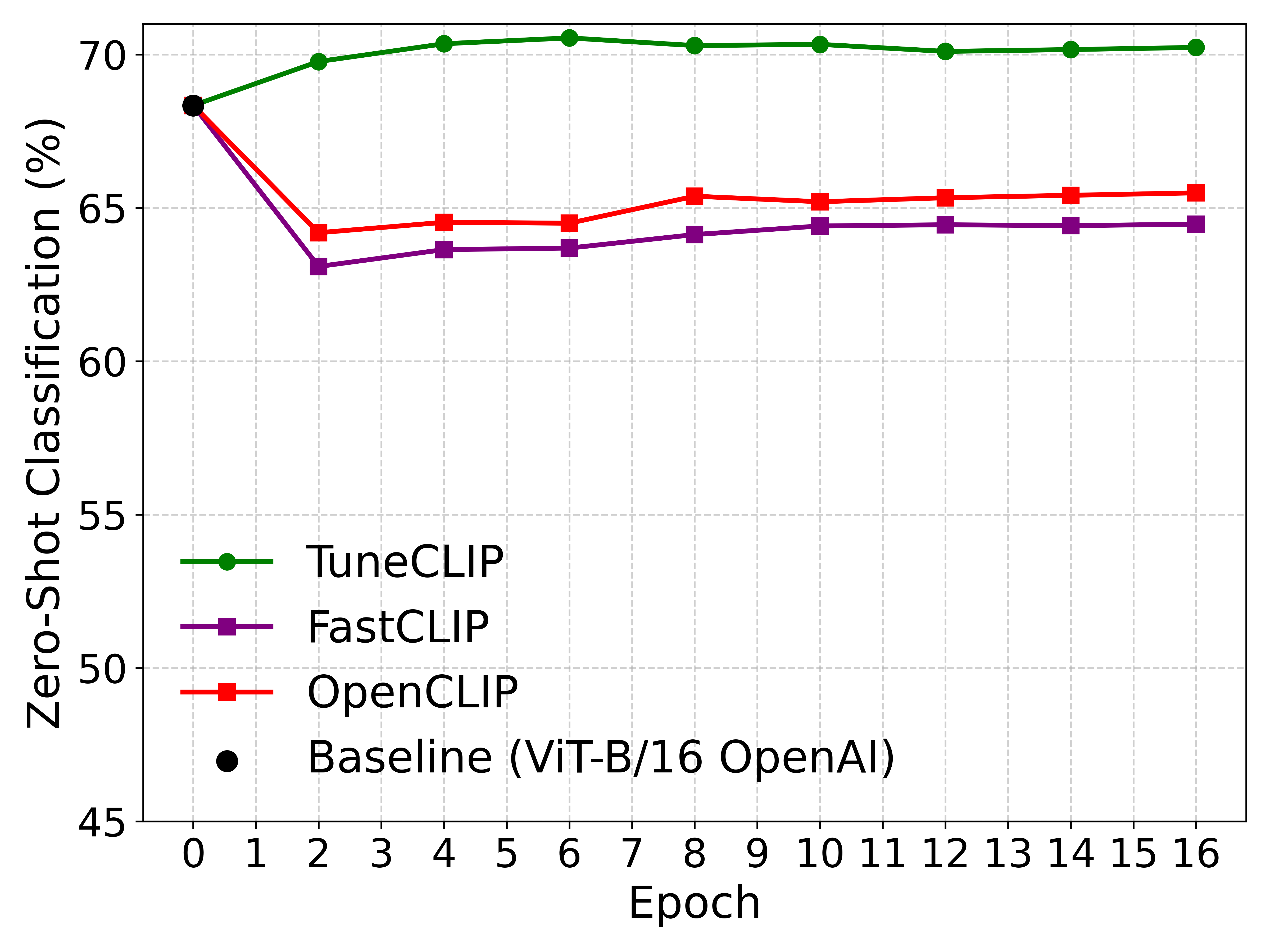}%
    \caption{\color{black}{Extended fine-tuning analysis of TuneCLIP and baselines on ImageNet-1k.}}
    \label{img:longer-ft}
\end{figure}
}

{\color{black}
As shown in Fig.~\ref{img:longer-ft}, extending fine-tuning beyond the used standard few (i.e. 5-epoch) schedule provides only diminishing returns for both the algorithms. Even when trained for up to 15 epochs, the performance curves plateau, indicating that additional compute does not meaningfully change the relative ordering of methods. Importantly, TuneCLIP preserves a consistent improvement over the baselines.}

\newpage

\section{Comprehensive results on 38 DataComp benchmarks}
\label{app:fullresults}
We report the gains of TuneCLIP over the base model, with improvements highlighted in \textcolor{CorrectGreen}{(+)}.

\begin{table}[H]
\centering
\caption{Performance of different CLIP models on DataComp Average. DataComp \cite{gadre2023datacomp} is a highly comprehensive benchmark that spans a diverse collection of datasets, tasks, and distributional variants. Even small improvements on DataComp are particularly meaningful, as they indicate stable gains across heterogeneous and challenging settings rather than isolated benefits on individual datasets.}
\begin{tabular}{llc}
\toprule
\textbf{Base Model} & \textbf{Method} & \textbf{DataComp Average} \\
\midrule

\multirow{4}{*}{\makecell[c]{OpenAI\\ViT-B/32}}
& Baseline      & 52.45 \\
& FastCLIP  & 49.78 \\
& OpenCLIP  & 51.02 \\
& CLIP-Refine  & 51.58 \\
\rowcolor{lightblue!100}
& TuneCLIP  & \textbf{54.34} \textcolor{CorrectGreen}{\textbf{(+1.89)}} \\
\midrule

\multirow{4}{*}{\makecell[c]{OpenAI\\ViT-B/16}}
& Baseline      & 56.26 \\
& FastCLIP  & 53.53 \\
& OpenCLIP  & 55.11 \\
& CLIP-Refine  & 56.18 \\
\rowcolor{lightblue!100}
& TuneCLIP  & \textbf{58.62} \textcolor{CorrectGreen}{\textbf{(+2.36)}} \\
\midrule

\multirow{4}{*}{\makecell[c]{SigLIP\\ViT-B/16}}
& Baseline      & 62.32 \\
& FastCLIP  & 45.80 \\
& OpenCLIP  & 48.10 \\
& CLIP-Refine  & 44.72 \\
\rowcolor{lightblue!100}
& TuneCLIP  & \textbf{63.47} \textcolor{CorrectGreen}{\textbf{(+1.15)}} \\
\midrule

\multirow{4}{*}{\makecell[c]{LAION\\ViT-B/32}}
& Baseline      & 56.94 \\
& FastCLIP  & 55.89 \\
& OpenCLIP  & 56.75 \\
& CLIP-Refine  & \textbf{57.39} \\
\rowcolor{lightblue!100}
& TuneCLIP  & 57.22 \textcolor{CorrectGreen}{\textbf{(+0.28)}} \\
\bottomrule
\end{tabular}
\label{tab:datacomp_average}
\end{table}

\begin{table}[H]
\centering
\caption{Performance of different CLIP models on small-scale classification benchmarks.}
\begin{tabular}{llccc}
\toprule
\textbf{Base Model} & \textbf{Method} & \textbf{CIFAR-10} & \textbf{CIFAR-100} & \textbf{STL-10} \\
\midrule

\multirow{4}{*}{\makecell[c]{OpenAI\\ViT-B/32}}
& Baseline    & 89.83 & 64.23 & 97.13 \\
& FastCLIP    & 90.54 & 69.51 & 92.68 \\
& OpenCLIP    & 91.75 & 71.18 & 96.36 \\
& CLIP-Refine    & 89.94 & 68.15 & 96.35\\
\rowcolor{lightblue!100}
& TuneCLIP    & \textbf{93.63} \textcolor{CorrectGreen}{\textbf{(+3.80)}}
               & \textbf{73.87} \textcolor{CorrectGreen}{\textbf{(+9.64)}}
               & \textbf{97.20} \textcolor{CorrectGreen}{\textbf{(+0.07)}} \\
\midrule

\multirow{4}{*}{\makecell[c]{OpenAI\\ViT-B/16}}
& Baseline    & 90.77 & 66.95 & 98.25 \\
& FastCLIP    & 92.67 & 71.33 & 96.58 \\
& OpenCLIP    & 92.76 & 70.99 & 97.87 \\
& CLIP-Refine    & 91.01 & 69.10 & 97.81\\
\rowcolor{lightblue!100}
& TuneCLIP    & \textbf{94.40} \textcolor{CorrectGreen}{\textbf{(+3.63)}}
               & \textbf{76.14} \textcolor{CorrectGreen}{\textbf{(+9.19)}}
               & \textbf{98.26} \textcolor{CorrectGreen}{\textbf{(+0.01)}} \\
\midrule

\multirow{4}{*}{\makecell[c]{SigLIP\\ViT-B/16}}
& Baseline    & 92.34 & 72.23 & 98.21 \\
& FastCLIP    & 83.66 & 53.86 & 91.18 \\
& OpenCLIP    & 85.06 & 56.77 & 93.81 \\
& CLIP-Refine    & 76.66 & 43.52 & 91.32\\
\rowcolor{lightblue!100}
& TuneCLIP    & \textbf{95.20} \textcolor{CorrectGreen}{\textbf{(+2.86)}}
               & \textbf{79.91} \textcolor{CorrectGreen}{\textbf{(+7.68)}}
               & \textbf{98.37} \textcolor{CorrectGreen}{\textbf{(+0.16)}} \\
\midrule

\multirow{4}{*}{\makecell[c]{LAION\\ViT-B/32}}
& Baseline    & 93.58 & 75.55 & \textbf{96.56} \\
& FastCLIP    & 92.38 & 75.95 & 91.73 \\
& OpenCLIP    & 94.13 & 76.10 & 95.16 \\
& CLIP-Refine    & 94.13 & 76.07 & 96.40\\
\rowcolor{lightblue!100}
& TuneCLIP    & \textbf{94.22} \textcolor{CorrectGreen}{\textbf{(+0.64)}}
               & \textbf{76.46} \textcolor{CorrectGreen}{\textbf{(+0.91)}}
               & 96.47 \textcolor{magenta}{\textbf{(-0.09)}} \\
\bottomrule
\end{tabular}
\label{tab:cifar_stl_all}
\end{table}

\begin{table}[H]
\centering
\caption{Performance of different CLIP models on ImageNet-1k, ImageNet-Sketch, ImageNet-V2.}
\begin{tabular}{llccc}
\toprule
\textbf{Base Model} & \textbf{Method} & \textbf{1K} & \textbf{Sketch} & \textbf{V2} \\
\midrule

\multirow{4}{*}{\makecell[c]{OpenAI\\ViT-B/32}}
& Baseline    & 63.32 & 42.29 & 55.92 \\
& FastCLIP    & 59.59 & 41.26 & 52.25 \\
& OpenCLIP    & 61.11 & 41.86 & 53.84 \\
& CLIP-Refine    & 62.16 & 41.50 & 55.12\\
\rowcolor{lightblue!100}
& TuneCLIP    & \textbf{65.29} \textcolor{CorrectGreen}{\textbf{(+1.97)}}
               & \textbf{45.84} \textcolor{CorrectGreen}{\textbf{(+3.55)}}
               & \textbf{57.45} \textcolor{CorrectGreen}{\textbf{(+1.53)}} \\
\midrule

\multirow{4}{*}{\makecell[c]{OpenAI\\ViT-B/16}}
& Baseline    & 68.34 & 48.24 & 61.88 \\
& FastCLIP    & 64.57 & 46.49 & 56.88 \\
& OpenCLIP    & 65.10 & 46.15 & 58.19 \\
& CLIP-Refine    & 66.36 & 46.46 & 59.51\\
\rowcolor{lightblue!100}
& TuneCLIP    & \textbf{70.57} \textcolor{CorrectGreen}{\textbf{(+2.23)}}
               & \textbf{51.16} \textcolor{CorrectGreen}{\textbf{(+2.92)}}
               & \textbf{64.11} \textcolor{CorrectGreen}{\textbf{(+2.23)}} \\
\midrule

\multirow{4}{*}{\makecell[c]{SigLIP\\ViT-B/16}}
& Baseline    & 76.04 & \textbf{67.92} & 68.93 \\
& FastCLIP    & 57.52 & 12.56 & 49.04 \\
& OpenCLIP    & 59.50 & 10.42 & 51.00 \\
& CLIP-Refine    & 57.67 & 10.86 & 49.00\\
\rowcolor{lightblue!100}
& TuneCLIP    & \textbf{76.41} \textcolor{CorrectGreen}{\textbf{(+0.37)}}
               & 65.78 \textcolor{magenta}{\textbf{(-2.14)}}
               & \textbf{69.02} \textcolor{CorrectGreen}{\textbf{(+0.09)}} \\
\midrule

\multirow{4}{*}{\makecell[c]{LAION\\ViT-B/32}}
& Baseline    & 66.56 & 53.65 & 58.15 \\
& FastCLIP    & 65.26 & 53.12 & 56.92 \\
& OpenCLIP    & 66.31 & 53.73 & 58.46 \\
& CLIP-Refine    & 66.65 & 53.83 & 58.79\\
\rowcolor{lightblue!100}
& TuneCLIP    & \textbf{67.14} \textcolor{CorrectGreen}{\textbf{(+0.58)}}
               & \textbf{54.46} \textcolor{CorrectGreen}{\textbf{(+0.81)}}
               & \textbf{59.10} \textcolor{CorrectGreen}{\textbf{(+0.95)}} \\
\bottomrule
\end{tabular}
\label{tab:imagenet_in}
\end{table}

\begin{table}[H]
\centering
\caption{Performance of CLIP models on ImageNet out-of-distribution variants \cite{yang2023imagenet} (A, O, R, ObjectNet). ObjectNet \cite{barbu2019objectnet} evaluates robustness under real-world viewpoint and background shifts.}
\begin{tabular}{llcccc}
\toprule
\textbf{Base Model} & \textbf{Method} & \textbf{A} & \textbf{O} & \textbf{R} & \textbf{ObjectNet} \\
\midrule

\multirow{4}{*}{\makecell[c]{OpenAI\\ViT-B/32}}
& Baseline    & \textbf{31.55} & 47.75 & 69.33 & 44.31 \\
& FastCLIP    & 24.84 & 47.20 & 65.21 & 44.32 \\
& OpenCLIP    & 27.10 & 48.85 & 67.13 & 46.02 \\
& CLIP-Refine    & 28.74 & 48.90 & 68.46 & 48.15\\
\rowcolor{lightblue!100}
& TuneCLIP    & 30.72 \textcolor{magenta}{\textbf{(-0.83)}}
               & \textbf{51.11} \textcolor{CorrectGreen}{\textbf{(+3.36)}}
               & \textbf{70.67} \textcolor{CorrectGreen}{\textbf{(+1.34)}}
               & \textbf{46.45} \textcolor{CorrectGreen}{\textbf{(+2.14)}} \\
\midrule

\multirow{4}{*}{\makecell[c]{OpenAI\\ViT-B/16}}
& Baseline    & \textbf{49.95} & 42.30 & 77.70 & 55.31 \\
& FastCLIP    & 41.37 & 43.30 & 75.41 & 55.37 \\
& OpenCLIP    & 41.57 & 44.05 & 73.23 & 56.67 \\
& CLIP-Refine    & 44.53 & 43.20 & 74.80 & 56.84\\
\rowcolor{lightblue!100}
& TuneCLIP    & 48.10 \textcolor{magenta}{\textbf{(-1.85)}}
               & \textbf{46.65} \textcolor{CorrectGreen}{\textbf{(+4.35)}}
               & \textbf{77.86} \textcolor{CorrectGreen}{\textbf{(+0.16)}}
               & \textbf{57.08} \textcolor{CorrectGreen}{\textbf{(+1.77)}} \\
\midrule

\multirow{4}{*}{\makecell[c]{SigLIP\\ViT-B/16}}
& Baseline    & 45.41 & 38.15 & \textbf{90.30} & 55.09 \\
& FastCLIP    & 21.80 & 35.85 & 59.08 & 38.72 \\
& OpenCLIP    & 22.73 & 39.50 & 58.27 & 40.08 \\
& CLIP-Refine    & 21.70 & 35.96 & 58.80 & 39.82\\
\rowcolor{lightblue!100}
& TuneCLIP    & \textbf{50.10} \textcolor{CorrectGreen}{\textbf{(+4.69)}}
               & \textbf{41.50} \textcolor{CorrectGreen}{\textbf{(+3.35)}}
               & 88.33 \textcolor{magenta}{\textbf{(-1.97)}}
               & \textbf{67.93} \textcolor{CorrectGreen}{\textbf{(+12.84)}} \\
\midrule

\multirow{4}{*}{\makecell[c]{LAION\\ViT-B/32}}
& Baseline    & 26.26 & 49.95 & 76.43 & 48.81 \\
& FastCLIP    & 24.84 & 48.00 & 76.04 & \textbf{51.72} \\
& OpenCLIP    & 26.92 & 50.40 & 76.05 & 51.01 \\
& CLIP-Refine    & 26.36 & \textbf{51.00 }& 76.37 & 50.37\\
\rowcolor{lightblue!100}
& TuneCLIP    & \textbf{27.04} \textcolor{CorrectGreen}{\textbf{(+0.78)}}
               & 50.85 \textcolor{CorrectGreen}{\textbf{(+0.90)}}
               & \textbf{76.45} \textcolor{CorrectGreen}{\textbf{(+0.02)}}
               & 51.34 \textcolor{CorrectGreen}{\textbf{(+2.53)}} \\
\bottomrule
\end{tabular}
\label{tab:imagenet_ood}
\end{table}

\begin{table}[H]
\centering
\caption{Performance of different CLIP models on VTAB and Fairness. The Visual Task Adaptation Benchmark (VTAB) \cite{zhai2019visual} evaluates performance across 12 diverse tasks. We also report performance on two fairness-oriented datasets, Dollar Street and GeoDE \cite{ramaswamy2023geode}, which measure robustness to geographic and socioeconomic diversity.}
\begin{tabular}{llcc}
\toprule
\textbf{Base Model} & \textbf{Method} & \textbf{VTAB Mean} & \textbf{Fairness Mean} \\
\midrule

\multirow{4}{*}{\makecell[c]{OpenAI\\ViT-B/32}}
& Baseline    & 51.81 & 68.02 \\
& FastCLIP    & 48.51 & 69.52 \\
& OpenCLIP    & 49.23 & 70.05 \\
& CLIP-Refine    & 49.26 & 70.25 \\
\rowcolor{lightblue!100}
& TuneCLIP    & \textbf{53.50} \textcolor{CorrectGreen}{\textbf{(+1.69)}}
               & \textbf{70.73} \textcolor{CorrectGreen}{\textbf{(+2.71)}} \\
\midrule

\multirow{4}{*}{\makecell[c]{OpenAI\\ViT-B/16}}
& Baseline    & 53.80 & 72.45 \\
& FastCLIP    & 50.44 & 74.02 \\
& OpenCLIP    & 52.46 & 74.23 \\
& CLIP-Refine    & 51.65 & \textbf{75.86} \\
\rowcolor{lightblue!100}
& TuneCLIP    & \textbf{54.44} \textcolor{CorrectGreen}{\textbf{(+0.64)}}
               & 74.96 \textcolor{CorrectGreen}{\textbf{(+2.51)}} \\
\midrule

\multirow{4}{*}{\makecell[c]{SigLIP\\ViT-B/16}}
& Baseline    & 60.39 & \textbf{78.48} \\
& FastCLIP    & 49.93 & 69.33 \\
& OpenCLIP    & 53.38 & 71.55 \\
& CLIP-Refine    & 47.88 & 71.59 \\
\rowcolor{lightblue!100}
& TuneCLIP    & \textbf{62.66} \textcolor{CorrectGreen}{\textbf{(+2.27)}}
               & 78.44 \textcolor{magenta}{\textbf{(-0.04)}} \\
\midrule

\multirow{4}{*}{\makecell[c]{LAION\\ViT-B/32}}
& Baseline    & 55.08 & 71.05 \\
& FastCLIP    & 55.55 & 69.69 \\
& OpenCLIP    & 54.68 & 71.94 \\
& CLIP-Refine    & \textbf{55.43} & \textbf{71.72} \\
\rowcolor{lightblue!100}
& TuneCLIP    & 55.09 \textcolor{CorrectGreen}{\textbf{(+0.01)}}
               & 71.03 \textcolor{magenta}{\textbf{(-0.02)}} \\
\bottomrule
\end{tabular}
\label{tab:vtab_mean}
\end{table}

\begin{table}[H]
\centering
\caption{Retrieval performance (Recall@1) on MSCOCO and Flickr datasets.}
\begin{tabular}{llcccc}
\toprule
\textbf{Base Model} & \textbf{Method} 
& \multicolumn{2}{c}{\textbf{MSCOCO}} 
& \multicolumn{2}{c}{\textbf{Flickr}} \\
\cmidrule(lr){3-4} \cmidrule(lr){5-6}
& & \textbf{IR@1} & \textbf{TR@1} & \textbf{IR@1} & \textbf{TR@1} \\
\midrule

\multirow{4}{*}{\makecell[c]{OpenAI\\ViT-B/32}}
& Baseline    & 30.44 & 50.12 & 58.78 & 78.90 \\
& FastCLIP    & 28.75 & 43.23 & 51.49 & 68.19 \\
& OpenCLIP    & 33.33 & 49.79 & 60.92 & 76.10 \\
& CLIP-Refine    & 34.09 & 52.79 & 61.82 & 79.00\\
\rowcolor{lightblue!100}
& TuneCLIP    & \textbf{36.74} \textcolor{CorrectGreen}{\textbf{(+6.30)}}
               & \textbf{56.16} \textcolor{CorrectGreen}{\textbf{(+6.04)}}
               & \textbf{64.71} \textcolor{CorrectGreen}{\textbf{(+5.93)}}
               & \textbf{83.30} \textcolor{CorrectGreen}{\textbf{(+4.40)}} \\
\midrule

\multirow{4}{*}{\makecell[c]{OpenAI\\ViT-B/16}}
& Baseline    & 33.09 & 52.42 & 62.16 & 82.20 \\
& FastCLIP    & 31.25 & 45.80 & 56.45 & 74.00 \\
& OpenCLIP    & 36.46 & 50.77 & 65.11 & 78.90 \\
& CLIP-Refine    & 34.75 & 54.68 & 64.13 & 81.69\\
\rowcolor{lightblue!100}
& TuneCLIP    & \textbf{40.45} \textcolor{CorrectGreen}{\textbf{(+7.36)}}
               & \textbf{59.78} \textcolor{CorrectGreen}{\textbf{(+7.36)}}
               & \textbf{69.66} \textcolor{CorrectGreen}{\textbf{(+7.50)}}
               & \textbf{86.59} \textcolor{CorrectGreen}{\textbf{(+4.39)}} \\
\midrule

\multirow{4}{*}{\makecell[c]{SigLIP\\ViT-B/16}}
& Baseline    & \textbf{47.78} & 65.74 & \textbf{74.68} & 89.10 \\
& FastCLIP    & 26.91 & 36.46 & 48.33 & 61.79 \\
& OpenCLIP    & 34.34 & 44.58 & 57.12 & 70.10 \\
& CLIP-Refine    & 33.91 & 40.40 & 57.59 & 67.40\\
\rowcolor{lightblue!100}
& TuneCLIP    & 47.64 \textcolor{magenta}{\textbf{(-0.14)}}
               & \textbf{66.36} \textcolor{CorrectGreen}{\textbf{(+0.62)}}
               & 74.44 \textcolor{magenta}{\textbf{(-0.24)}}
               & \textbf{89.30} \textcolor{CorrectGreen}{\textbf{(+0.20)}} \\
\midrule

\multirow{4}{*}{\makecell[c]{LAION\\ViT-B/32}}
& Baseline    & 39.34 & 56.32 & 66.78 & \textbf{84.10} \\
& FastCLIP    & 33.37 & 49.34 & 58.66 & 76.99 \\
& OpenCLIP    & 38.34 & 54.42 & 65.03 & 81.30 \\
& CLIP-Refine    & 39.31 & 56.62 & 66.01 & 83.39\\
\rowcolor{lightblue!100}
& TuneCLIP    & \textbf{39.55} \textcolor{CorrectGreen}{\textbf{(+0.21)}}
               & \textbf{57.80} \textcolor{CorrectGreen}{\textbf{(+1.48)}}
               & \textbf{66.82} \textcolor{CorrectGreen}{\textbf{(+0.04)}}
               & 83.20 \textcolor{magenta}{\textbf{(-0.90)}} \\
\bottomrule
\end{tabular}
\label{tab:retrieval_coco_flickr}
\end{table}

\newpage
\section{\textcolor{black}{Compute cost analysis}}
\label{app:cost-analysis}

\begin{table}[ht]
\centering
\caption{\color{black}Batch size specifications of different CLIP variants under our distributed training setup.}
\begin{tabular}{lcc}
\toprule
{\color{black}\textbf{Model}} 
& {\color{black}\textbf{Local Batch Size}} 
& {\color{black}\textbf{Global Batch Size}} \\
\midrule
{\color{black}OpenAI CLIP ViT-B/32}     & {\color{black}512} & {\color{black}4096} \\
{\color{black}OpenAI CLIP ViT-B/16}     & {\color{black}256} & {\color{black}2048} \\
{\color{black}LAION CLIP ViT-B/32}      & {\color{black}512} & {\color{black}4096} \\
{\color{black}SigLIP ViT-B/16}          & {\color{black}256} & {\color{black}2048} \\
\bottomrule
\end{tabular}
\label{tab:batch_specs}
\end{table}

\begin{table}[H]
\centering
\caption{\color{black}Unified comparison of training cost and performance across CLIP backbones.}
\begin{tabular}{llccc}
\toprule
{\color{black}\textbf{Base Model}} 
& {\color{black}\textbf{Method}} 
& {\color{black}\textbf{Wall-Clock Time (hrs)}} 
& {\color{black}\textbf{GPU-hours}} 
& {\color{black}\textbf{DataComp}} \\
\midrule

\multirow{3}{*}{\makecell[c]{\color{black}OpenAI\\\color{black}ViT-B/32}}
& {\color{black}OpenCLIP}  
& {\color{black}2.66} 
& {\color{black}21.28} 
& {\color{black}51.02} \\
& {\color{black}FastCLIP}
& {\color{black}2.22} 
& {\color{black}17.76} 
& {\color{black}49.78} \\
& {\color{black}CLIP-Refine}
& {\color{black}0.58} 
& {\color{black}4.64} 
& {\color{black}51.58} \\
& {\color{black}TuneCLIP}
& {\color{black}\textbf{5.05}} 
& {\color{black}\textbf{40.40}}
& {\color{black}\textbf{54.34}} \\
\midrule

\multirow{3}{*}{\makecell[c]{\color{black}OpenAI\\\color{black}ViT-B/16}}
& {\color{black}OpenCLIP} 
& {\color{black}5.46} 
& {\color{black}43.68} 
& {\color{black}55.11} \\
& {\color{black}FastCLIP} 
& {\color{black}4.21} 
& {\color{black}33.68} 
& {\color{black}53.53} \\
& {\color{black}CLIP-Refine}
& {\color{black}1.01} 
& {\color{black}8.08} 
& {\color{black}56.18} \\
& {\color{black}TuneCLIP}
& {\color{black}\textbf{8.62}} 
& {\color{black}\textbf{68.96}}
& {\color{black}\textbf{58.62}} \\
\midrule

\multirow{3}{*}{\makecell[c]{\color{black}SigLIP\\\color{black}ViT-B/16}}
& {\color{black}OpenCLIP} 
& {\color{black}7.55} 
& {\color{black}60.40} 
& {\color{black}48.10} \\
& {\color{black}FastCLIP} 
& {\color{black}4.28} 
& {\color{black}34.24} 
& {\color{black}45.80} \\
& {\color{black}CLIP-Refine}
& {\color{black}1.07} 
& {\color{black}8.56} 
& {\color{black}44.72} \\
& {\color{black}TuneCLIP}
& {\color{black}\textbf{9.27}} 
& {\color{black}\textbf{74.16}}
& {\color{black}\textbf{63.47}} \\
\midrule

\multirow{3}{*}{\makecell[c]{\color{black}LAION\\\color{black}ViT-B/32}}
& {\color{black}OpenCLIP} 
& {\color{black}3.10} 
& {\color{black}24.80} 
& {\color{black}56.75} \\
& {\color{black}FastCLIP} 
& {\color{black}2.33} 
& {\color{black}18.64} 
& {\color{black}55.89} \\
& {\color{black}CLIP-Refine}
& {\color{black}0.55} 
& {\color{black}4.40} 
& {\color{black}\textbf{57.39}} \\
& {\color{black}TuneCLIP}
& {\color{black}\textbf{4.32}} 
& {\color{black}\textbf{36.24}}
& {\color{black}57.22} \\
\bottomrule
\end{tabular}
\label{tab:combined_costs}
\end{table}

\textcolor{black}{
Table~\ref{tab:batch_specs} summarizes the local and global batch sizes used for
each backbone,
where the global batch size is given by
$B_{\text{global}} = 8 \times B_{\text{local}}$. Across all backbones,
Table~\ref{tab:combined_costs} additionally reports the wall-clock time
(measured as the elapsed time between the start and end of fine-tuning) and the
corresponding GPU-hours for all approaches. We compute
GPU-hours using the standard relation
\[
\text{GPU-hours} = \text{wall-clock time (hours)} \times \text{\#GPUs},
\]
so that, in our case, GPU-hours directly reflect wall-clock time scaled by a
factor of $8$. Since TuneCLIP uses a two-stage procedure (OSR followed by HGCL
fine-tuning), it naturally incurs higher compute than single-stage baselines,
typically increasing the wall-clock time by about $1.5$--$2\times$ for a given
backbone. However, this additional cost remains modest and is consistently mirrored by improved performance across all evaluated models. Thus, TuneCLIP provides a favorable cost--performance trade-off with relatively small extra computational overhead compared to baselines and it reliably converts additional compute into gains on related benchmarks.} 

\textcolor{black}{
In addition to reporting the full TuneCLIP cost, we further dissect its
two-stage schedule into OSR-only and HGCL-only components in
Table~\ref{tab:combined_costs_osr_hgcl}. Across all backbones, each stage
incurs a wall-clock time comparable to standard single-stage fine-tuning
for OpenCLIP/FastCLIP.}

\begin{table}[H]
\centering
{\color{black}
\caption{\color{black}Isolated computational cost of the OSR and HGCL stages in TuneCLIP.}
\label{tab:combined_costs_osr_hgcl}
\begin{tabular}{llcc}
\toprule
{\color{black}\textbf{Base Model}} 
& {\color{black}\textbf{Method}} 
& {\color{black}\textbf{Wall-Clock (hrs)}} 
& {\color{black}\textbf{GPU-hours}} \\
\midrule

\multirow{4}{*}{\makecell[c]{\color{black}OpenAI\\\color{black}ViT-B/32}}
& {\color{black}OpenCLIP}         & {\color{black}2.66} & {\color{black}21.28} \\
& {\color{black}FastCLIP}        & {\color{black}2.22} & {\color{black}17.76} \\
& {\color{black}OSR only}        & {\color{black}\textbf{2.50}} & {\color{black}\textbf{20.00}} \\
& {\color{black}HGCL only}       & {\color{black}\textbf{2.55}} & {\color{black}\textbf{20.40}} \\
\midrule

\multirow{4}{*}{\makecell[c]{\color{black}OpenAI\\\color{black}ViT-B/16}}
& {\color{black}OpenCLIP}         & {\color{black}5.46} & {\color{black}43.68} \\
& {\color{black}FastCLIP}        & {\color{black}4.21} & {\color{black}33.68} \\
& {\color{black}OSR only}        & {\color{black}\textbf{4.27}} & {\color{black}\textbf{34.16}} \\
& {\color{black}HGCL only}       & {\color{black}\textbf{4.35}} & {\color{black}\textbf{34.80}} \\
\midrule

\multirow{4}{*}{\makecell[c]{\color{black}SigLIP\\\color{black}ViT-B/16}}
& {\color{black}OpenCLIP}         & {\color{black}7.55} & {\color{black}60.40} \\
& {\color{black}FastCLIP}        & {\color{black}4.28} & {\color{black}34.24} \\
& {\color{black}OSR only}        & {\color{black}\textbf{4.52}} & {\color{black}\textbf{36.16}} \\
& {\color{black}HGCL only}       & {\color{black}\textbf{4.75}} & {\color{black}\textbf{38.00}} \\
\midrule

\multirow{4}{*}{\makecell[c]{\color{black}LAION\\\color{black}ViT-B/32}}
& {\color{black}OpenCLIP}         & {\color{black}3.10} & {\color{black}24.80} \\
& {\color{black}FastCLIP}        & {\color{black}2.33} & {\color{black}18.64} \\
& {\color{black}OSR only}        & {\color{black}\textbf{2.20}} & {\color{black}\textbf{17.60}} \\
& {\color{black}HGCL only}       & {\color{black}\textbf{2.12}} & {\color{black}\textbf{16.96}} \\
\bottomrule
\end{tabular}
}
\end{table}

\textbf{On FLOPS.} Since all baseline methods use the same vision and text encoders, GPUs, and training configurations at a given model scale, the FLOPs per training step are effectively identical.

\section{\textcolor{black}{Comparing other Cold-start bias mitigation strategies}}

\textcolor{black}{
To evaluate simpler cold-start bias mitigations, we consider two realistic alternatives to OSR. First, we apply a short learning-rate warm-up of $500$ iterations using momentum SGD, gradually increasing the learning rate from $1\times10^{-6}$ to $1\times10^{-5}$ before switching to the standard fine-tuning stage with the same optimizer. Second, we simulate a large-batch warm-up by computing gradients with a larger global batch size (e.g., increasing OpenAI CLIP ViT-B/16's batch size from $2048$ to $4096$) while keeping model weights frozen, allowing gradient moments to accumulate before performing normal fine-tuning in the second stage. Infact, this serves as a cheaper approximation to OSR. As shown in Table~\ref{tab:cold_start_baselines}, both strategies provide mild stabilization but yield noticeably smaller improvements than OSR, indicating that OSR remains most effective \& reliable approach for mitigating cold-start bias.}

\label{app:cold-start_bias}
\begin{table}[h]
\centering
{\color{black}
\caption{\color{black}Comparison of cold-start mitigation strategies for TuneCLIP (on OpenAI CLIP ViT-B/16).}
\label{tab:cold_start_baselines}
\begin{tabular}{lccccc}
\toprule
\textbf{Cold-Start Mitigation}& \textbf{Stage II}& \textbf{OSR} & \textbf{IN \& Variants} & \textbf{Retrieval} & \textbf{DataComp} \\
\midrule
Base Model &N/A & $\times$ & 57.67 & 57.46 & 56.26 \\
Momentum SGD&Momentum SGD & $\times$ & 54.65 & 59.32 & 54.82 \\
Larger Batch  Gradients & AdamW  & $\times$ & 58.27 & 62.92 & 58.05 \\
OSR & Momentum SGD& $\checkmark$ & 57.99 & 62.08 & 57.82 \\
\rowcolor{lightblue!100}
OSR& AdamW  & $\checkmark$ & \textbf{59.36} & \textbf{64.12} & \textbf{58.62} \\
\bottomrule
\end{tabular}
}
\end{table}

\section{TuneCLIP on State-of-the-art CLIP}
\label{app:h14}

TuneCLIP achieves state-of-the-art performance on ImageNet and its distributional variants, improving accuracy from 71.8\% to 73.23\%. On retrieval and DataComp, the results are slightly lower, but remain within a tolerable band of 1\% relative to the baseline.

\begin{table}[H]
\centering
\caption{Performance of TuneCLIP on SOTA ViT-H/14-quickgelu across evaluation suites.}
\begin{tabular}{lcc}
\toprule
\textbf{Category} & \textbf{Baseline} & \textbf{TuneCLIP} \\
\midrule
\rowcolor{lightblue!100}
IN \& Variants   & 71.80 & \textbf{73.23} \\
Retrieval        & 74.78 & 73.78 \\
DataComp         & 69.61 & 69.23 \\
\bottomrule
\end{tabular}
\label{tab:vith14_results}
\end{table}

\section{TuneCLIP on CNN-based CLIP}
\label{app:resnet}
\begin{table}[H]
\centering
\caption{Performance of TuneCLIP on OpenAI CLIP with a ResNet-50 backbone.}
\label{tab:resnet50_results}
\begin{tabular}{lcc}
\toprule
\textbf{Method} & \textbf{ImageNet-1k} & \textbf{MS COCO} \\
\midrule
OpenAI CLIP ResNet-50 & 59.82 & 38.69 \\
\rowcolor{lightblue!100}
TuneCLIP & \textbf{60.95} & \textbf{45.19} \\
\bottomrule
\end{tabular}
\end{table}

To evaluate TuneCLIP beyond Vision Transformer backbones, we apply it to
the OpenAI CLIP ResNet-50 model. As shown in Table~\ref{tab:resnet50_results}, TuneCLIP improves performance
across classification as well as retrieval. These results demonstrate its applicability to CNN-based CLIP models.

\newpage
\section{False Negative mitigation through HGCL}
\label{app:fn_hgcl}

\setlength{\fboxrule}{1pt} 
\setlength{\fboxsep}{1pt}  

\begin{figure*}[htbp]
  \centering
  
  \colorbox{red!40}{%
    \setlength{\fboxsep}{1pt}%
    \begin{subfigure}{0.40\textwidth}
      \centering
      \includegraphics[width=\linewidth]{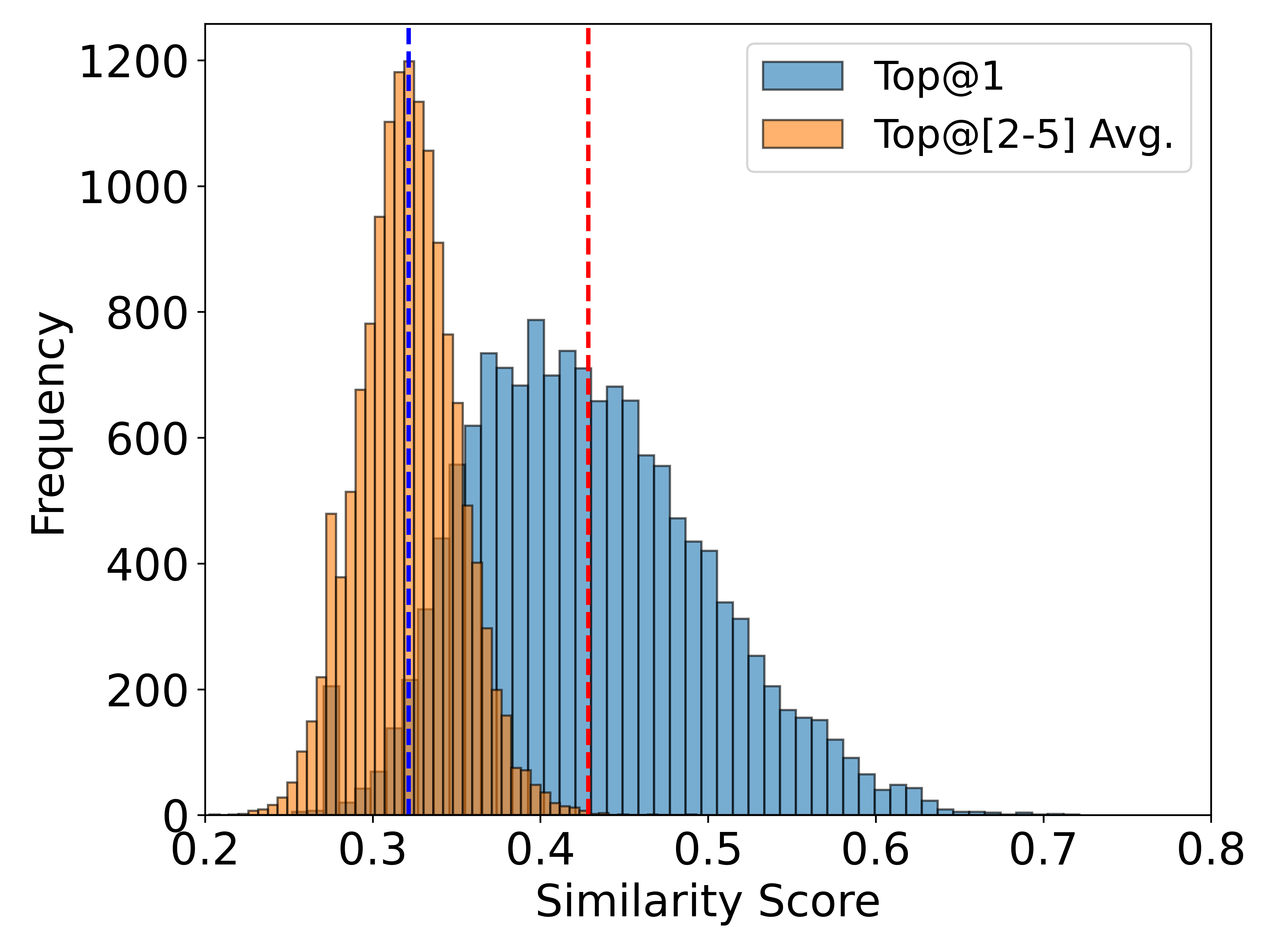}
      \caption{OSR+GCL (TP vs FN)}
      \label{fig:gcl_dist}
    \end{subfigure}%
  }\hfill
  
  \fbox{%
    \begin{subfigure}{0.40\textwidth}
      \centering
      \includegraphics[width=\linewidth]{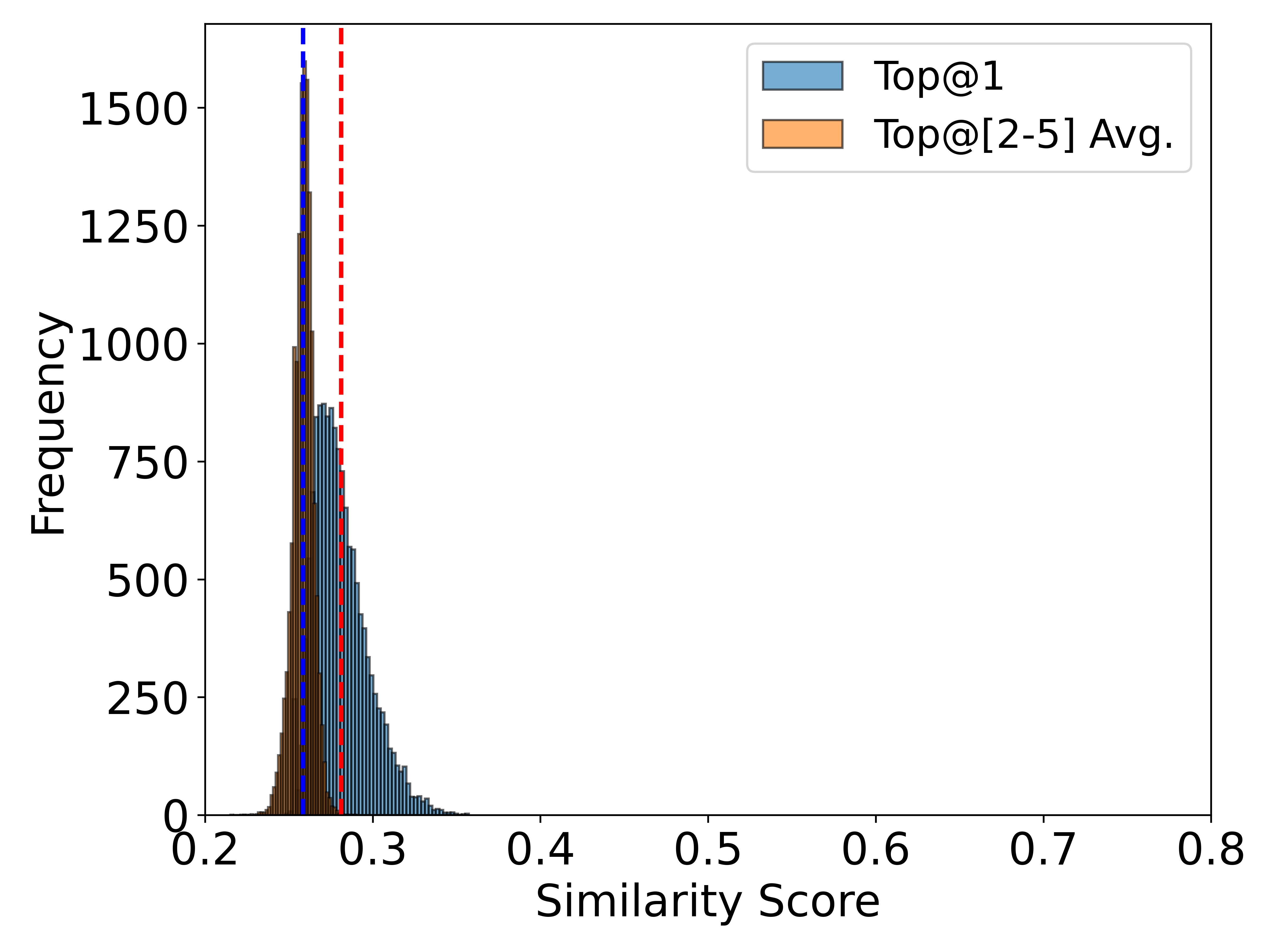}
      \caption{OSR+HGCL (TP vs FN)}
      \label{fig:hgcl_dist}
    \end{subfigure}\hfill
    \begin{subfigure}{0.40\textwidth}
      \centering
      \includegraphics[width=\linewidth]{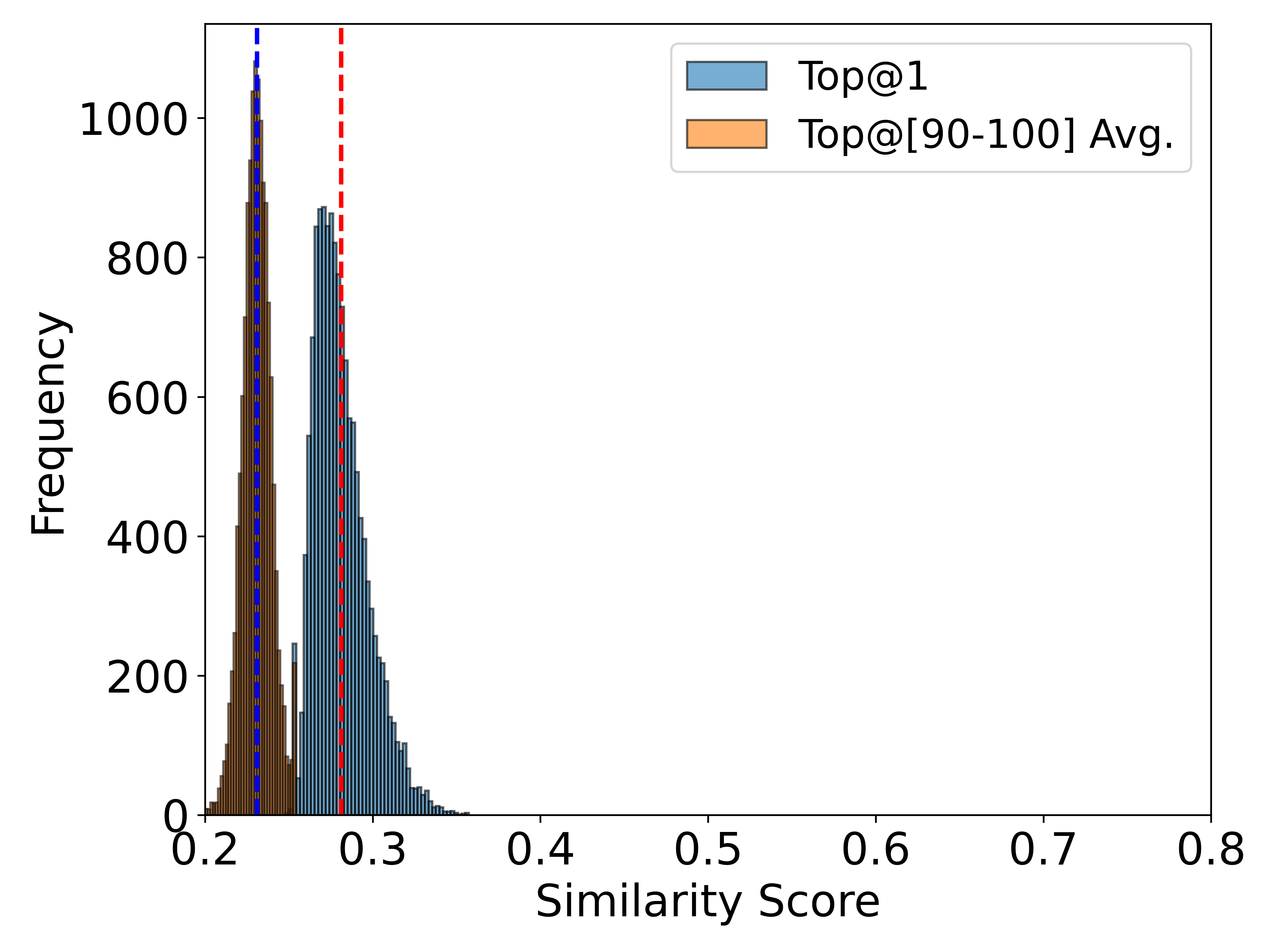}
      \caption{OSR+HGCL (TP vs TN)}
      \label{fig:TP_TN_hgcl}
    \end{subfigure}
  }

  \caption{
    OSR+GCL shows higher variance among false negatives (std.\ dev.\ $0.030$). While, OSR+HGCL yields a substantially more compact distribution (std.\ dev.\ $0.0061$), reducing false negative bias and at the same time ensuring a clear separation between true positives ($\mu=0.28$) and true negatives ($\mu=0.23$).
  }
  \label{fig:distributions-combined}
\end{figure*}

As shown in Figures~\ref{fig:gcl_dist} and~\ref{fig:hgcl_dist}, 
HGCL compresses the gap between true positives and false negatives, producing distributions that are both more compact and overlapping than those under GCL. This analysis, conducted on 15{,}000 randomly sampled pairs from DFN-12M, highlights HGCL’s capacity to reduce variance and counteract false-negative bias. By preserving higher similarity for semantically related negatives, HGCL prevents over-suppression and encourages the model to maintain only a fine-grained distinctions across closely related concepts (true positives and false negatives). This in turn improves generalization, since the learned similarity space better reflects true semantic structure rather than being distorted by aggressive penalization of false negatives. Additionally, to approximate true negatives, we sample from the bottom-ranked retrievals (Top@[90-100]), as these examples are least likely to share semantic overlap with the query and thus provide a reliable baseline for unrelated pairs. Table~\ref{tab:false_negatives} shows qualitative examples from the above distributions.

\newcommand{\rowimgheight}{2.0cm}

\begin{table}[t]
\centering
\scriptsize
\caption{Qualitative examples of false negatives (text captions) and their similarity scores with the anchor image. 
Here, \colorbox{blue!10}{$s_1$} denotes the similarity under \colorbox{blue!10}{OSR+GCL} and \colorbox{blue!10}{$s_2$} under \colorbox{blue!10}{OSR+HGCL}. 
These examples illustrate how OSR+HGCL mitigates excessive suppression of false negatives, allowing them to retain higher similarity scores compared to OSR+GCL. 
Conversely, false negatives that already exhibit reasonable similarity with the anchor tend to remain stable or slightly reduced, compensating for cases where suppression was more severe. 
Overall, this yields a calibrated similarity structure in which false negatives are assigned scores that better reflect semantic relatedness.}

\begin{tabular}{|m{2.8cm}|m{10.2cm}|}
\hline
\textbf{Anchor Image} & \raggedright\arraybackslash \textbf{False Negative Captions \colorbox{blue!10}{($s_1$, $s_2$)} in Top-5 Retrieved texts} \\
\hline

\parbox[c]{\linewidth}{%
  \centering
  \includegraphics[height=\rowimgheight,width=\linewidth,keepaspectratio]{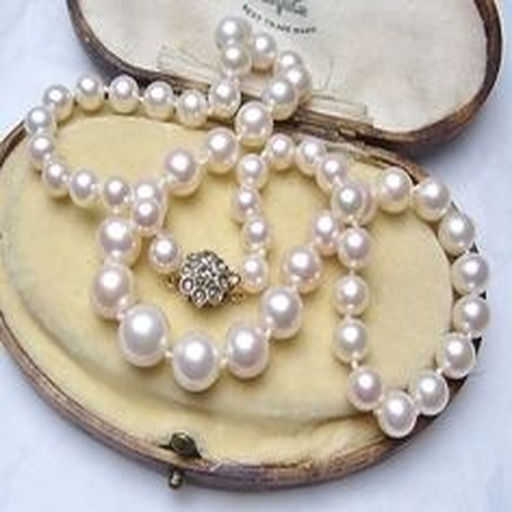}\\
  \colorbox{yellow!15}{%
    \parbox{0.95\linewidth}{\centering \scriptsize \textit{Caption: FABULOUS PEARL NECKLACE DIAMOND CLUSTER 14K GOLD CLASP | eBay}}
  }
}
&
\begin{itemize}\setlength\itemsep{0.2em}
\item Women's Chain Necklaces \;\colorbox{blue!10}{\textbf{s$_1$=0.2053, s$_2$=0.2467}}
\item necklace image New Arrivals \;\colorbox{blue!10}{\textbf{s$_1$=0.1720, s$_2$=0.2429}}
\end{itemize} \\
\hline

\parbox[c]{\linewidth}{%
  \centering
  \includegraphics[height=\rowimgheight,width=\linewidth,keepaspectratio]{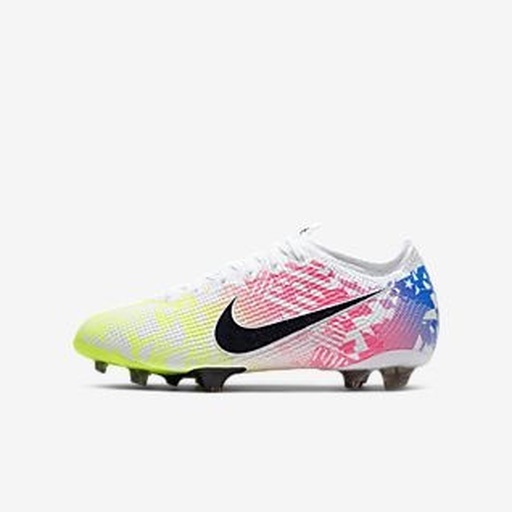}\\
  \colorbox{yellow!15}{%
    \parbox{0.95\linewidth}{\centering \scriptsize \textit{Caption: Boys' Soccer Cleats 0026 Shoes. Nike.com}}
  }
}
&
\begin{itemize}\setlength\itemsep{0.2em}
\item running shoes \;\colorbox{blue!10}{\textbf{s$_1$=0.1561, s$_2$=0.2346}}
\end{itemize} \\
\hline

\parbox[c]{\linewidth}{%
  \centering
  \includegraphics[height=\rowimgheight,width=\linewidth,keepaspectratio]{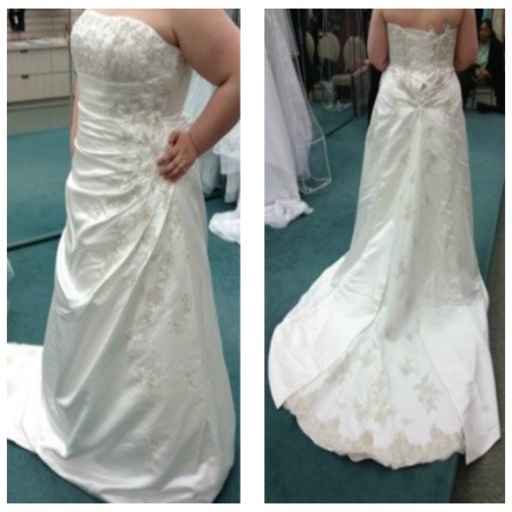}\\
  \colorbox{yellow!15}{%
    \parbox{0.95\linewidth}{\centering \scriptsize \textit{Caption: dress}}
  }
}
&
\begin{itemize}\setlength\itemsep{0.2em}
\item muslim wedding dresses 3d flower burgundy muslim wedding dresses 2018 arabic custom plus \;\colorbox{blue!10}{\textbf{s$_1$=0.1713, s$_2$=0.2238}}
\item Evening Gowns For Mother Of The Bride In Singapore - Prom Dresses \;\colorbox{blue!10}{\textbf{s$_1$=0.1913, s$_2$=0.2333}}
\end{itemize} \\
\hline

\parbox[c]{\linewidth}{%
  \centering
  \includegraphics[height=\rowimgheight,width=\linewidth,keepaspectratio]{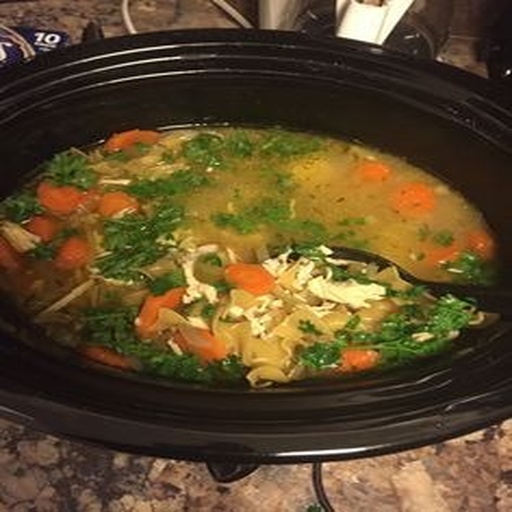}\\
  \colorbox{yellow!15}{%
    \parbox{0.95\linewidth}{\centering \scriptsize \textit{Caption: Chicken noodle soup}}
  }
}
&
\begin{itemize}\setlength\itemsep{0.2em}
\item top down view of bowl filled with white bean tomatillo soup with items surrounding. \;\colorbox{blue!10}{\textbf{s$_1$=0.2061, s$_2$=0.2047}}
\item Butternut soup with sriracha. Made it. \;\colorbox{blue!10}{\textbf{s$_1$=0.2046, s$_2$=0.2354}}
\end{itemize} \\
\hline

\parbox[c]{\linewidth}{%
  \centering
  \includegraphics[height=\rowimgheight,width=\linewidth,keepaspectratio]{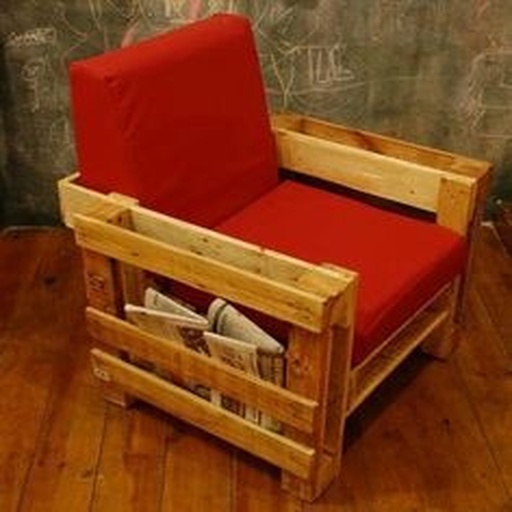}\\
  \colorbox{yellow!15}{%
    \parbox{0.95\linewidth}{\centering \scriptsize \textit{Caption: pallets chair}}
  }
}
&
\begin{itemize}\setlength\itemsep{0.2em}
\item B32 Office Chair by Armet 3 \;\colorbox{blue!10}{\textbf{s$_1$=0.0747, s$_2$=0.2189}}
\item Burlap Ruffle Chair \#burlap \#ruffle \#furniture \;\colorbox{blue!10}{\textbf{s$_1$=0.3096, s$_2$=0.2583}}
\end{itemize} \\
\hline

\parbox[c]{\linewidth}{%
  \centering
  \includegraphics[height=\rowimgheight,width=\linewidth,keepaspectratio]{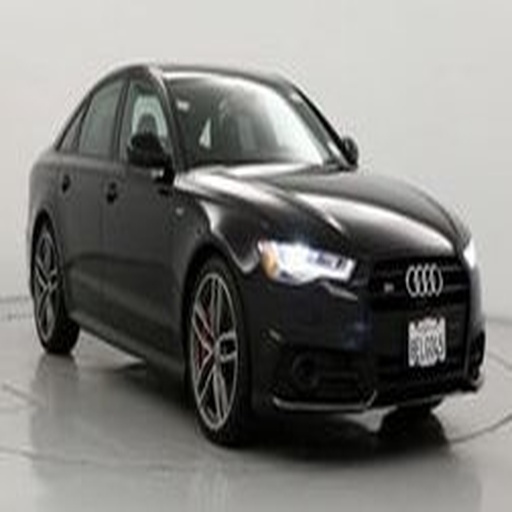}\\
  \colorbox{yellow!15}{%
    \parbox{0.95\linewidth}{\centering \scriptsize \textit{Caption: 2017 Audi S6 4.0T quattro Premium Plus}}
  }
}
&
\begin{itemize}\setlength\itemsep{0.2em}
\item gebraucht Audi S8 plus V8 4.0TFSI tiptr. UPE 154.100,- HeadUp/SD \;\colorbox{blue!10}{\textbf{s$_1$=0.2358, s$_2$=0.2272}}
\item Audi Q3 und Audi RS Q3: Facelift und mehr Leistung. \;\colorbox{blue!10}{\textbf{s$_1$=0.1056, s$_2$=0.2174}}
\end{itemize} \\
\hline

\parbox[c]{\linewidth}{%
  \centering
  \includegraphics[height=\rowimgheight,width=\linewidth,keepaspectratio]{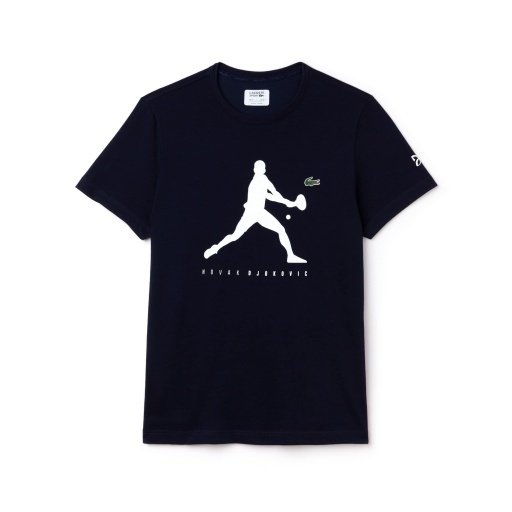}\\
  \colorbox{yellow!15}{%
    \parbox{0.95\linewidth}{\centering \scriptsize \textit{Caption: Men's Crew Neck Jersey T-Shirt - Support With Style Collection for Novak Djokovic}}
  }
}
&
\begin{itemize}\setlength\itemsep{0.2em}
\item Unique Scoop Collar Printed T Shirt \;\colorbox{blue!10}{\textbf{s$_1$=0.1978, s$_2$=0.2433}}
\item T-shirt z nadrukiem - white \;\colorbox{blue!10}{\textbf{s$_1$=0.1707, s$_2$=0.2407}}
\item HX Hot Sale Colorful Skull 3D Print Harajuku T Shirt Grim Reaper Skull Casual T Shirt Men/women Streetwear T Shirt Tops HX768 1 \;\colorbox{blue!10}{\textbf{s$_1$=0.1560, s$_2$=0.2301}}
\end{itemize} \\
\hline

\end{tabular}
\label{tab:false_negatives}
\end{table}

\end{document}